\documentclass[pdflatex,sn-apa]{sn-jnl}

\usepackage{graphicx}%
\usepackage{multirow}%
\usepackage{amsmath,amssymb,amsfonts}%
\usepackage{amsthm}%
\usepackage{mathrsfs}%
\usepackage[title]{appendix}%
\usepackage{xcolor}%
\usepackage{textcomp}%
\usepackage{manyfoot}%
\usepackage{booktabs}%
\usepackage{algorithm}%
\usepackage{algorithmicx}%
\usepackage{algpseudocode}%
\usepackage{listings}%

\usepackage{pdflscape} 
\usepackage{etoolbox}

\usepackage{hyperref}

\DeclareMathOperator{\E}{\mathbb{E}}

\usepackage{tabularx}
\usepackage{longtable}

\newtoggle{anonymous}
\togglefalse{anonymous}

\newtoggle{arxiv}
\toggletrue{arxiv}

\iftoggle{arxiv}{}{
  \usepackage{xr}
  \externaldocument{supplementary_material}    
}

\theoremstyle{thmstyleone}%
\newtheorem{property}{Property} 
\newtheorem{corollary}{Corollary} 

\theoremstyle{thmstyletwo}%

\theoremstyle{thmstylethree}%

\begin{document}

\title[Who is the root?]{Who is the root in a syntactic dependency structure?}


\iftoggle{anonymous}{
   \author*[1]{...}
   \affil[1]{...}
}
{   
   \author*[1]{\fnm{Ramon} \sur{Ferrer-i-Cancho}}\email{rferrericancho@cs.upc.edu}

   \author[2]{\fnm{Marta} \sur{Arias}}\email{marias@cs.upc.edu}


   \affil*[1]{Quantitative, Mathematical and Computational Linguistics Research Group, \orgdiv{Department of Computer Science}, \orgname{Universitat Politècnica de Catalunya}, \orgaddress{\street{Jordi Girona Salgado 1-3}, \city{Barcelona}, \postcode{08034}, \state{Catalonia}, \country{Spain}}}
}


\abstract{The syntactic structure of a sentence can be described as a tree that indicates the syntactic relationships between words. In spite of significant progress in unsupervised methods that retrieve the syntactic structure of sentences, guessing the right direction of edges is still a challenge. As in a syntactic dependency structure edges are oriented away from the root, the challenge of guessing the right direction can be reduced to finding an undirected tree and the root. The limited performance of current unsupervised methods demonstrates the lack of a proper understanding of what a root vertex is from first principles.
We consider an ensemble of centrality scores, some that only take into account the free tree (non-spatial scores) and others that take into account the position of vertices (spatial scores).
We test the hypothesis that the root vertex is an important or central vertex of the syntactic dependency structure. 
We confirm the hypothesis in the sense that root vertices tend to have high centrality and that vertices of high centrality tend to be roots. The best performance in guessing the root is achieved by novel scores that only take into account the position of a vertex and that of its neighbours.
We provide theoretical and empirical foundations towards a universal notion of rootness from a network science perspective.}




\maketitle



\tableofcontents

\textcolor{blue}{
\begin{itemize}
\item
The hyperlinking of the PDF does not work, e.g. by clicking on Table B9 one does not go to Table B9 in the Appendix but, incorrectly, to Table 9 in the main article. If you wish to check a table from the Appendix please search for it manually by its text label.
\item
Within proofs, the journal template increases font size for text in itemize or enumerate environments.
\end{itemize}
}

\section{Introduction}

\label{sec:introduction}

\begin{figure}[!b]
\begin{center}
\includegraphics[width = 0.6\textwidth]
{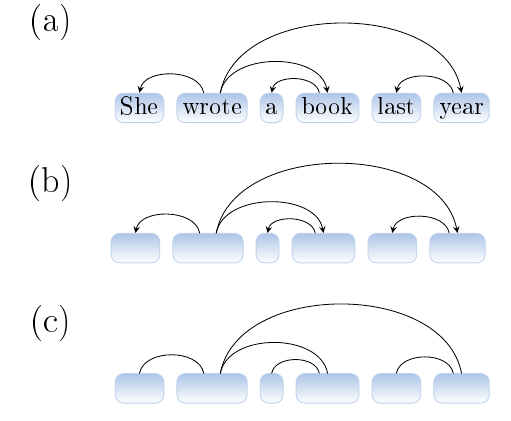}
\end{center}
\caption{\label{fig:syntactic_dependency_structures} (a) The syntactic dependency structure of a sentence. (b) The corresponding rooted tree. (c) The corresponding free tree. }
\end{figure}

The syntactic structure of a sentence can be described as a rooted tree that indicates the syntactic relationships between its words as in Figure \ref{fig:syntactic_dependency_structures} \citep{Melcuk1988}. In these trees, there is a particular vertex, called root, that has no incoming edges. The backbone of the tree is the free tree (Figure \ref{fig:syntactic_dependency_structures} (c)), that is the undirected tree that results from removing link direction from the rooted tree (Figure \ref{fig:syntactic_dependency_structures} (b)).  

The question of who is the root of a syntactic dependency structure arises in two contexts. In a theoretical context, when one wishes to understand the foundations of syntactic dependency structures and characterize what a root vertex is.
In the context of natural language processing, there has been considerable research on extracting those trees automatically from texts using unsupervised methods \citep{Marecek2016a,Han2020a}. These methods are critical when there is no training dataset because very little is known about that language, e.g., in low-resourced languages, languages that deviate from the safe frame of Indo-European languages or that do not have a large number of speakers. A serious limitation of these methods is that they misidentify arc directions. Namely, these methods often guess correctly that two words $u$ and $v$ are linked but they fail to guess if $u \rightarrow v$ or $u \leftarrow v$. For these reason, they are often evaluated just in terms of whether they have guessed that there is an undirected edge between $u$ and $v$ \citep{Marecek2016a}. 
There are distinct ways the right direction of the arcs can be guessed. One is by identifying the root in the free tree and then assigning arc direction consistently from that root (from the root away to the leaves). \footnote{Notice that, in a rooted tree, no two vertices can point to the same vertex.} That takes us back to the theoretical question of who is the root of a syntactic dependency structure from first principles thinking. 

The general objective of the present article is two-fold. First, to achieve a theoretical understanding of what the root of a sentence is. Our focus is on generalizations that are valid across languages (rather than what a root is in a specific language). Second, to contribute to the development of unsupervised methods to guess the root node of a free tree when the root of a tree is unknown or unreliable, either as a part of powerful parsing methods or simply methods to assess the reliability of the root obtained by an unsupervised parser. 

Alternatively, we could investigate our research question by means of other approaches to the syntactic structure of a sentence such as headed context-free grammars. However, syntactic dependency trees are a particular case of spatial network where the network structure is a rooted tree and the space is defined by the linear order of the vertices in a sentence \citep{Barthelemy2011a}. Thus, our choice fully aligns with this forum, that is a journal of network science. 
Crucially, our choice of the dependency grammar framework (figure \ref{fig:syntactic_dependency_structures}) is supported by the community of computational linguistics that gradually abandoned phrase structure grammars (as headed context-free grammar) in favour of syntactic dependency grammar to the point that dependency structure became the {\em de facto} standard \citep{Kuebler2009a}. Dependency grammar offers several advantages as a formalism \citep{Covington2001a} and is neurophysiologically supported \citep{Chen2018a}. A key advantage of dependency grammar is its simplicity and parsimony \citep{Covington2001a, Chen2018a} which eases theoretical reasoning. Ascertaining if such simplicity constitutes a severe limit for our purposes is beyond the reach of the present article.

Here we will present and test unsupervised methods to guess the root vertex in a simplified setting that still sheds light on a language-independent notion of rootness. 
Given a sentence, the model guesses the root based only on information from that sentence (other sentences are neglected) and the kind of information the model exploits from a syntactic dependency structure is restricted. As for the latter, the syntactic structure of a sentence can be seen as a three-fold entity consisting of 
\begin{enumerate}
\item
A rooted tree.
\item
A linear arrangement. Typically a table that indicates the position in the sentence of every vertex of the tree.
\item
Additional labels. Labels attached to vertices of the rooted tree indicate the word form of each vertex. Labels attached to edges indicate the syntactic function (e.g., subject or direct object for verb arguments).
\end{enumerate}
The methods introduced in the present article discard the additional labels and focus on exploiting the free tree of the syntactic dependency structure.
To illustrate the setting, the model cannot exploit
\begin{enumerate}
\item
Information about the word that corresponds to the vertex (its string, its part-of-speech,...).
\item
The language of the sentence where the vertex appears. This excludes precious information such as the branching direction in the language \citep{Liu2010a} or the likely placement of the main verb.  
\item
As the prediction has to be made on the free tree, information on the rooted tree cannot be used. For instance, one cannot use the in-degree of the vertex. That would make the problem trivial because the root vertex is the vertex that has zero in-degree. 
\item
Information outside the sentence, namely ontologies or word embeddings. 
\end{enumerate}

The specific goal of this article is two-fold. First, to test the hypothesis that the root vertex is an important or central vertex of a syntactic dependency structure across languages. Second, to bring order to the diversity and quantity of existing centrality scores for trees. 
S{\o}gaard proposed that a ``dependency structure is, among other things, a partial order on the nodes in terms of centrality or saliency.'' \citep{Soegaard2012a}.
In particular, he hypothesized that roots are words of high PageRank \citep{Soegaard2012a,Soegaard2012b} and presented an unsupervised parsing method that operates in two phases. In the 1st phase, various rules are used to build a directed graph representation of the sentence that is used to compute the PageRank of each word. In the second phase, a parsing algorithm obtains the syntactic dependency structure representation of the sentence by setting the word of highest PageRank as the root.
Here we will test a specific version of the hypothesis, namely that the root vertex is an important or central vertex of the free tree or both the free tree and the linear order of a syntactic dependency structure.
Our approach differs from S{\o}gaard's in the sense that he used the importance of words to find a rooted tree when there is still no rooted or free tree available, while we assume that a free tree is already available. In addition, the 1st stage of the algorithm uses additional linguistic information (e.g. the word form, part of speech tags) to build a graph that is used to compute the PageRank of the words in the sentence. We exclude that kind of information from the root-finding problem.

In particular, we tackle the problem of guessing the root by means of centrality scores from two perspectives. First, as a binary classification problem where the goal is to predict whether a vertex is a root or not. Second, as a ranking problem, where the goal is to sort vertices by their centrality, ideally ranking the root at the top. 
   
The organization of the remainder of the article is as follows. Section \ref{sec:centrality} reviews the centrality scores that will be used in this article. 
Section \ref{subsec:centrality_toolbox} presents the scores that are borrowed from the standard toolbox while Section \ref{subsec:new_spatial_centrality_scores} and \ref{subsec:new_non_spatial_centrality_scores} present new centrality scores that are put forward in this article. Section \ref{subsec:relationships_among_centrality_scores} presents known and new theoretical relationships among scores. Section \ref{sec:material} presents the parallel treebanks and annotation styles used to evaluate the models. Section \ref{sec:methods} presents the models that apply the centrality scores in Section \ref{sec:centrality} to guess the root,  the metrics used to evaluate them and further methodological details. Section \ref{sec:results} presents the results of the evaluation of the models and Section \ref{sec:discussion} discusses the implications for the nature of root vertices.

\section{Vertex centrality}

\label{sec:centrality}

\subsection{The standard toolbox}

\label{subsec:centrality_toolbox}

\begin{table}
\caption{\label{tab:subtree_sizes} The sizes of the subtrees that are produced when a vertex from Figure \ref{fig:syntactic_dependency_structures} is removed. }
\begin{tabular}{ll}
\toprule
Vertex & Subtree sizes \\
\midrule
She & 5 \\
wrote & 1, 2, 2 \\
a & 5 \\
book & 1, 4 \\
last & 5 \\
year & 1, 4 \\

\botrule
\end{tabular}
\end{table}

\begin{table}
\caption{\label{tab:centrality_scores_example_sentence}
The outcome of each centrality score on the vertices of the sentence in Figure \ref{fig:syntactic_dependency_structures}. Boldface is used to mark the root of the sentence and the optimal vertices for each centrality score. }
\centering
\begin{tabular}{lllllll}
\toprule
Centrality & She & {\bf wrote} & a & book & last & year \\
\midrule
degree & 1 & {\bf 3} & 1 & 2 & 1 & 2 \\
eccentricity & 3 & {\bf 2} & 4 & 3 & 4 & 3 \\
closeness & 0.53 & {\bf 0.8} & 0.48 & 0.67 & 0.48 & 0.67 \\
max subtree size & 5 & {\bf 2} & 5 & 4 & 5 & 4 \\
subtree size 2nd moment & 25 & {\bf 3} & 25 & 8.5 & 25 & 8.5 \\
betweenness & 0 & {\bf 8} & 0 & 4 & 0 & 4 \\
all-subgraphs & 3.3 & {\bf 4.2} & 3 & 3.8 & 3 & 3.8 \\
$D$ & 1 & {\bf 7} & 1 & 3 & 1 & 5 \\
corrected $D$ & 0.033 & {\bf 1.2} & 0.033 & 0.2 & 0.033 & 0.67 \\
coverage & 1 & {\bf 5} & 1 & 2 & 1 & 4 \\
straightness & 1.4 & 1.8 & 0.73 & 1.2 & 0.93 & {\bf 1.9} \\
 
\botrule
\end{tabular}
\end{table}

Network science provides a large toolbox of scores of the importance (or centrality) of a vertex in a network such as a free tree: degree centrality, PageRank centrality, closeness centrality, betweenness centrality, among others (see \cite{Koschutzki2005a}, \citet[Chapter 7]{Newman2010a} and \citet{Barthelemy2011a} for an overview).
One of the simplest centrality scores is the degree centrality of a vertex, namely the number of links of the vertex. The degree center, also known as hub, is the vertex (or vertices) that maximize the degree. The degree of the vertex ``wrote'' in Figure \ref{fig:syntactic_dependency_structures} (a) is 3. Indeed, ``wrote'' is the only degree center of that sentence (Table \ref{tab:centrality_scores_example_sentence}). 
Hereafter we refer to the degree centrality of vertex $v$ as $k(v)$.
Various more complex scores take into account the shortest paths between two vertices. Suppose that $\delta(u, v)$ is the network shortest path distance in edges between two vertices $u$ and $v$ in the network. 
$\delta(u, v)$ is known as the topological distance between $u$ and $v$.
The mean topological distance of a vertex $v$ to all other vertices is \cite[Chapter 7]{Newman2010a}
\begin{equation}
l(v) = \frac{1}{n - 1} \sum_{u \in V \setminus\{v\}} \delta(u, v),
\label{eq:mean_geodesic_distance}
\end{equation}
where $n$ is the number of vertices of the network. The vertices that minimize $l(v)$ in a graph are known as median vertices \citep{Koschutzki2005a}. The most popular definition of closeness centrality is \citep{Koschutzki2005a, Newman2010a}
\begin{equation}
closeness(v) = \frac{1}{\sum_{u \in V \setminus\{v\}} \delta(u, v)}
\label{eq:popular_closeness}
\end{equation}
and then 
\begin{equation}
closeness(v) = \frac{1}{(n - 1) l(v)}.
\end{equation}
The closeness centrality score of vertex $v$ can also be defined as \cite[Chapter 7]{Newman2010a}
\begin{equation}
closeness(v) = \frac{1}{n - 1} \sum_{u \in V \setminus\{v\}} \frac{1}{\delta(u, v)}.
\label{eq:Newmans_closeness}
\end{equation}
Then $1/closeness(v)$ is the harmonic mean of the network distance of $v$ to all other vertices.   
The betweenness centrality of a vertex $v$ can be defined as \cite[Chapter 7]{Newman2010a} 
\begin{equation}
betweenness(v) = \sum_{s < t} \frac{\sigma_{st}(v)}{\sigma_{st}},
\label{eq:betweenness}
\end{equation}
where $\sigma_{st}(v)$ is the number of shortest paths between vertices $s$ and $t$ passing through vertex $v$ and  
$\sigma_{st}$ is the number of shortest paths between vertices $s$ and $t$, that is
$$\sigma_{st} = \sum_{v} \sigma_{st}(v).$$

Network science also offers specific scores for spatial networks, networks where vertices have coordinates in some space \citep{Barthelemy2011a}. A syntactic dependency structure is a spatial network on the 1-dimensional space defined by the linear arrangement of the words of the sentence. 
Straightness centrality is defined as average of the ratio between the network distance and the physical distance between the vertices \citep{Crucitti2006a}. If $d(u, v)$ is the Euclidean distance between vertices $u$ and $v$, the straightness centrality of $v$ can be defined as \citep{Crucitti2006a} 
\begin{equation}
straightness(v) = \frac{1}{n - 1} \sum_{u \in V \setminus\{v\}} \frac{d(u,v)}{\delta(u, v)}.
\label{eq:straightness}
\end{equation}
In a syntactic dependency structure, the physical distance between vertices is often measured as the Euclidean distance between them (in words units) 
\iftoggle{anonymous}{\citep{Lin1996a,anonymous,Liu2008a}}
{\citep{Lin1996a,Ferrer2004b,Liu2008a}}, 
$$d(u, v) = |\pi(u)- \pi(v)|,$$ 
where $\pi(v)$ is the position of $v$ in the linear arrangement ($1 \leq \pi(v) \leq n$).

Now we turn our attention to the sizes of the connected components that the removal of $v$ produces, namely $n_1,n_2,...,n_k$, where the size of the tree is 
\begin{equation}
n = 1 + \sum_{i=1}^{k} n_i. \label{eq:sum_of_subtree_sizes}
\end{equation}
Each of these components is a subtree of the original tree. 
It is well-known that the definition of the betweenness centrality simplifies for trees,  where $\sigma_{st} = 1$ and then equation \ref{eq:betweenness} becomes \citep{Raghavan_Unnithan2014a,Britz2019a} 
\begin{align}
betweenness(v) & = \sum_{s < t} \sigma_{st}(v) \nonumber \\
               & = \sum_{i < j} n_i n_j. \label{eq:betweenness_trees}
\end{align}

In the centrality scores above, the importance of a vertex is positively or negatively correlated with the value of the score. Within classic graph theory one finds criteria to locate ``central'' vertices that are based on the optimization of some vertex parameter and that have been investigated theoretically in the context of trees \cite[35-36]{Harari1969a}.
\begin{itemize}
\item
In classic graph theory, the center is a vertex that minimizes the eccentricity. The eccentricity of vertex $v$ is 
$$e(v) = \max_u \delta(v, u).$$ 
That center is also known as Jordan center \citep{Jordan1869a}.
To distinguish it from other centers that can be retrieved by optimizing scores other than $e(v)$, we will refer to it as eccentricity center or Jordan center. 
Then an eccentricity center is a vertex $v$ such that the greatest distance $\delta(u,v)$ to other vertices $u$ is minimal.  
In a tree, there can be one or two eccentricity centers. The sentence in Figure \ref{fig:syntactic_dependency_structures} (a) has a single center that is ``wrote'' because that is the vertex that minimizes eccentricity (Table \ref{tab:topological_distances} and Table \ref{tab:centrality_scores_example_sentence}).
\item
The centroid, a vertex that minimizes the maximum size of its subtrees. For a vertex $v$, that maximum size is
$$n_{max}(v) = \max_{1 \leq i \leq k} n_i.$$
Equivalently, a centroid is a vertex such that, when removed, it produces connected components (subtrees) whose number of vertices does not exceed $n/2$, where $n$ is the size of the original tree. 
\footnote{Formally, $v$ is a centroid if $n_i \leq n/2$ for all $i$ such that $1 \leq i \leq k$.}
In a tree, there can be one or two centroidal vertices. Figure \ref{fig:syntactic_dependency_structures} (a) has $n = 6$ and a single centroid that is ``wrote'' because that vertex is the only vertex whose removal produces subtrees of size $\leq n/2 = 3$ (Table \ref{tab:subtree_sizes}). The leaves of a tree with $n > 2$ (the vertices ``She'', ``a'', ``last'' in the example) cannot be centroids. When $n > 2$, the removal of a leaf produces a subtree of size $n - 1$ (and $n - 1 > n/2$ when $n > 2$).     
\end{itemize}
See \citet{Steinbach_1995_centers_and_centroids} for a gentle definition of eccentricity center and centroid and a beautiful gallery of free trees with their eccentricity centers and centroids. \footnote{Freely available from \url{https://oeis.org/A000055/a000055_7.pdf.}}

\begin{table}
\caption{\label{tab:topological_distances} The distance matrix, showing $\delta(v, u)$ for each pair of vertices $v$ and $u$. }
\begin{tabular}{lcccccc}
\toprule
Vertex & She & wrote & a & book & last & year\\
\midrule
She & 0 & 1 & 3 & 2 & 3 & 2 \\
wrote & 1 & 0 & 2 & 1 & 2 & 1 \\
a & 3 & 2 & 0 & 1 & 4 & 3 \\
book & 2 & 1 & 1 & 0 & 3 & 2 \\
last & 3 & 2 & 4 & 3 & 0 & 1 \\
year & 2 & 1 & 3 & 2 & 1 & 0 \\

\botrule
\end{tabular}
\end{table}

\subsection{A hawk-eye view of centrality scores}

The set of possible centrality scores is too wide \citep{Koschutzki2005a, Newman2010a,Barthelemy2011a,Riveros2023a} even for trees \citep{Reid2010a} and thus a research strategy is needed. Furthermore, expanding the set of scores has the increased risk of finding a score that adapts to languages in a sample by chance but it is unlikely that it works well on languages outside that sample. We are interested in a universal notion of root vertex, not a notion that fits specific languages in a typological sense. 
Thus, we identify a core of suitable centrality scores from first principles. 

First, we proceed in a purely mathematical fashion to bring some order to the diversity of centrality scores. A centrality score satisfies the tree rooting property, namely that, if for any free tree, it retrieves just one or two vertices of maximum centrality, and if it retrieves two vertices, these two vertices are connected \citep{Riveros2023a}. If a centrality score satisfies that property, importance decreases from the most central vertices in all directions. Among classic centrality scores, closeness (as defined in Equation \ref{eq:popular_closeness}) and eccentricity satisfy this property, but degree, betweenness and PageRank do not \citep{Riveros2023a}. For this reason, we also consider a recently introduced score, all-subgraphs centrality \citep{Riveros2020a}, that satisfies the tree rooting property \citep{Riveros2023a}. The all-subgraphs centrality of a vertex $v$, is defined as $log_2 A$, where $A$ is the number of connected subgraphs that contain $v$. The calculation of all-subgraphs centrality is a computationally hard problem but for the case of trees, a simple algorithm is forthcoming \citep[Algorithm 2, p. 16]{Riveros2020a}. The tree rooting property and the accompanying theoretical apparatus is an important step towards a taxonomy of the myriad of centrality scores available. However, harnessing the huge diversity of centrality scores \citep{Reid2010a} is beyond the scope of the present article. Here we do not address the theoretical question of whether the tree rooting property is {\em a priori} convenient to find the root of syntactic dependency structures.

Second, we will justify the theoretical importance of the centroid and then we will shift to proposals of new centrality scores (Section \ref{subsec:new_spatial_centrality_scores} and Section \ref{subsec:new_non_spatial_centrality_scores}) and uncovering relationships among scores (Section \ref{subsec:relationships_among_centrality_scores}). 
To set the stage, centroids stem from a notion of centrality that satisfies the tree rooting property: every tree has either one centroid or just two adjacent centroids \citep{Jordan1869a} \footnote{See also \cite[Theorem 4.3, p. 36]{Harari1969a}.} and are equivalent to medians on trees \citep{Kang1975a,Slater1975a}.
Our main point is that centroids are crucial vertices in optimal linear arrangements of trees. An optimal linear arrangement is a total order of the vertices of a free tree that minimizes the sum of distances between linked vertices \citep{Shiloach1979,Chung1984}. In minimum planar linear arrangements, namely minimum linear arrangements such that edges do not cross, the centroid has to be placed in the middle of the linear ordering of the sentence surrounded by subtrees that are sorted by size around the centroid in a specific way
\iftoggle{anonymous}{\citep{Iordanskii1987a, Hochberg2003a,anonymous}.}{\citep{Iordanskii1987a, Hochberg2003a,Alemany2021a}.}
In minimum unconstrained linear arrangements, the centroid is also a key vertex for building an optimal linear arrangement \citep{Shiloach1979, Chung1984}. Thus the saliency of the centroid follows from first principles. 
On the one hand, for its critical role in the theory of optimal linear arrangements of trees \citep{Shiloach1979, Chung1984, Iordanskii1987a, Hochberg2003a}.
On the other hand, for the suitability of that theory for real sentences, where the distance between syntactically related words is smaller than expected by chance \iftoggle{anonymous}{\citep{Liu2008a,Futrell2015a,anonymous}} 
{\citep{Liu2008a,Futrell2015a,Ferrer2020b}} as expected by the principle of syntactic dependency distance minimization \citep{Rijkhoff1986a,Lin1996a,Ferrer2004b}.

\subsection{New spatial scores}
\label{subsec:new_spatial_centrality_scores}

We use the term spatial score to refer to a centrality score that takes into account both the free tree and the linear arrangement of the vertices. 
The challenge is to find a criterion that is valid for any language, and thus no parameter tuning (training) is required in the model that uses that score to guess the root or its ranking. Notice that the preferred placement of a the root may depend on the language: certain languages may have a bias for a late placement of root, that is typically the main verb (as in SOV languages), while other languages may have a bias for a placement of the root in the middle (as in SVO languages) and still some languages may have a bias for an early placement of the verb (as in VSO languages). Therefore, we must reflect on what a root vertex is in an axiomatic sense. One could argue that a root is a vertex that unites distinct components of the sentence and thus it will naturally form long distance dependencies. A prototypical example is the main verb, that unites the major kinds of components of a clause: the subject, the object, the complements and the adjuncts \footnote{\url{https://dictionary.cambridge.org/grammar/british-grammar/adjuncts}}. For that definition, degree centrality would not suffice because, in addition, the root has to link components that are far away in the sentence.  
Accordingly, we put forward the first spatial centrality score: the sum of the edge distances of a vertex. For a vertex $v$, it is defined as
\begin{equation}
D(v) = \sum_{u \in \Gamma(v)} d(u, v),
\label{eq:sum_of_dependency_distances}
\end{equation}
where $\Gamma(v)$ is the set of neighbours of $v$ and $d(u, v)$ is the distance between vertices $u$ and $v$ in the linear arrangement. $D = 1 + 2 + 4 = 7$ for the vertex ``wrote'' in Figure \ref{fig:syntactic_dependency_structures} (a). 
The Euclidean distance center is the vertex (or vertices) that maximize $D(v)$. Indeed, ``wrote'' is the only Euclidean distance center of that sentence (Table \ref{tab:centrality_scores_example_sentence}).
A further justification of the vertex that maximizes $D(v)$ as a likely root from first principles is the structure of optimal projective and planar arrangements. In an optimal projective arrangement, the root has to be surrounded by its optimal projective arrangements following a specific ordering by subtree sizes \citep{Gildea2007a}. If the root is a centroid, then the optimal projective arrangement is also a planar projective arrangement 
\iftoggle{anonymous}{\citep{Iordanskii1987a,Hochberg2003a,anonymous}}
{\citep{Iordanskii1987a,Hochberg2003a,Alemany2021a}}
furthermore, the fact that the root is a centroid warrants that the subtree sizes do not exceed $n/2$, which may increase the chance that the root maximizes $D(v)$.

Guessing that the root is the vertex (or vertices) that maximize $D(v)$ is potentially problematic because one wishes to distinguish the main root from other heads, e.g., the main verb of a subordinate clause or the heads of complex noun phrases. Indeed, it has been shown that dependency distances are naturally maximized (against the principle of syntactic dependency distance minimization) in simple noun phrases \citep{Ferrer2023b} or in short sequences \citep{Ferrer2019a,Ferrer2020b}. Thus we consider an alternative centrality score, that we call coverage, that is simply the distance between the left-most and the right-most vertex among $v$ and its neighbours. The coverage of a vertex $v$ is defined as
\begin{equation*}
C(v) = \max_{u \in \Gamma'(v)} \pi(u) - \min_{u \in \Gamma'(v)} \pi(u),
\end{equation*}
where 
$\Gamma' = \Gamma \cup \{v \}$. Notice that $1\leq C(v) \leq n - 1$.
Finally, we consider a correction of $D(v)$ that takes into account the fraction of the whole linear arrangement covered by $v$ and its neighbours, that is defined as 
\begin{equation*}
D'(v) = \frac{C(v)}{n - 1} D(v).
\end{equation*}
Then $D'(v) \leq D(v)$.

\subsection{Relationships between scores}

\label{subsec:relationships_among_centrality_scores}

\subsubsection{Hard versus soft centrality scores}

\label{subsubsec:soft_versus_hard_centrality_scores}

We say that a centrality score is a hard score if it is an optimum of a certain sample of values and thus retains just one value in the sample; we say that a centrality score is soft if it aggregates (e.g., averages) the values in a sample. Eccentricity is a hard score, i.e. the maximum topological distance of a vertex to the remainder of the vertices. Eccentricity has soft correlates that are averages: the closeness centrality in Equation \ref{eq:popular_closeness}, that is proportional to the inverse of the arithmetic mean, as well as the closeness centrality in Equation \ref{eq:Newmans_closeness}, that is the inverse of the harmonic mean.

This "soft/hard" terminology is not standard in statistics but is borrowed by analogy from machine learning. In that field, a "soft prediction" refers to a continuous score (e.g., a probability) that retains model confidence, while a "hard prediction" is the discrete, categorical outcome (e.g., a class label) derived from it. The analogy holds because "hard" metrics, much like hard predictions, are extreme aggregations that discard information present in their "soft" counterparts.

Let us consider centroids, the vertices that are retrieved by a hard score, minimizing the maximum subtree size produced by their removal. At first glance, our centrality score toolbox seems to lack a soft correlate, in the form of a simple aggregation of the values of these subtree sizes. For trees, betweenness centrality is usually defined in terms of a sum of pairwise products of subtree sizes (equation \ref{eq:betweenness_trees}) \citep{Raghavan_Unnithan2014a,Britz2019a}. However, the following property shows that betweenness reduces to a sum of squared subtree sizes.
\begin{property}
If a vertex $v$ has $k$ neighbours, then
$$betweenness(v) = \frac{1}{2}\left[(n - 1)^2 - \sum_{i=1}^k n_i^2 \right].$$
\label{prop:betweenness_versus_subtree_sizes_2nd_moment}
\end{property}
\begin{proof}
Equation \ref{eq:betweenness_trees} can be expressed equivalently 
\begin{align*}
betweenness(v) & = \frac{1}{2} \sum_{i=1}^k n_i \left(\sum_{j=1}^k n_j - n_i\right) \\
               & = \frac{1}{2} \left[\left(\sum_{i=1}^k n_i \right)^2 - \sum_{i=1}^k n_i^2\right]. \\
\end{align*}
Then the application of equation \ref{eq:sum_of_subtree_sizes} produces the desired result. 
\end{proof}

Thus, betweenness centrality is indeed a straightforward soft correlate of the maximum subtree size.
The following property shows that the mean ($m$) and the variance ($V$) of the subtree sizes produced by the removal of a vertex $v$ have simple expressions. 

\begin{property}
If a vertex $v$ has $k$ neighbours, then
\begin{align*}
m(v) & = \frac{n - 1}{k} \\
V(v) & = \frac{1}{k}\left(\sum_{i=1}^k n_i^2 - \frac{(n-1)^2}{k}\right).
\end{align*}
\end{property}
\begin{proof}
First, 
\begin{equation}
m(v) = \frac{1}{k} \sum_{i=1}^k n_i = \frac{n - 1}{k}
\label{eq:mean_of_subtree_sizes}
\end{equation}
thanks to equation \ref{eq:sum_of_subtree_sizes}.
Then the substitution of $m(v)$ in the definition
\begin{equation*}
V(v) = \frac{1}{k} \sum_{i=1}^k n_i^2 - m(v)^2
\end{equation*}
yields the final expression for $V(v)$.
\end{proof}

\subsubsection{A new soft score}

\label{subsec:new_non_spatial_centrality_scores}

Given the form of betweenness in Property  \ref{prop:betweenness_versus_subtree_sizes_2nd_moment}, we also consider the 2nd moment of subtree sizes as an alternative soft score for max subtree size. 
The mean of the subtree sizes is the 1st moment about zero of the subtree sizes, i.e. $m_1(v) = m(v)$ (Equation \ref{eq:mean_of_subtree_sizes}) while their 2nd moment about zero is  
\begin{equation}
m_2(v) = \frac{1}{k}\sum_{i=1}^k n_i^2
\label{eq:subtree_sizes_2nd_moment}
\end{equation}
and then the variance of the subtree sizes is 
$$Var(v) = m_2(v) - m_1(v)^2.$$ 
Thus, the betweenness becomes 
$$betweenness(v) = \frac{1}{2}[(n - 1)^2 - k m_2(v)].$$

\subsubsection{Local versus global centrality scores}

Another important distinction, beyond soft and hard scores, involves whether centrality measures use local (neighbor-based) or global (whole network) information.
Local scores, like degree centrality, consider only neighboring nodes, while global scores leverage data from across the entire tree. The local scores are degree-centrality and
the new spatial scores. The remaining scores exploit global information about the free tree (e.g. the shortest path distances in the tree).
{\em Ceteris paribus,} global scores are expected to perform better than local scores because they leverage more comprehensive information about the tree.

\subsubsection{Degree centrality versus other scores}
\label{subsubsec:degree_versus_new_spatial_scores}

Degree centrality ($k(v)$) is a local non-spatial score.
Due to the simplicity of its definition, degree centrality serves as a control or baseline for other scores. Therefore we investigate some relationships between degree and other scores.

The following property shows lower and upper bounds for betweenness uncovering a dependency with degree centrality.
\begin{property}
\begin{equation*}
\frac{(n-1)^2}{2} \left(1 - \frac{1}{k} \right) \leq betweenness(v) \leq {n - 1 \choose 2} \\
\end{equation*}
\end{property}
\begin{proof}
First, we will derive lower and upper bounds for $\sum_{i=1}^k n_i^2$. 
On the one hand, 
\begin{equation*}
\sum_{i=1}^k n_i^2 \geq \sum_{i=1}^k n_i = n - 1
\end{equation*}
thanks to equation \ref{eq:sum_of_subtree_sizes}.
On the other hand, $V(v) \geq 0$ yields
\begin{equation}
\sum_{i=1}^k n_i^2 \leq \frac{1}{k}(n-1)^2.
\end{equation}
Then the lower and upper bounds of betweenness follow after plugging the lower and upper bounds of $\sum_{i=1}^k n_i^2$ into the simple definition of betweenness in Property \ref{prop:betweenness_versus_subtree_sizes_2nd_moment}.
\end{proof}

Degree centrality is an obvious control for other scores based on topological neighbours of a vertex. The following property shows a relationship with Euclidean distance centrality ($D(v)$).
\begin{property}
Suppose a syntactic dependency structure of $n$ vertices ($n$ words). In a random linear arrangement (namely a random shuffling of the words of the sentence), the expected value of $D(v)$ is 
\begin{equation}
\E[D(v)] = k(v) \frac{n + 1}{3}.
\label{eq:expected_sum_of_dependency_distances}
\end{equation}
\item
$D(v)$ is bounded below and above by a quadratic function of $k(v)$, namely,
\begin{equation*}
\left\lfloor \frac{1}{4}(k(v) + 1)^2 \right\rfloor \leq D(v) \leq \frac{1}{2}k(v)(2n - 1 - k(v)).
\end{equation*}
\end{property}
\begin{proof}
The expected value of $D(v)$ in a random linear arrangement is  
\begin{equation*}
\E[D(v)] = \sum_{\{u,v\} \in E} \E[d(u, v)]
\end{equation*}
thanks to Equation \ref{eq:sum_of_dependency_distances} and the linearity of expectation.
Knowing that \citep{Ferrer2004b} \footnote{\iftoggle{anonymous}{See \citet{anonymous} for a more detailed argument.}{See \citet[Section 2.2]{Alemany2021b} for a detailed derivation.}}
$$\E[d(u, v)] = \frac{n+1}{3},$$
we finally obtain Equation \ref{eq:expected_sum_of_dependency_distances}.
As for the range of variation of $D(v)$, notice $v$ and its neighbours in the free tree form a star tree of $n = k(v) + 1$ vertices. On the one hand, $D(v)$ is minimized by a minimum linear arrangement of such star tree. Recalling that the minimum sum of edge distances of a tree of $n$ vertices is \citep{Iordanskii1974a}
\begin{equation*}
\left\lfloor \frac{1}{4}n^2 \right\rfloor,
\end{equation*}
it follows that $D(v)$ is bounded below by
\begin{equation*}
\left\lfloor \frac{1}{4}(k(v) + 1)^2 \right\rfloor.
\end{equation*}
Second, $D(v)$ is maximized by placing $v$ at one end of the linear arrangement and its neighbours at the other end, which yields the following upper bound of $D(v)$
\begin{equation*}
\sum_{i = 1}^{k(v)} (n - i) = \frac{1}{2}k(v)(2n - 1 - k(v)).
\end{equation*}
\end{proof}
The first relationship (Equation \ref{eq:expected_sum_of_dependency_distances}) indicates that degree centrality would be equivalent to the Euclidean distance centrality if the order of a sentence was arbitrary. However, it is well-known that dependency distances only achieve lengths that are neither shorter nor longer than expected by chance in short sentences \citep{Ferrer2019a} or exceptionally in languages depending on the annotation style \citep{Ferrer2020b}. \footnote{The exceptions were Telugu and Warlpiri when using UD annotation style; no exception when SUD annotation style was used \citep{Ferrer2020b}.}

\subsubsection{Consistency among centrality scores}
\label{subsec:consistency}

\begin{figure}
\includegraphics[width = \textwidth]{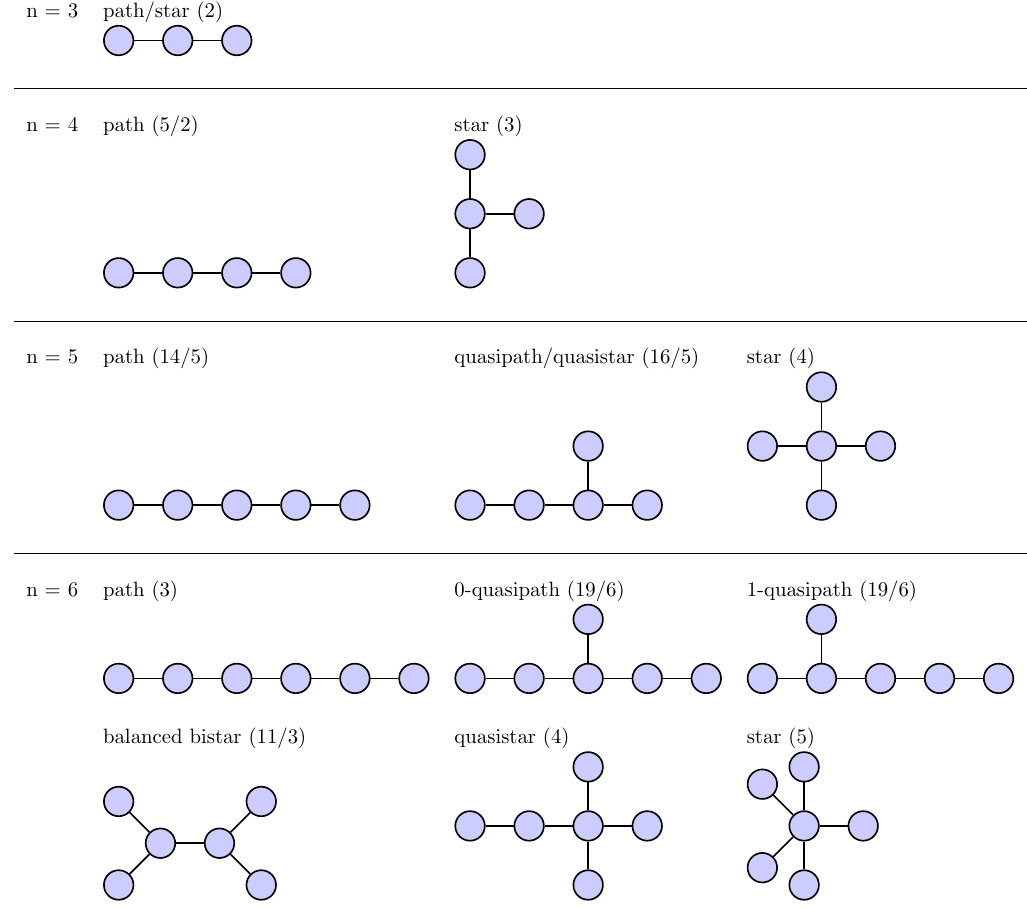}
\caption{\label{fig:all_unlabelled_trees} All unlabelled trees between 3 and 6 vertices and their canonical names. The trees of the same size are sorted by increasing degree variance. The number in parenthesis next to the tree name is $\left< k^2 \right>$, the 2nd moment of degree about zero (equation \ref{eq:degree_2nd_moment_zero}). } 
\end{figure}

An important result is that, in a tree, the centroid vertices are the same as the median vertices \citep{Kang1975a,Slater1975a}, that is, the vertices that minimize the maximum subtree size (centroids) coincide with those that maximize the popular closeness (Equation \ref{eq:popular_closeness}) (or minimize the mean topological distance; equation  \ref{eq:mean_geodesic_distance}). \footnote{See \cite{Koschutzki2005a} for an updated overview.}
Therefore, the maximum subtree size, the mean topological distance and the popular closeness are consistent given any tree. Next we will consider a more restrictive notion of consistency.

Given an unlabelled tree, we say that two or more centrality scores are consistent if they give the same center (or centers).
We investigate the consistency of the centrality scores on the following kinds of trees (Figure \ref{fig:all_unlabelled_trees})
\begin{enumerate}
\item 
Star tree. A tree with a vertex of maximum degree. That vertex is called the hub of the star tree. 
A star tree of $n$ vertices has a hub of degree $n-1$.
\item 
Quasistar tree. A quasistar tree of $n$ vertices is formed by attaching a vertex to one of the leaves of a star tree of $n-1$ vertices.
\item 
Path tree. A tree where the maximum degree is two. A path tree has two leaves (that have degree 1) and $n-2$ internal vertices (that have degree 2). A path tree has one middle vertex (when $n$ is odd) or two middle vertices (when $n$ is even). 
\item 
Quasipath tree. A quasipath tree of $n$ vertices is formed by attaching a vertex to one of the internal vertices of a path tree of $n-1$ vertices.  
\item
$d$-quasipath tree. A $d$-quasipath tree of $n$ vertices is a quasipath tree that is formed by (a) taking a path tree of $n-1$ vertices (b) selecting an internal vertex of the path tree that is at distance $d$ of the middle vertex or vertices, where $0 \leq d \leq \frac{n}{2} - 1$ and (c) attaching a leave to that internal vertex. This definition of quasipath tree requires $n \geq 4$ because the existence of a path tree with at least one internal vertex requires $n = 3$. 
\item
Balanced bistar tree. A bistar tree is obtained by linking the respective hubs of two star trees. These two hubs are the hubs of the bistar tree. A balanced bistar tree of $n$ vertices is formed by two stars of size $\lfloor n/2 \rfloor -1$ and $\lceil n/2 \rceil -1$.
\end{enumerate}
The names of these trees are borrowed from \cite{Ferrer2020a}. Quasipath tree is a name we introduce in this article. 
Figure \ref{fig:all_unlabelled_trees} shows all the unlabelled free trees up to 6 vertices with their names (notice that the same tree may receive different names according to the definitions above; the figure shows the canonical name we use in this article). \footnote{All the possible unlabelled trees up to $n = 10$ can be seen in Harari's classic graph theory book \citep{Harari1969a}.} The free tree in Figure \ref{fig:syntactic_dependency_structures} is a 0-quasipath of 6 vertices.

It is convenient to sort the trees of same size by their degree variance as in Figure \ref{fig:all_unlabelled_trees}. Over trees of same size, the average degree, $2 - 2/n$, is constant and then the degree variance  is determined by the 2nd moment of degree about zero, i.e. 
\begin{equation}
\left< k^2 \right> = \frac{1}{n} \sum_{i=1}^n k_i^2,
\label{eq:degree_2nd_moment_zero}
\end{equation}
where $k_i$ is the degree of the $i$-th vertex. Thus degree variance reduces to $\left< k^2 \right>$ in trees of same size. $\left< k^2 \right>$ is minimized by path trees and maximized by star trees \citep{Ferrer2013b}. $\left< k^2 \right>$ is a measure of hubiness or star-likeness, namely a measure of the similarity with respect to a star tree \citep{Ferrer2017a}.  

All the centrality scores on the free tree that are used in this article satisfy the following consistency properties.
\begin{property}
\label{prop:consistency}
The consistency among the non-spatial scores by tree kind \citep{Ferrer2020a}, in order of increasing hubiness, is as follows 
\begin{enumerate}
\item 
Path tree. All scores except vertex degree are consistent on linear trees. The degree centrality finds $\max(n - 2, n)$ centers (all vertices if $n < 3$ or the $n - 2$ vertices of degree two if $n \geq 3$) whereas the remainder of the centrality scores find the middle vertices of the path.
\item 
Balanced bistar tree (with $n > 3$; when $n \leq 3$ the tree becomes a star tree). All centrality scores are consistent on balanced bistar trees when $n$ is even; when $n$ is odd all centrality scores except eccentricity are consistent. When $n$ is even, all the centrality scores agree that the two hubs of that tree are the center; when $n$ is odd, all centrality scores agree that the hub with highest degree is the center except eccentricity, which determines that the two hubs are indeed the centers.   
\item 
Quasistar tree with $n > 4$ (when $n \leq 4$ the quasistar is also a star). All scores except eccentricity are consistent on quasistar trees. 
Eccentricity finds that the centers are the hub and the vertex of degree $2$ whereas the remainder of centrality scores agree that the hub (the vertex of degree $n - 2$) is the only center.  
\item
Star tree. The scores are consistent for star trees, that is, for any star tree, all centrality scores give the same center, that is the hub vertex of the star.
\end{enumerate}
\end{property}
\begin{proof}
We have verified computationally the consistency properties above. \footnote{Indeed, we have checked that this is true for $n \leq 10^3$. That is more than enough for syntactic dependency structures. A rigorous mathematical proof is a tedious exercise.}
\end{proof}

\begin{figure}
\includegraphics[width = 0.92\textwidth]{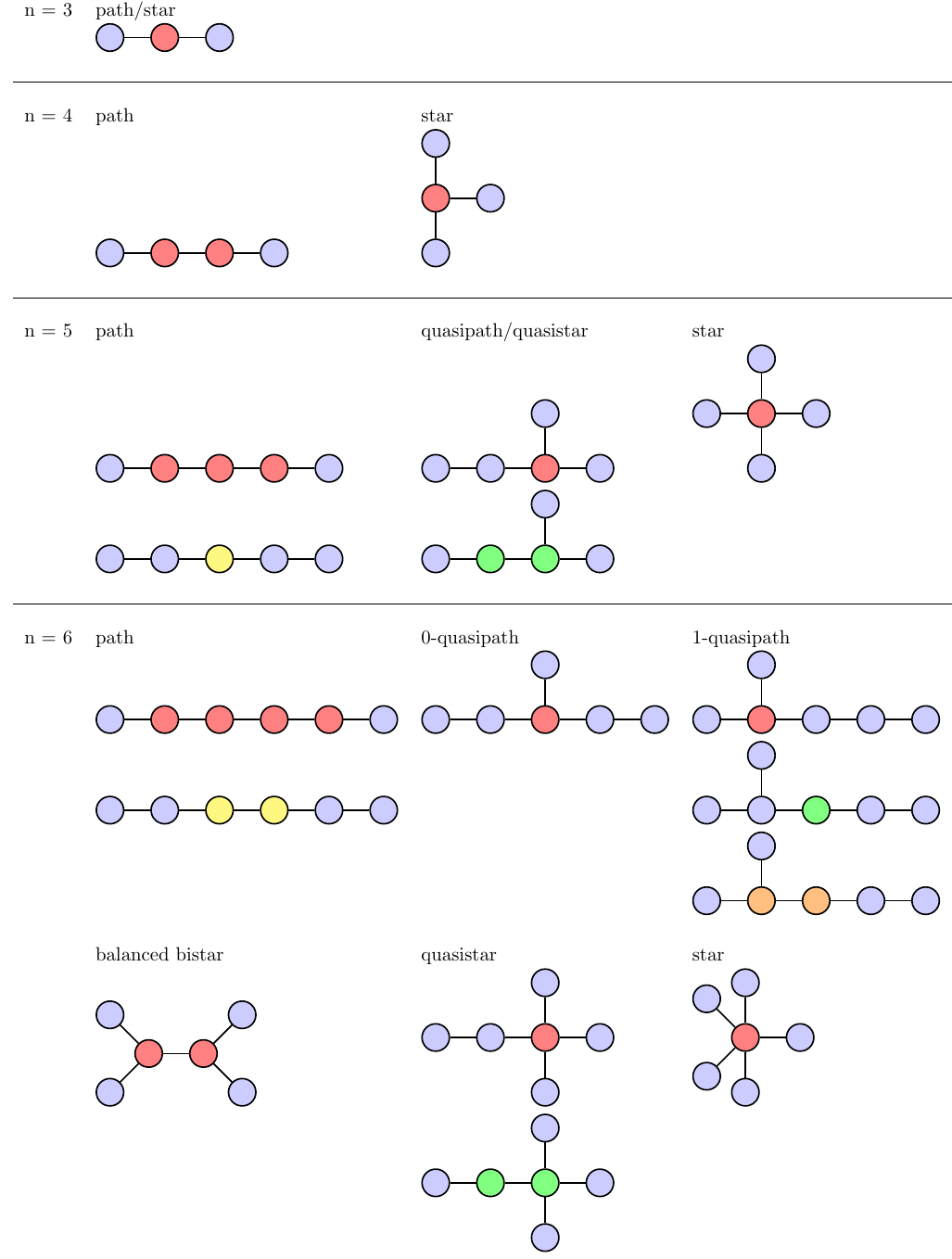} 
\caption{\label{fig:all_unlabelled_trees_with_centers} 
All unlabelled trees between 3 and 6 vertices, their canonical names and the centers retrieved by the centrality scores on the free tree (the non-spatial scores). 
The trees of same size are sorted by increasing degree variance.
Centers are colored according to the representative of the class that retrieves them: degree centrality (red), eccentricity (green and yellow; green when eccentricity is the only member of the class) and maximum subtree size (orange). 
For each tree, we show only a representative of the class of equivalence that results from conditioning on both the tree kind and its size.
}
\end{figure}

Property \ref{prop:consistency} has two important consequences described in the following corollaries and illustrated in Figure \ref{fig:all_unlabelled_trees_with_centers}. 
\begin{corollary}
\label{cor:consistency_up_to_n4}
~\\
\begin{enumerate}
\item
All the non-spatial centrality scores on the free trees used in this article are consistent if $1 \leq n \leq 4$, given any tree with $1 \leq n \leq 4$ all produce the same centers because within that range of $n$, the trees are either star trees or linear trees ($n=4$) or both ($n < 4$).
\item 
Given only a free tree with $1 \leq n \leq 4$, all centrality scores will agree that the node(s) with highest degree must be the roots. 
\end{enumerate}
\end{corollary}
\begin{proof}
Trivial given Property \ref{prop:consistency}.
\end{proof}

\begin{corollary}
\label{cor:classes_of_equivalence_on_path_quasistar_star}
The consistency among centrality scores defines classes of equivalence among non-spatial scores on the free tree given a kind of tree (in order of increasing hubiness):
\begin{itemize}
\item 
Path tree. There are two classes, one that contains degree alone and another one, represented by eccentricity, that covers the remainder of the scores. 
\item
Balanced bistar tree (with $n > 3$). When $n$ is even, there is only one class, represented by degree centrality. When $n$ is odd, there are two classes, one that contains eccentricity alone and another one, represented by degree, that covers the remainder of the scores. 
\item 
Quasistar tree. There are two classes, one that contains eccentricity alone and another one, represented by degree centrality, that covers the remainder of the scores. 
\item
Star tree. There is only one class that is represented by degree centrality.
\end{itemize}
\end{corollary}
\begin{proof}
Trivial given Property \ref{prop:consistency}.
\end{proof}

As for spatial scores, the analysis of consistency among centrality scores is beyond the scope of the present article but some simple results on star trees are worth mentioning and help simplify the reporting of results.
\begin{property}
\label{prop:new_spatial_scores_on_star}
On star trees, 
\begin{enumerate}
\item
$D(v)$ and $D'(v)$ are consistent, namely both centrality scores retrieve the same center that is the hub.
\item
$C(v)$ and straightness are not consistent with $D(v)$, $D'(v)$ and any other non-spatial score. $C(v)$ and straightness can retrieve more than one center. 
\end{enumerate}
\end{property}
\begin{proof}
We use $leaf$ and $hub$ to refer, respectively, to a leaf and the hub of a star tree. 
\begin{enumerate}
\item
By definition of $D(v)$ and $C(v)$, it is easy to see that $D(leaf) < D(hub)$ and that $C(leaf) \leq C(hub) = n - 1$.
By integrating the previous properties into the definition of $D'(v)$, one obtains $D'(hub) = D(hub)$ and also $D'(leaf) \leq D(leaf) < D(hub) = D'(hub)$. Thus the hub is always the center retrieved by both $D(v)$ and $D'(v)$.
\item
On a star tree, all non-spatial scores retrieve the hub as the only center (Property \ref{prop:consistency}). It is easy to see that $C(v)$ and straightness can retrieve more than one vertex and thus they are inconsistent with any other non-spatial score. When $n \geq 3$ and the hub of the star tree is placed at one of the ends of the linear arrangement, the coverage of both the hub and the leaf at the other end coincides, giving two centers instead of one. To see that straightness centrality can retrieve multiple centers, consider $n=3$ and that the hub is placed at the center of the linear arrangement. Then all vertices are centers because they all yield the same centrality: the straightness centrality of the hub is (Equation \ref{eq:straightness})
$$\frac{1}{2}\left(\frac{1}{1} + \frac{1}{1}\right) = 1$$
while the straightness centrality of the leaves is
$$\frac{1}{2}\left(\frac{1}{1} + \frac{2}{2}\right) = 1.$$
\end{enumerate}
\end{proof}
Thus, $D(v)$ and $D'(v)$ find the same center (the hub) as the scores on the free tree independently of the linear order of the words of the sentences. 

\subsubsection{Short sequences}

\label{subsec:small_sequences}

We aim to investigate the performance of the centrality scores in short sequences. The reason is four-fold. First, the placement of the head is more predictable in short sequences. In particular, it has been shown that the hub of star trees is more likely to be placed at one of the ends (either first or last) in small sentences \citep{Ferrer2019a,Ferrer2020b} or in short noun-phrases \citep{Ferrer2023b}. Second, 
to control for the kind of tree in a setting where the number of distinct unlabelled trees is small. Third, to take advantage of the classes of equivalence among centrality scores (Section \ref{subsec:consistency}). Fourth, to set some foundations for research (a) on simple sentences or on languages where subordination is lacking or debated \citep{Pullum2024a} and also (b) on the short sequences that other species produce \citep{Sigmundson2025a,Ferrer2012c}.

For the sake of simplicity, we restrict the analysis of short sequences to sentences with $3 \leq n \leq 6$ (Figure \ref{fig:all_unlabelled_trees}). That range comprises the average sentence length in a Pirahã corpus, that is 5.9 words \citep{Futrell2016a} and the average length of dolphin whistle sequences, that does not exceed 6 across individuals \citep[Table 1]{Ferrer2012c}.
We need to identify all classes of equivalence for any free tree that result from conditioning on $n$ for $n$ within that range. The next property deals with the case of $n=6$.

\begin{property}
\label{prop:classes_given_tree_kind_and_n6}
The classes of equivalence of non-spatial scores that result from conditioning both on tree kind and $n = 6$ are as follows (in order of increasing hubiness of the tree)
\begin{itemize}
\item
Path tree. Degree centrality (only member of its class) and eccentricity (representative for the class formed by the remainder of scores).
\item 
0-quasipath tree. Degree centrality (representative of the single class formed by all scores). 
\item
1-quasipath tree. Eccentricity as only member of the class, maximum subtree size (centroid) as only member of the class and degree as representative of the class that contains the remainder of the scores. Degree centrality finds a single center that is the hub (the vertex of degree three); eccentricity finds a single center, that is the internal vertex that is the closest to the hub; finally, the maximum subtree size finds the union of the centers found by degree and eccentricity, namely the hub and the eccentricity center (Figure \ref{fig:all_unlabelled_trees}). 
\item
Balanced bistar tree. Degree centrality (representative of the only class). 
\item
Quasistar. Degree centrality (the class of all scores except eccentricity) and eccentricity (the only member of its class). 
\item
Star. Degree centrality (only member of its class).
\end{itemize}
\end{property}
\begin{proof}
We examine all the possible kinds of tree when $n = 6$
\begin{itemize}
\item
Path tree. The classes of equivalence for path trees in Property \ref{cor:classes_of_equivalence_on_path_quasistar_star} are not altered by the fact of conditioning on $n = 6$.
\item 
0-quasipath tree. We already know that there is only one class (Figure \ref{fig:syntactic_dependency_structures} (a); Table \ref{tab:centrality_scores_example_sentence}) and then we use degree centrality as representative. 
\item
1-quasipath tree. The 1-quasipath tree of 6 vertices is the smallest tree with two centroids \cite[Figure 4.4, p. 36]{Harari1969a}.
Computing the center according to each centrality scores as we did for the 0-quasipath tree of 6 vertices, we find that there are three classes of equivalence: one class the contains eccentricity alone, another class that contains maximum subtree size (centroid), and another class that contains the remainder of the scores, that is represented by degree. More precisely, degree centrality finds a single center that is the hub (the vertex of degree three); eccentricity finds a single center, that is the internal vertex that is the closest to the hub; finally, the maximum subtree size finds the union of the centers found by degree and eccentricity, namely the hub and the Jordan center (Figure \ref{fig:all_unlabelled_trees}). 
\item
Balanced bistar tree. As $n$ is even, there is only one class for balanced bistar trees (Corollary \ref{cor:classes_of_equivalence_on_path_quasistar_star}), that is represented by degree centrality. 
\item
Quasistar. The classes of equivalence for quasitar trees in Property \ref{cor:classes_of_equivalence_on_path_quasistar_star} are not altered by conditioning on $n = 6$.
\item
Star. Trivial as there is only one class of equivalence for star trees before conditioning on $n = 6$ (Property \ref{cor:classes_of_equivalence_on_path_quasistar_star}).
\end{itemize}
\end{proof}

The next property presents the centers that are retrieved by each class of non-spatial centrality score when both the tree kind and its size are given. 
\begin{property}
\label{prop:classes_given_tree_kind_and_tree_size}
For $3 \leq n \leq 6$, the classes of equivalence conditioning both on tree kind and tree size $n$ are the ones shown in Figure \ref{fig:all_unlabelled_trees_with_centers}. 
\end{property}
\begin{proof}
When $n = 6$, the classes of equivalence are borrowed from Property \ref{prop:classes_given_tree_kind_and_n6}. 
When $n < 6$, all trees are either star, quasistar or path (Figure \ref{fig:all_unlabelled_trees}) and the classes of equivalence for each tree kind given $n$ are the same as when not conditioning on tree size (Corollary \ref{cor:classes_of_equivalence_on_path_quasistar_star}), except for path trees with $n=4$.  
For path trees there are two classes of equivalence when not conditioning on tree size are represented by degree and eccentricity. However, these two representatives retrieve the same vertices of a path tree with $n=4$ and thus there is just a single class of equivalence that we represent by degree centrality.
\end{proof}

We combine the representatives of each class for all trees of a given size $n$ so as to form a minimal set of representatives of centrality scores on the free tree that are strictly necessary given the tree size, which yields the following minimal sets of representatives (Figure \ref{fig:all_unlabelled_trees_with_centers}).
\begin{property}
\label{prop:classes_given_tree_size}
When only the tree size is given, the minimal set of representatives of non-spatial centrality scores that are strictly necessary to cover any score in our ensemble of non-spatial scores are
\begin{itemize}
\item
$n = 3$ or $n = 4$. Degree centrality.
\item
$n = 5$. Degree and eccentricity.
\item
$n = 6$. Degree, eccentricity and maximum subtree size.
\end{itemize} 
\end{property}
\begin{proof}
We examine each tree size ($n$) as follows
\begin{itemize}
\item 
$n = 3$ (the tree is both a star tree and a linear tree). Just degree centrality according to Corollary \ref{cor:classes_of_equivalence_on_path_quasistar_star}.
\item 
$n = 4$ (the trees are path or star trees). Just degree centrality because degree, the representative for path tree and star tree, and eccentricity, the representative for path trees (Corollary \ref{cor:classes_of_equivalence_on_path_quasistar_star}) retrieves the same vertices for $n = 4$ (Figure \ref{fig:all_unlabelled_trees_with_centers}).
\item 
$n = 5$ (the trees are path, quasistar or star trees). Degree and eccentricity (Corollary \ref{cor:classes_of_equivalence_on_path_quasistar_star}) because one cannot replace the other.
\item 
$n = 6$ (the trees are path, 0-quasipath, 1-quasipath, balanced bistar, quasistar or star trees). Degree, eccentricity and maximum subtree size because none of them can replace another score in the triple.
\end{itemize}
\end{proof}
Given all the results so far, we will show the strictly necessary centrality scores when reporting on the performance of the scores in Section \ref{sec:results} and in the Appendix.

\subsection{Summary of scores}

\begin{sidewaystable}
\caption{\label{tab:features_of_centrality_scores}
Summary of the features of each centrality score (``Centrality''). ``Spatial'' indicates if the score is spatial or not. 
``Central vertex'' is the name of the vertex or vertices that optimize the centrality score.
``Abbreviation'' is the abbreviated name of the scores used for tables and figures.
``Information'' is the information that the scores take as input from the free tree. 
``Optimum'' is the kind of optimization of centrality value required to produce a center (``max'' for maximization of centrality value and ``min'' for minimization).
``Local'' indicates if the centrality score exploits local information.
``Can root'' indicates if the centrality score has the tree rooting property \citep{Riveros2023a}. 
``Hardness'' indicates if the score is hard or soft (in case of a hard score, the abbreviated names of its soft scores are indicated in parenthesis).
}
\centering
{\footnotesize 
\begin{tabular}{lllllllllll}
\toprule
Spatial & Centrality & Central vertex & Abbreviation & Information & Optimum & Local & Can root & Hardness\\
\midrule
no & degree & hub & $k$ & degree & max & yes & no & soft & \\
no & eccentricity & Jordan center & eccentricity & topological distance & min & yes & hard ($l$, closeness)\\
no & popular closeness (eq. \ref{eq:popular_closeness}) & median & $l$ & topological distance & max & no & yes & soft \\
no & Newman's closeness (eq. \ref{eq:Newmans_closeness}) & - & closeness & topological distance & max &  no & no\footnotemark[1] & soft \\
no & max subtree size & centroid & $n_{max}$ & subtree size & min & no & no & hard ($m_2$, betweenness) \\
no & subtree size 2nd moment & - & $m_2$ & subtree size & min & no & no\footnotemark[1] & soft \\
no & betweenness & - & betweenness & subtree size & max & no & no & soft \\
no & all-subgraphs & - & all-subg & containing subgraphs & max & no & yes & soft \\
yes & $D$ & - & $D$ & neighbours & max & yes & - & - \\
yes & corrected $D$ & - & $D'$ & Euclidean distance & max & yes & - & - \\
yes & coverage & - & $C$ & Euclidean distance & max & yes & - & - \\
yes & straightness & - & straightness & topological + & max & no & - & - \\
    &              &   &              & Euclidean distance \\

\botrule
\end{tabular}
}
\footnotetext[1]{This score does not satisfy the tree rooting property because it does not retrieve either two disconnected centers or more than two centers on the dataset in Section \ref{sec:material}.}
\end{sidewaystable}

The spatial scores in our study comprise straightness centrality and all the Euclidean distance centrality scores. Non-spatial scores are scores that only take into account the structure of the free tree.
Table \ref{tab:features_of_centrality_scores} summarizes the main features of the centrality scores used in this study and Table \ref{tab:centrality_scores_example_sentence}
shows their value for each vertex in the example sentence (Figure \ref{fig:syntactic_dependency_structures}). Straightness centrality is the only score that fails to identify the root, although the root has the second largest centrality value. For simplicity, we exclude the popular definition of closeness (Equation \ref{eq:popular_closeness}) from our statistical analyses because it retrieves the same centers as max subtree size (centroid) \citep{Kang1975a,Slater1975a}and exhibits the tree rooting property \citep{Riveros2023a}. However, we include Newman's closeness (equation \ref{eq:Newmans_closeness}) because it does not exhibit the tree rooting property. 

From a methodological standpoint, the rationale behind the set of centrality scores used in this article is as follows. Concerning established scores, it covers the typical scores considered in the literature \citep{Barthelemy2011a,Crucitti2006a}. The non-spatial scores give a reference point to spatial scores. We wish to know how powerful a non-spatial centrality score can be. 
Some centrality scores are justified or designed by first principles (centroid, all-subgraphs centrality and the new spatial scores) while others are just included for being representative of the network science or graph theory toolbox (the remainder). Given their simplicity, some scores serve as reference for others. In particular, degree centrality serves as a control for the new spatial scores, which in turn yield a simple reference for non-spatial scores and also provide a baseline to complex spatial scores such as straightness centrality.

\section{Material}

\label{sec:material}

The source data is the Parallel Universal Dependencies (PUD) collection \citep{conll2017st_APS}. PUD consists of a series of sentences and their syntactic dependency annotation from 21 languages belonging to 9 linguistic families (Table \ref{tab:languages}).
That collection is chosen to control for the content or the source text of the treebanks. In particular, we borrow PUD from the 2.14 release of the Universal Dependencies treebank collection. \iftoggle{anonymous}{\footnote{In previous quantitative dependency syntax research, the 2.6 release of PUD was used  \citep{anonymous}.}}{\footnote{In previous quantitative dependency syntax research, the 2.6 release of PUD was used  \citep{Ferrer2021a,Ferrer2020b}.}}

\begin{table}
\caption{\label{tab:languages} The 21 languages in the PUD collection grouped by linguistic family. For each language, we also indicate the dominant order of  subject (S), verb (V) and direct object (O) according to WALS \citep{wals}.}
\begin{tabularx}{\textwidth}{lXX}
Family & Languages & Dominant order \\
\hline
Afro-Asiatic & Arabic & VSO \\
Austronesian & Indonesian & SVO \\
Koreanic & Korean & SOV \\
Indo-European & Czech, English, French, Galician, German, Hindi, Icelandic, Italian, Polish, Portuguese, Russian, Spanish, Swedish & SVO for all languages except German (SOV or SVO) and Hindi (SOV). \\
Japonic & Japanese & SOV \\
Sino-Tibetan & Chinese & SVO \\ 
Tai-Kadai & Thai & SVO \\
Turkic & Turkish & SOV \\
Uralic & Finnish & SVO \\
\end{tabularx}
\end{table}

By default, the PUD treebank collection follows the UD annotation style \citep{ud26_APS}. To control for annotation style, we also use the SUD annotation style \citep{sud}. SUD stands for Surface-Syntactic Universal
Dependencies. 
The preprocessing method is borrowed from a recent study \citep{Ferrer2020b} and involves the removal
of punctuation marks and reparalellization to warrant there is no loss of parallelism after punctuation mark removal and setting the minimum sentence length to $n = 3$. As a result the reparallelization process, all languages end up having $N_S = 995$ sentences.

\section{Methods}

\label{sec:methods}
  
\subsection{Evaluation}

\label{subsec:evaluation}

\subsubsection{Ranking}

We evaluate the centrality scores by their capacity to rank the root vertex near the top.
Depending on the score, the centrality score will be minimized or maximized (Table \ref{tab:features_of_centrality_scores}).
Suppose a centrality score that is to be maximized to find the root. Then we sort all vertices decreasingly by centrality. An ideal score would leave the root vertex in the first position of the ranking. In practice, that may not happen and the centrality score may produce the same value for distinct vertices. For this reason our first evaluation metric is the rank as defined in non-parametric statistics, that is, if the there is a maximal sequence of tied vertices starting in position $i$ and ending in position $j$ of the order, all these vertices get a rank that is the average position of the vertices \citep{Conover1999a}, i.e. 
\begin{align}
  r = r(i, j) & = \frac{1}{j-i+1} \sum_{k=i}^j k \nonumber \\  
          & = \frac{1}{j-i+1} (i + j)(j - i + 1)/2 \nonumber \\
          & = \frac{i + j}{2}.\label{eq:average_rank}
\end{align}
If the centrality score is such that it has to be minimized to find the root (e.g., eccentricity, maximum subtree size) the procedure is the same but vertices are sorted increasingly by centrality.

As ranks from sentences of different length are not comparable, we transform all ranks into numbers between 0 and 1 knowing that $1 \leq r \leq n$. The normalized rank is 
\begin{align*}
\overline{r} & = \frac{r - 1}{n - 1}.
\end{align*}
The performance of a score on a language is the mean $\bar{r}$, namely the average value of $\bar{r}$ over all the sentences of that language. It is easy to see that the expected average normalized rank of the random baseline, that consists of selecting a random vertex as root, is 1/2, as the following property states. 
\begin{property}
\label{prop:ranking}
The expectation of $r$ and $\bar{r}$ according to a random baseline that picks a random vertex as root of a tree of size $n$ are
\begin{align*}
\mathbb{E}[r] & = \frac{n + 1}{2} \\
\mathbb{E}[\overline{r}] & = \frac{1}{2}. \\
\end{align*}
\end{property}
\begin{proof}
Simply (recall Equation \ref{eq:average_rank}),
$$\mathbb{E}[r] = r(1, n) = (n+1)/2.$$ 
By the linearity of expectation, 
\begin{equation*}
\mathbb{E}[\overline{r}] = \frac{1}{n - 1}(\mathbb{E}[r] - 1) = 1/2.
\end{equation*}
\end{proof}

\subsubsection{Classification}

We also evaluate the centrality scores by their capacity to classify a vertex as root. Each score is used to build a binary classification model. Suppose a centrality score that is to be maximized to find the root.
The model classifies the vertex or vertices that maximize the score as root and the other as non root. The random baseline model selects a vertex uniformly at random, that is classified as root, while the remainder of vertices are classified as non root. 

All the classification models are evaluated by means of traditional scores from the field of supervised machine learning: precision, recall and the $F$-measure, that is the harmonic mean of precision and recall, i.e.
$$F\mbox{-}measure = \frac{2}{\frac{1}{precision} + \frac{1}{recall}}.$$
We define $N_M$ as the number of pairs produced by the model and $N_S$ as the number of actual pairs. Notice that $N_S$ is also the number of sentences of the treebank, as every sentence has a single root. 
We define $h$ as the number of hits (true positives), namely the number of pairs produced by the model that are also found among the actual pairs. Then 
\begin{align*}
precision & = \frac{h}{N_M}\\
recall & = \frac{h}{N_S}.
\end{align*}
For the random baseline model, the following property indicates that the expected value of each of the evaluation metrics is just the inverse of the harmonic mean of sentence length.

\begin{property}
\label{prop:baseline}
\begin{equation*}
\E[precision] = \E[recall] = \E[F\mbox{-}measure] = \frac{1}{N_S} \sum_{i=1}^{N_S} \frac{1}{\nu_i},
\end{equation*}
where $\nu_i$ is the sentence length (in words) of the $i$-th sentence and $N_S$ is the total number of sentences.
\end{property}
\begin{proof}
For the baseline model, $N_M = N_S$ because the model only one makes one random guess per sentence and then 
\begin{equation*}
precision = recall = F\mbox{-}measure = \frac{h}{N_S}. \\
\end{equation*} 
Since $N_S$ is constant, the expected value of precision and recall can be expressed as
\begin{equation*}
\E[precision] = \E[recall] = \E\left[\frac{h}{N_S}\right] = \frac{\E[h]}{N_S}.
\end{equation*}
$h$ can be decomposed as
\begin{equation*}
h = \sum_{i=1}^{N_S} h_i,
\end{equation*}
where $h_i$ is a Bernoulli variable that indicates if the baseline model has guessed the correct root vertex for the $i$-th sentence ($h_i = 1$ it the guess is right; $h_i = 0$ otherwise).
The probability that the baseline model guesses the right root for the $i$-th sentence is 1/$\nu_i$, where $\nu_i$ is the number of words of the $i$-th sentence.
Then 
\begin{equation*}
\E[h] = \E\left[\sum_{i=1}^{N_S} h_i\right] = \sum_{i=1}^{N_S} \E[h_i] = \sum_{i=1}^{N_S} \frac{1}{\nu_i}.
\end{equation*}
Finally,
\begin{equation*}
\E[precision] = \E[recall] = \E[F\mbox{-}measure] = \frac{1}{N_S} \sum_{i=1}^{N_S} h_i. 
\end{equation*}
\end{proof}

Let us consider $g$, the number of guesses that a model produces for a given sentence. 
For a model based on the center or the centroid, $1 \leq g \leq 2$, because each tree has one or two center and one or two centroids. 
For the degree centrality model, $1 \leq g \leq n - l$ where $l$ is the number of leaves of the free tree. $g$ is minimum for a star tree, where $l = n - 1$ and $g = 1$, and maximum for a path (or linear tree), where $l = n - 2$ and $g = n - 2$. Since every sentence has one root, the number of false positives that a model produces for a sentence, is at least $g - 1$.

The false discovery rate is 
$$FDR = \frac{fp}{tp + fp},$$
where $tp$ is the number of true positives and $fp$ is the number of false positives. 
Then precision can also be defined equivalently as 
\begin{equation*}
precision =  1 - FDR
\end{equation*}
Since $N_M = tp + fp$ and $fp \geq N_M - N_S$, 
it turns out that 
\begin{align}
precision & \leq \frac{N_S}{N_M} \label{eq:precision_upper_bound} \\ 
FDR & \geq 1 - \frac{N_S}{N_M}. \nonumber
\end{align}
Thus, precision is limited in models that produce more than one guess per sentence (Figure \ref{fig:all_unlabelled_trees_with_centers}). 

\subsection{Small sequences}

\begin{figure}
\begin{center}
\includegraphics[width = 0.8 \textwidth]{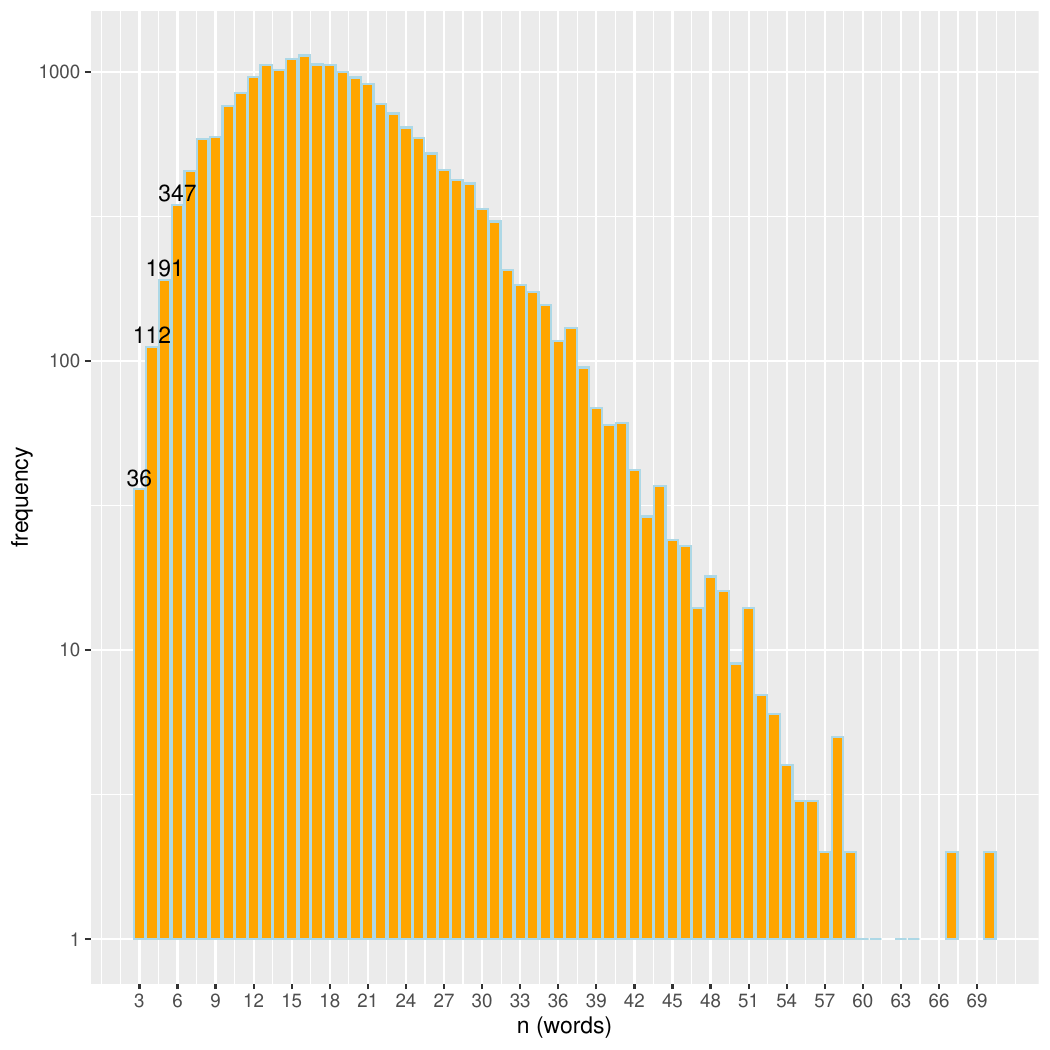}
\end{center}
\caption{\label{fig:sentence_length_distribution} The distribution of sentence length ($n$) given an annotation style over all languages. The frequency of lengths up to $n = 6$ is shown on top of each bar. Notice that the $y$-axis is in logarithmic scale. Only $3.3\%$ of sentences are of length 6 or smaller. }
\end{figure}

When investigating the performance of the centrality scores on short sentences for a given $n$, we mix the sentences of distinct languages because of the scarcity of short sentences and our focus on a language-independent notion of rootness. As for the former reason, Figure \ref{fig:sentence_length_distribution} shows the distribution of sentence lengths when mixing languages, that is identical for each annotation style. The fact that there are 21 languages but only 36 sentences when $n=3$ and $112$ sentences when $n=4$ motivates the mixing.

\subsection{Visualization}

In boxplots, the thick line indicates the median. The lower and upper hinges correspond to the first and third quartiles (the 25th and 75th percentiles). The upper whisker extends from the hinge to the largest value no further than $1.5 \cdot IQR$ from the hinge (where $IQR$ is the inter-quartile range, or distance between the first and third quartiles). The lower whisker extends from the hinge to the smallest value at most $1.5 \cdot IQR$ of the hinge. Data beyond the end of the whiskers are called ``outlying'' points and are plotted individually. Individual points correspond to languages.

\subsection{Implementation}

A word of caution is required for the implementation of centrality scores that do not produce an integer number: closeness and straightness centralities, that involve sums of rational numbers. 
If the sums of rational numbers are implemented as sums of real numbers it is possible that two vertices that indeed have the same centrality get distinct centrality values because of numerical precision errors. This can be addressed in three ways. First, neglecting the problem, assuming that the problem will have a low frequency among the vertices of maximum centrality. If the problem is not addressed, a likely consequence is to finding just one of the vertices of maximum centrality and then introducing a bias towards higher precision or lower precision. Another problem is also possible: that two vertices end up having the same centrality because of numerical precision problems while they do not have actually the same centrality. The second solution consists of introducing a tolerance error in the comparison of centrality values, which raises the question of the appropriate value of that threshold and introduces an additional parameter into the analyses. The third solution, the one we have adopted, is computing these sums as sums of rational numbers with an exact method. \footnote{The method consist of keeping the numerator and the denominator of a rational numbers as smalls as possible by means of the $gcd$ (the greatest common divisor); when summing to rational numbers, the magnitude of the denominator is reduced by using $lcm$ (the least common multiplier), of the denominators of the summands.}

\section{Results}

\label{sec:results}

We analyze the performance of the scores from two perspectives: ranking, i.e. by their ability to rank the root at the top (Section \ref{subsec:ranking}) or as binary classification problem, i.e. by their ability to identify the root vertex in general (Section \ref{subsec:classification}) or in short sentences (Section \ref{subsec:short_sequences}). We consider a series of evaluation metrics, i.e. mean normalized rank (mean $\bar{r}$) for ranking, as well as precision, recall and $F$-measure for binary classification. 
Then, every score is evaluated with respect to these metrics on each language. 
On top of that, we determine the best and the worst score according to a certain evaluation metric by considering both the mean and the median of the score over languages so as to get more robust conclusions. For instance, we say that a certain centrality score is the best in terms of precision if its mean and median are smaller than that of any of the other centrality scores. We extend the criterion to groups of centrality scores. 
For instance, we say that a set of centrality score contains the best scores in terms of precision if their means and medians over languages are smaller than that of any of the other centrality scores. 

\subsection{Ranking}

\label{subsec:ranking}

\begin{figure}
\centering
\includegraphics[height = 0.85 \textheight]{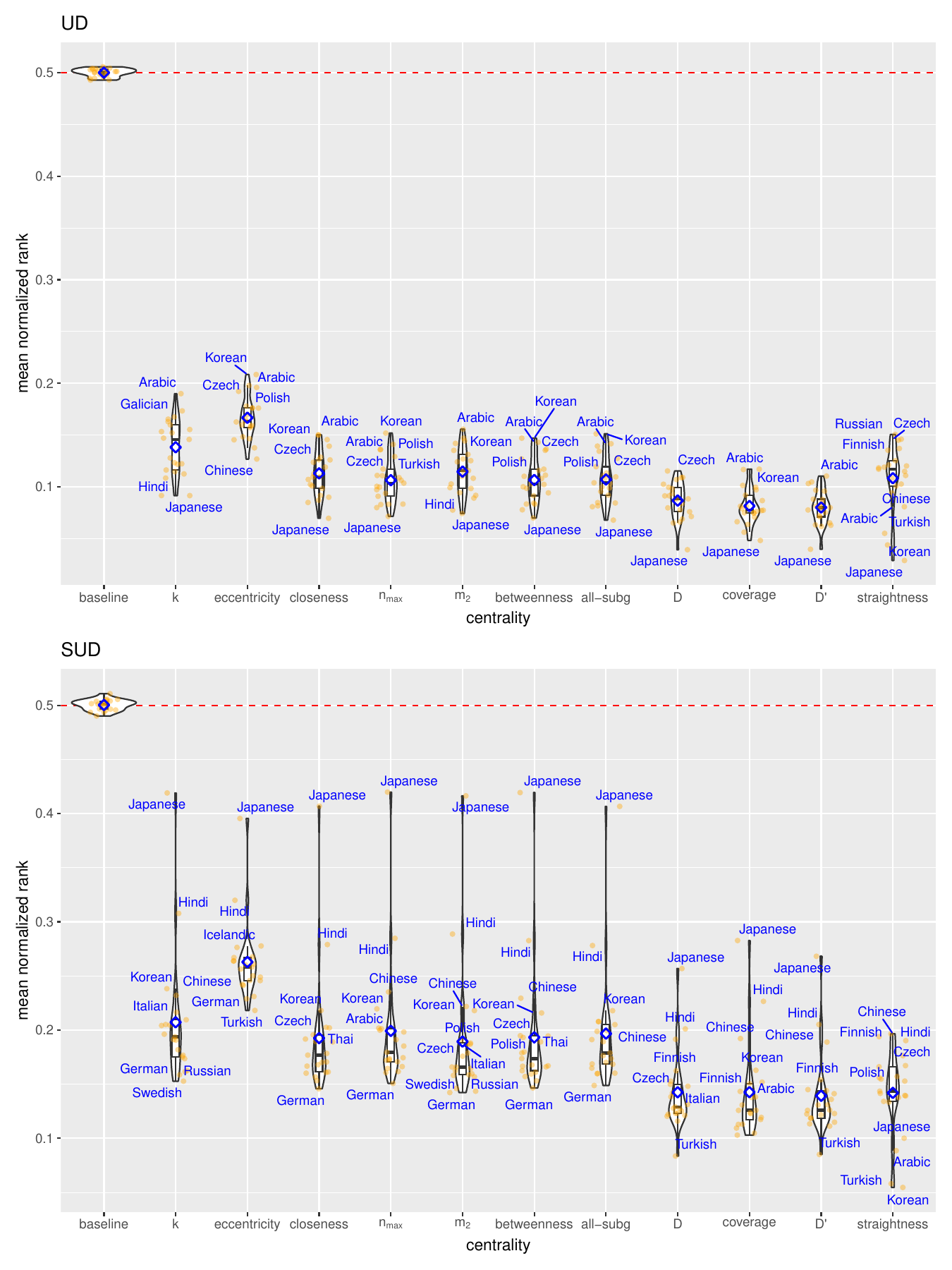}
\caption{\label{fig:boxplot_centrality_scores_ranking} The distribution of the mean $\bar{r}$, the mean normalized rank, by means of a combined boxplot and violin plot across languages for each centrality score when using UD (top) and SUD (bottom) annotation style. The mean normalized rank of a score is computed for each language by averaging the normalized rank for that score over all  sentences. For each centrality score, black thick lines indicate medians while blue diamonds indicate means. 
``baseline'' refers to the random baseline. The red dashed line indicates the expected normalized rank according to the random baseline (Property \ref{prop:ranking}).  
}
\end{figure}

Figure \ref{fig:boxplot_centrality_scores_ranking} summarizes the performance of the centrality scores according their ability to rank the root vertex. A perfect centrality score would assign rank 1 (i.e. $\bar{r} = 0$), to the root vertex. Independently of the annotation style, we find that
\begin{enumerate}
\item
All centrality scores tend to put the root near top positions (near rank 1). The mean $\bar{r}$ over all sentences of a language is far from the 1/2 predicted by the random baseline (Property \ref{prop:ranking}) except for non-spatial scores in Japanese when using SUD annotation style, where the average rank is $\approx 0.4$.
All the scores, except eccentricty, perform better than the degree centrality score.
\item 
The best scores are the new spatial scores in UD and all the spatial scores in SUD, that manage to get closer to top positions.  
\item 
Among the non-spatial scores, degree centrality and eccentricity are clearly the worst scores for UD whereas eccentricity is the worst for SUD. 
\item
The performance of the scores is generally higher with UD annotation style (for instance, languages with a normalized average rank above 0.2 are exceptional in UD but they abound in SUD). In addition, SUD shows marked outliers (Japanese and Hindi).
\end{enumerate} 
\iftoggle{arxiv}{Appendix \ref{appendix:ranking_summary}}{Supplementary Material, Section \ref{appendix:ranking_summary}}
shows further details on the distribution of the performance of each centrality score across languages according to rank-based 
evaluation metrics for each annotation style. It also considers a state-of-the-art ranking score from the field of information retrieval, i.e. discounted cumulative gain (DCG)  \citep{Croft2010a}. DCG was originally designed for evaluating systems that can retrieve a large number of documents and then introduces a logarithmic correction on ranks that is not powerful enough for the problem of retrieving the root vertex because most sentence lengths vary within the same order of magnitude (Figure \ref{fig:sentence_length_distribution}). 
Then it is not surprising that the qualitative results above are almost the same when rank is replaced by DCG \iftoggle{arxiv}{(Appendix \ref{appendix:ranking_summary}).}
{(Supplementary Material, Section \ref{appendix:ranking_summary}).}

\subsection{Classification}

\label{subsec:classification}

\begin{figure}
\centering
\includegraphics[height = 0.88 \textheight]{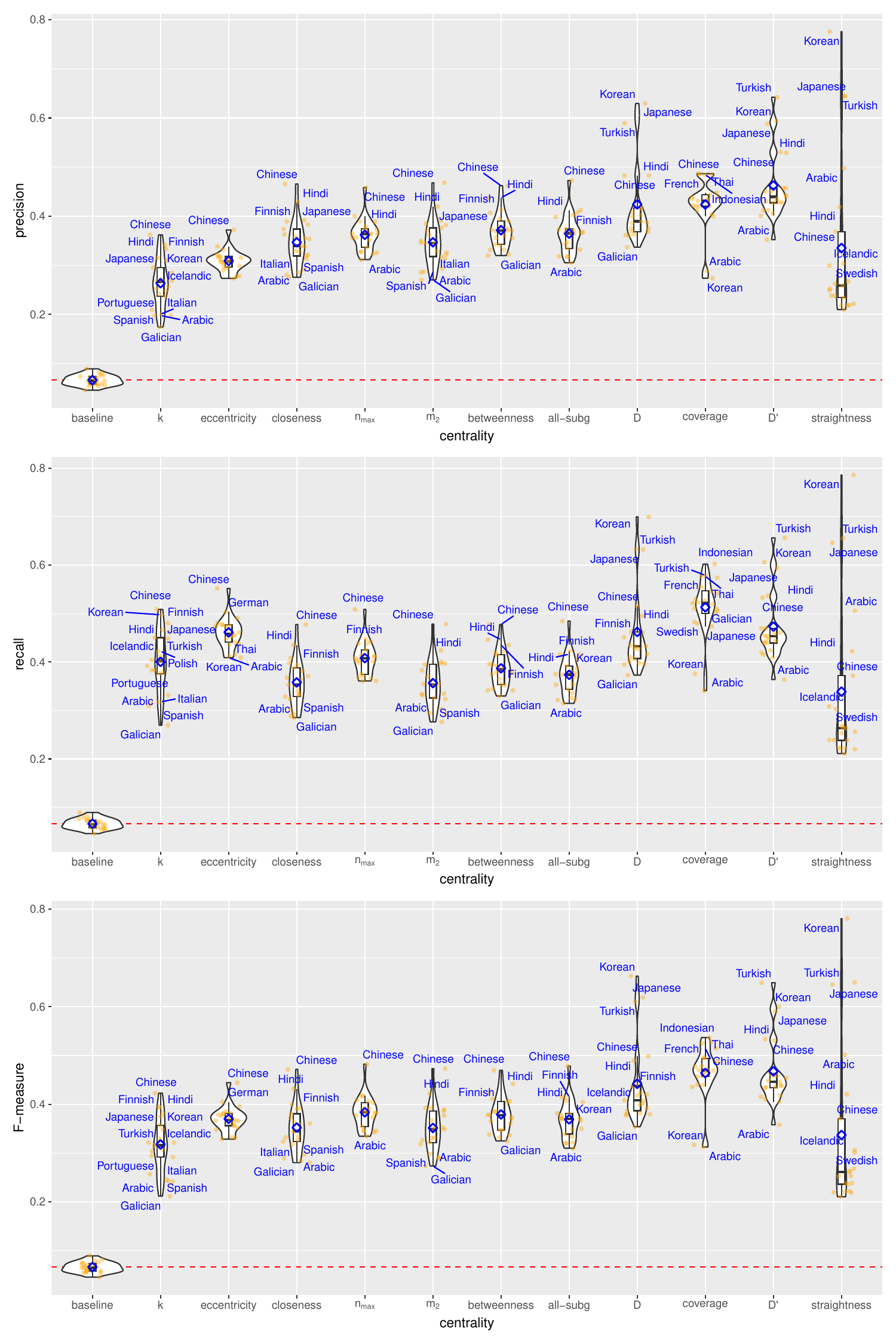}
\caption{\label{fig:boxplot_centrality_scores_UD} 
The distribution of the performance of each classification model across languages when using UD annotation style. Performance is evaluated using the following metrics: precision (top), recall (middle) and $F$-measure (bottom). For each centrality score, black thick lines indicate medians while blue diamonds indicate means. ``baseline'' refers to the random baseline. The red dashed line indicates the mean (over languages) of the expected value of the evaluation score (Property \ref{prop:baseline}).
}
\end{figure}

\begin{figure}
\centering
\includegraphics[height = 0.88 \textheight]{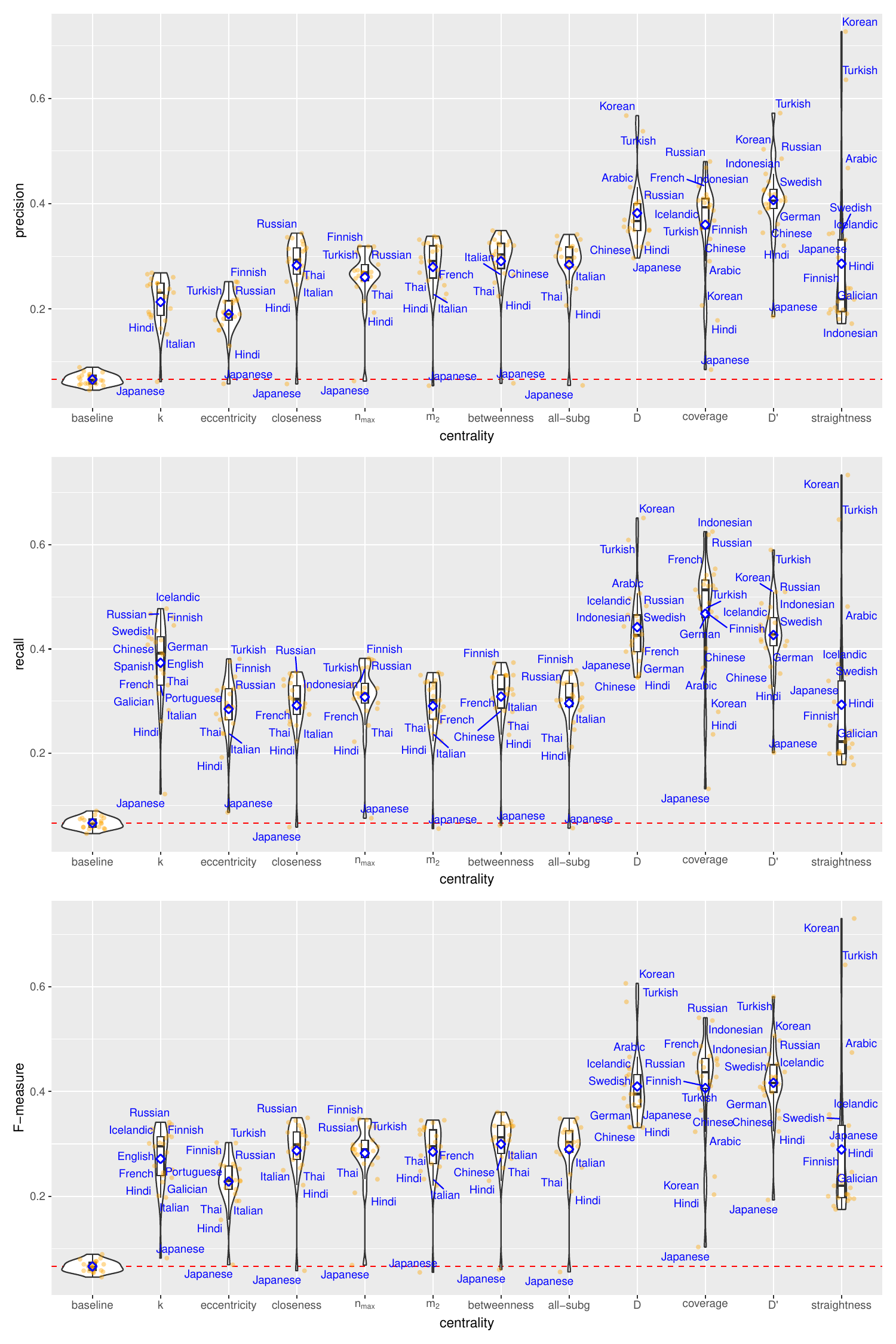}
\caption{\label{fig:boxplot_centrality_scores_SUD} 
The distribution of the performance of each classification model across languages when using SUD annotation style depending on the evaluation metric. The format is the same as in Figure \ref{fig:boxplot_centrality_scores_UD}.
}
\end{figure}

Figures \ref{fig:boxplot_centrality_scores_UD} and \ref{fig:boxplot_centrality_scores_SUD} summarize the performance of the classification models according to standard evaluation metrics. The main results are 
\begin{itemize}
\item 
Random baseline. The random baseline model is always the worst model by far independently of the evaluation metric and the annotation style.
\item 
Precision. Both in UD and SUD, the new corrected $D$ is the centrality score with highest precision. The three scores with highest precision are the new spatial scores. 
The degree centrality baseline is the worst non-random score in UD and the second worst (after eccentricity) in SUD.
\item
Recall. Both in UD and SUD, coverage is the centrality score with highest recall. In UD, the other new spatial scores and eccentricity are the following best scores. In SUD, the three best scores are the new spatial scores.
The degree centrality baseline is better than all or almost all non-spatial scores and straightness but degree centrality is never better than any of the new spatial scores.
\item 
$F$-measure. Both in UD and SUD, the three scores with highest $F$-measure are the three new spatial scores. There is no clear single best when just focusing on UD or SUD.  
The degree centrality baseline recapitulates the behavior for precision: is the worst non-random score in UD and the second worst (after eccentricity) in SUD.
\end{itemize}
See \iftoggle{arxiv}{Appendix \ref{appendix:evaluation_summary}}{Supplementary Material Section \ref{appendix:evaluation_summary}} for further details about the performance of the classification models.

\subsection{Short sentences}

\label{subsec:short_sequences}

\begin{table}[h]
\caption{\label{tab:performance_versus_tree_size_UD} The correlation between tree size and the performance of the centrality scores for trees of given kind on short sequences ($3 \leq n \leq 6$) mixing languages in PUD and using UD annotation style. We only show the representatives of the non-spatial centrality scores according to the classes of equivalence for each kind of tree (Corollary \ref{cor:classes_of_equivalence_on_path_quasistar_star}). For star trees, we omit the spatial centrality scores that are equivalent to degree centrality, i.e. $D(v)$ and $D'(v)$ (Property \ref{prop:new_spatial_scores_on_star}).
We report the sample size ($N$), that is the number of trees of given kind such that $3 \leq n \leq 6$, and the Kendall rank correlation ($\tau$) (we do not report the $p$-value of a two-sided test because $N$ is below 5, the critical value needed to find significance in a two-side correlation test and a significance level $\alpha = 0.05$; see \cite[Table 3]{Ferrer2012a}).
}
\begin{tabular}{llllll}
\toprule
Kind & Centrality score & Evaluation score & $N$ & $\tau$ \\
\midrule
star & $k$ & precision & 4 & 1\\
star & $k$ & recall & 4 & 1\\
star & $k$ & F-measure & 4 & 1\\
star & coverage & precision & 4 & 1\\
star & coverage & recall & 4 & 0.667\\
star & coverage & F-measure & 4 & 1\\
star & straightness & precision & 4 & 0.333\\
star & straightness & recall & 4 & 0.667\\
star & straightness & F-measure & 4 & 0.667\\
path & $k$ & precision & 4 & -1\\
path & $k$ & recall & 4 & 0.913\\
path & $k$ & F-measure & 4 & -1\\
path & eccentricity & precision & 4 & -1\\
path & eccentricity & recall & 4 & -0.333\\
path & eccentricity & F-measure & 4 & -1\\
path & $D$ & precision & 4 & -1\\
path & $D$ & recall & 4 & -0.667\\
path & $D$ & F-measure & 4 & -1\\
path & coverage & precision & 4 & -0.667\\
path & coverage & recall & 4 & -1\\
path & coverage & F-measure & 4 & -1\\
path & straightness & precision & 4 & -1\\
path & straightness & recall & 4 & -1\\
path & straightness & F-measure & 4 & -1\\
 \\
\botrule
\end{tabular}
\end{table}

\begin{table}[h]
\caption{\label{tab:performance_versus_tree_size_SUD} The correlation between tree size and the performance of the centrality scores for trees of given kind on short sequences ($3 \leq n \leq 6$) mixing languages in PUD and using SUD annotation style. The format is the same as in Table \ref{tab:performance_versus_tree_size_UD}.}
\begin{tabular}{llllll}
\toprule
Kind & Centrality score & Evaluation score & $N$ & $\tau$ \\
\midrule
star & $k$ & precision & 4 & 1\\
star & $k$ & recall & 4 & 1\\
star & $k$ & F-measure & 4 & 1\\
star & coverage & precision & 4 & 1\\
star & coverage & recall & 4 & 0.667\\
star & coverage & F-measure & 4 & 1\\
star & straightness & precision & 4 & $0$\\
star & straightness & recall & 4 & 1\\
star & straightness & F-measure & 4 & 0.333\\
path & $k$ & precision & 4 & -1\\
path & $k$ & recall & 4 & 0.667\\
path & $k$ & F-measure & 4 & -1\\
path & eccentricity & precision & 4 & -1\\
path & eccentricity & recall & 4 & -0.333\\
path & eccentricity & F-measure & 4 & -0.667\\
path & $D$ & precision & 4 & -1\\
path & $D$ & recall & 4 & -0.667\\
path & $D$ & F-measure & 4 & -1\\
path & coverage & precision & 4 & -1\\
path & coverage & recall & 4 & -1\\
path & coverage & F-measure & 4 & -1\\
path & straightness & precision & 4 & -1\\
path & straightness & recall & 4 & -1\\
path & straightness & F-measure & 4 & -1\\
 \\
\botrule
\end{tabular}
\end{table}

We highlight some results about the performance of the scores on short sentences with $3 \leq n \leq 6$ (Tables \ref{tab:complete_centrality_scores_UD_small_sequences} and \ref{tab:complete_centrality_scores_SUD_small_sequences}). We find two major patterns. First, the performance of the scores tends to increase as $n$ increases in star trees while we find an opposite effect in path trees, the performance of the centrality scores tends to reduce as $n$ increases both in UD and SUD (Tables \ref{tab:performance_versus_tree_size_UD} and \ref{tab:performance_versus_tree_size_SUD}). The only exception to this pattern is that the performance only increases as tree size increases in path trees when the centrality score is $k$ and the performance is measured by recall. Second, we find that the performance of the scores tends to decrease as the hubiness of the tree increases.
(Tables \ref{tab:performance_versus_hubiness_UD} and \ref{tab:performance_versus_hubiness_SUD}). 
The effect is more marked in SUD, where the trend is only broken by recall in a few cases ($k$ when $n = 4$ and $n = 6$; $D$ when $n = 4$); in UD, the trend is broken by recall ($k$ and $D$ when $n = 4$ and $n = 6$; eccentricity when $n = 6$), precision (only for coverage and $n = 6$) and $F$-measure (coverage and straightness when $n = 6$). 
To simplify detailed reporting, we focus on the $F$-measure. For specific unlabelled trees with a given $n$, we find (Tables \ref{tab:complete_centrality_scores_UD_small_sequences} and \ref{tab:complete_centrality_scores_SUD_small_sequences}):
\begin{itemize}
\item 
Path trees ($n \geq 3$). Both in UD and SUD, the best or 2nd best score belongs to the degree centrality class or it is an Euclidean distance score (excluding straightness centrality). Recall that $k$ is consistent with $D$ and $D'$ on star trees (Property \ref{prop:new_spatial_scores_on_star}).
\item 
Quasipath trees and balanced bistar trees with $n = 6$. The degree centrality class or the spatial scores (except straightness centrality) give the best or 2nd best performance in terms of $F$-measure.
\item
Quasistar trees ($n \geq 5$). The best score is always an Euclidean distance score (other than straightness). The difference in performance of the best score with respect to the degree centrality class is small.
\item
Star trees ($n \geq 3$). The best scores are in the degree centrality class. Recall that $k$ is consistent with $D$ and $D'$ on star trees (Property \ref{prop:new_spatial_scores_on_star}). Thus, taking into account space (linear order) does not help. 
\end{itemize}
When all trees of same $n$ are mixed, the best scores are the new spatial scores in most cases (see results for conditioning just on $n$ in Tables \ref{tab:complete_centrality_scores_UD_small_sequences} and \ref{tab:complete_centrality_scores_SUD_small_sequences}).

\begin{table}[h]
\caption{\label{tab:performance_versus_hubiness_UD} The correlation between the hubiness of a tree structure and the performance of the centrality scores for short sequences of same length ($n$) mixing languages in PUD and using UD annotation style. We show all spatial centrality scores. For non-spatial scores, we use only a strictly necessary representative of the classes of equivalence 
that result from conditioning on the size of the tree (Property \ref{prop:classes_given_tree_size}). 
We report the sample size ($N$), that is the number unlabelled trees of $n$ vertices (Figure \ref{fig:all_unlabelled_trees}) and the Kendall rank correlation ($\tau$) (we do not report the $p$-value of a two-sided test because $N$ is in most cases below 5 or too close to 5, the critical value needed to find significance in a two-side correlation test and a significance level $\alpha = 0.05$; see \cite[Table 3]{Ferrer2012a}).
}
{\tiny
\begin{tabular}{llllll}
\toprule
$n$ & Centrality score & Evaluation score & $N$ & $\tau$ \\
\midrule
4 & $k$ & precision & 2 & 1\\
4 & $k$ & recall & 2 & -1\\
4 & $k$ & F-measure & 2 & 1\\
4 & $D$ & precision & 2 & 1\\
4 & $D$ & recall & 2 & -1\\
4 & $D$ & F-measure & 2 & 1\\
4 & coverage & precision & 2 & 1\\
4 & coverage & recall & 2 & 1\\
4 & coverage & F-measure & 2 & 1\\
4 & $D'$ & precision & 2 & 1\\
4 & $D'$ & recall & 2 & 1\\
4 & $D'$ & F-measure & 2 & 1\\
4 & straightness & precision & 2 & 1\\
4 & straightness & recall & 2 & 1\\
4 & straightness & F-measure & 2 & 1\\
5 & $k$ & precision & 3 & 0.333\\
5 & $k$ & recall & 3 & 0.333\\
5 & $k$ & F-measure & 3 & 0.333\\
5 & eccentricity & precision & 3 & 0.333\\
5 & eccentricity & recall & 3 & 0.333\\
5 & eccentricity & F-measure & 3 & 0.333\\
5 & $D$ & precision & 3 & 0.333\\
5 & $D$ & recall & 3 & 1\\
5 & $D$ & F-measure & 3 & 0.333\\
5 & coverage & precision & 3 & 0.333\\
5 & coverage & recall & 3 & 0.333\\
5 & coverage & F-measure & 3 & 0.333\\
5 & $D'$ & precision & 3 & 0.333\\
5 & $D'$ & recall & 3 & 1\\
5 & $D'$ & F-measure & 3 & 0.333\\
5 & straightness & precision & 3 & 0.333\\
5 & straightness & recall & 3 & 1\\
5 & straightness & F-measure & 3 & 1\\
6 & $k$ & precision & 6 & 0.276\\
6 & $k$ & recall & 6 & -0.071\\
6 & $k$ & F-measure & 6 & 0.138\\
6 & eccentricity & precision & 6 & 0.414\\
6 & eccentricity & recall & 6 & $0$\\
6 & eccentricity & F-measure & 6 & 0.138\\
6 & $n_{max}$ & precision & 6 & 0.276\\
6 & $n_{max}$ & recall & 6 & 0.414\\
6 & $n_{max}$ & F-measure & 6 & 0.138\\
6 & $D$ & precision & 6 & 0.357\\
6 & $D$ & recall & 6 & -0.276\\
6 & $D$ & F-measure & 6 & 0.138\\
6 & coverage & precision & 6 & $0$\\
6 & coverage & recall & 6 & 0.138\\
6 & coverage & F-measure & 6 & $0$\\
6 & $D'$ & precision & 6 & 0.138\\
6 & $D'$ & recall & 6 & -0.138\\
6 & $D'$ & F-measure & 6 & 0.138\\
6 & straightness & precision & 6 & 0.276\\
6 & straightness & recall & 6 & 0.138\\
6 & straightness & F-measure & 6 & $0$\\
 \\
\botrule
\end{tabular}
}
\end{table}

\begin{table}[h]
\caption{\label{tab:performance_versus_hubiness_SUD} 
The correlation between the hubiness of a tree structure and the performance of the centrality scores for short sequences of same length ($n$) mixing languages in PUD and using SUD annotation style. The format is the same as in Table \ref{tab:performance_versus_hubiness_UD}. }
{\tiny
\begin{tabular}{llllll}
\toprule
$n$ & Centrality score & Evaluation score & $N$ & $\tau$ \\
\midrule
4 & $k$ & precision & 2 & 1\\
4 & $k$ & recall & 2 & -1\\
4 & $k$ & F-measure & 2 & 1\\
4 & $D$ & precision & 2 & 1\\
4 & $D$ & recall & 2 & -1\\
4 & $D$ & F-measure & 2 & 1\\
4 & coverage & precision & 2 & 1\\
4 & coverage & recall & 2 & 1\\
4 & coverage & F-measure & 2 & 1\\
4 & $D'$ & precision & 2 & 1\\
4 & $D'$ & recall & 2 & 1\\
4 & $D'$ & F-measure & 2 & 1\\
4 & straightness & precision & 2 & 1\\
4 & straightness & recall & 2 & 1\\
4 & straightness & F-measure & 2 & 1\\
5 & $k$ & precision & 3 & 0.333\\
5 & $k$ & recall & 3 & 0.333\\
5 & $k$ & F-measure & 3 & 0.333\\
5 & eccentricity & precision & 3 & 0.333\\
5 & eccentricity & recall & 3 & 0.333\\
5 & eccentricity & F-measure & 3 & 0.333\\
5 & $D$ & precision & 3 & 0.333\\
5 & $D$ & recall & 3 & 0.333\\
5 & $D$ & F-measure & 3 & 0.333\\
5 & coverage & precision & 3 & 0.333\\
5 & coverage & recall & 3 & 0.333\\
5 & coverage & F-measure & 3 & 0.333\\
5 & $D'$ & precision & 3 & 0.333\\
5 & $D'$ & recall & 3 & 0.333\\
5 & $D'$ & F-measure & 3 & 0.333\\
5 & straightness & precision & 3 & 0.333\\
5 & straightness & recall & 3 & 1\\
5 & straightness & F-measure & 3 & 0.333\\
6 & $k$ & precision & 6 & 0.414\\
6 & $k$ & recall & 6 & $0$\\
6 & $k$ & F-measure & 6 & 0.138\\
6 & eccentricity & precision & 6 & 0.414\\
6 & eccentricity & recall & 6 & 0.138\\
6 & eccentricity & F-measure & 6 & 0.138\\
6 & $n_{max}$ & precision & 6 & 0.276\\
6 & $n_{max}$ & recall & 6 & 0.414\\
6 & $n_{max}$ & F-measure & 6 & 0.138\\
6 & $D$ & precision & 6 & 0.276\\
6 & $D$ & recall & 6 & 0.071\\
6 & $D$ & F-measure & 6 & 0.138\\
6 & coverage & precision & 6 & 0.138\\
6 & coverage & recall & 6 & 0.214\\
6 & coverage & F-measure & 6 & 0.138\\
6 & $D'$ & precision & 6 & 0.138\\
6 & $D'$ & recall & 6 & 0.138\\
6 & $D'$ & F-measure & 6 & 0.138\\
6 & straightness & precision & 6 & 0.276\\
6 & straightness & recall & 6 & 0.138\\
6 & straightness & F-measure & 6 & 0.276\\
 \\
\botrule
\end{tabular}
}
\end{table}

\section{Discussion}

\label{sec:discussion}

We have investigated the general problem of finding the root of a free tree \citep{Riveros2023a} in the context of syntactic dependency structures. 
We have 
found support for the hypothesis that the root of a syntactic dependency structure is a word of high centrality in the free tree (or both the free tree and the linear arrangement) in the sense that all centrality scores tend to put the root vertex in top positions (Figure \ref{fig:boxplot_centrality_scores_ranking}).

\subsection{The baselines}

It may not be surprising that all the centrality scores perform better than the random baselines. For that reason, we presented degree centrality as a stronger baseline in Section \ref{sec:centrality}. The new spatial scores never perform worse than degree centrality in spite of their close theoretical relationship with vertex degree (Section \ref{subsubsec:degree_versus_new_spatial_scores}). When considering the ability of a centrality score to put the root in top positions, we find that eccentricity tends to perform worse than degree centrality (Figure \ref{fig:boxplot_centrality_scores_ranking}). When considering the classification models, degree centrality never performs better than the new spatial scores but performs better in the following conditions: recall of closeness, $m_2(v)$ and betweenness in UD; precision of eccentricity in SUD, recall of all other non-spatial scores in SUD and $F$-measure of eccentricity in SUD (Figures \ref{fig:boxplot_centrality_scores_UD} and \ref{fig:boxplot_centrality_scores_SUD}). Thus, our findings on recall in SUD indicate that all the complexity of the non-spatial scores is totally useless in that setting. 

Japanese yields a shocking result in SUD annotation: precision and recall (and therefore $F$-measure) are close to the random baseline for non-spatial scores (\ref{fig:boxplot_centrality_scores_SUD}), which questions the validity of our hypothesis about the nature of roots for Japanese in SUD annotation. However, when examining the mean normalized rank for these centrality scores (Figure \ref{fig:boxplot_centrality_scores_ranking}), Japanese consistently deviates from the random baseline, indicating that the hypothesis still holds for Japanese in SUD, but only as a weak trend for these scores.

\subsection{The best scores}

{\em A priori,} spatial scores are expected to outperform non-spatial scores because they exploit more information (both the free tree structure and vertex positions). By the same token, all else being equal, scores that exploit global information about the free tree (e.g., the shortest path distances in the tree) should perform better than scores that exploit only local information (e.g., the neighbours of a vertex). This suggests that the best score should be a global spatial score (namely, straightness). Contrary to our expectation, we have found that the best scores are local spatial scores. Details follow.

Unsurprisingly, the new spatial centrality scores ($D$, $D'$ and coverage) have the highest ability to place the root vertex in top positions. Surprisingly, straightness centrality, a spatial score that exploits global information of the free tree, has a lower performance even with respect to non-spatial scores especially in UD (Figures \ref{fig:boxplot_centrality_scores_ranking}, \ref{fig:boxplot_centrality_scores_UD} and \ref{fig:boxplot_centrality_scores_SUD}). It is interesting that the new spatial scores beat the non-spatial scores just by exploiting local information.

It could be argued that the average performance of the best classification model in sentences of any length is poor (about $42-47\%$)\footnote{$D'$ reaches an average $F$-measure over languages of 0.417 in UD and 0.468 in SUD; Tables \ref{tab:summary_centrality_scores_UD} and \ref{tab:summary_centrality_scores_SUD}\iftoggle{arxiv}{.}{ in Supplementary Material.}} with respect to state-of-the-art unsupervised dependency parsing \citep{Han2020a} or the recently introduced deep supervised parsing methods \citep{Kulmizev2019a}. The latter are able to guess the correct arc and the corresponding label with an accuracy of $85\%$ or more. \footnote{Here we refer to labelled attachment score, that is just percentage of correct arcs, relative to
the gold standard, but ignoring arc labels. 
} Notice however, that our classification models are parameter-less and that they are not taking into account any information about the words attached to the free tree vertices, e.g., the word form or its part-of-speech \citep{Han2020a},\footnote{Recall that \cite{Soegaard2012a,Soegaard2012b} also exploited that information.} neither any information outside the sentence such as word ontologies or word embeddings, as it is customary in traditional supervised and unsupervised parsing methods \citep{Jurafsky2024a,Han2020a}. 

\subsection{Why do the new local spatial scores work?}

We have not found a single score that is able to find unequivocally, or with very high accuracy, the root. We have found, however, that our local spatial scores exhibit a high sensitivity to the rootness of a vertex. 
We introduced the corrected $D$ ($D'$) hoping that it would perform better than $D$. There is a slight tendency of $D'$ to perform better than $D$ (Figure \ref{fig:boxplot_centrality_scores_ranking}), that becomes evident when looking at the performance of the classification models (Figures \ref{fig:boxplot_centrality_scores_UD} and \ref{fig:boxplot_centrality_scores_SUD}). Then our fear that simply $D$ could retrieve heads due to anti dependency distance minimization in short spans \citep{Ferrer2019a,Ferrer2020b,Ferrer2023b} was totally justified and demonstrates the power of word order theory. Our findings suggest that roots are words that form long dependencies, not because dependency distance minimization is surpassed by other word order principles but rather because they connect distant elements in the sentence as expected in optimal projective arrangements \citep{Gildea2007a,Alemany2021a}. Furthermore, these scores may be able to break ties between centers with respect to non-spatial scores (Figure \ref{fig:all_unlabelled_trees_with_centers}) by combining information on the free tree with positional information.

\subsection{Soft versus hard}

We introduced the distinction of soft versus hard scores to organize a map of centrality scores (Table \ref{tab:features_of_centrality_scores}; Section \ref{subsubsec:soft_versus_hard_centrality_scores}). Our theoretical analysis of node-network metrics revealed that various apparently distinct metrics are, in fact, measuring the same underlying phenomenon, differing only in their final aggregation mechanism (Table \ref{tab:features_of_centrality_scores}). Some metrics summarize a sample of values by aggregating all of them (e.g., using a mean), while others use only an extreme value (e.g., a maximum). We termed the former {\em soft} aggregations and the latter {\em hard} aggregations. For example, we have found a connection between centroid (hard score) and betweenness (soft counterpart) in the context of trees, which was surprising and unknown to us (Section \ref{subsubsec:soft_versus_hard_centrality_scores}). However, in practice, the distinction has not been as fruitful as initially expected. Results are mixed. Details follow.
 
On the side of negative results, we introduced $m_2(v)$ hoping that it would be a soft centrality score that would perform better than its hard correlate, i.e. $n_{max}(v)$ (centroid). The fact is that $m_2(v)$ is worse than $n_{max}(v)$ in UD and slightly better in SUD, both in terms of normalized rank (Figure \ref{fig:boxplot_centrality_scores_ranking}).  
Regarding the classification models, $m_2(v)$ performs worse than $n_{max}(v)$ (in terms of precision, recall and $F$-measure) in UD whereas performance depends on the evaluation metric in SUD (Figures \ref{fig:boxplot_centrality_scores_UD} and \ref{fig:boxplot_centrality_scores_SUD}). Thus, $m_2(v)$ does not show a clear general improvement with respect to its hard version.
The fact that betweenness, which shares ingredients with $m_2(v)$ (Section \ref{prop:betweenness_versus_subtree_sizes_2nd_moment}), yields always better classification models (Figures \ref{fig:boxplot_centrality_scores_UD} and \ref{fig:boxplot_centrality_scores_SUD}), suggests that $m_2(v)$ does not make any addition to the literature on standard centrality scores. Instead, betweenness centrality seems to yield the soft version of $n_{max}(v)$ that we were looking for.  

On the side of positive results, there is a hard centrality score that has been beaten by soft correlates. Eccentricity is a hard score whose soft correlate is closeness. 
Newman's closeness yields higher precision than eccentricity (Figures \ref{fig:boxplot_centrality_scores_UD} and \ref{fig:boxplot_centrality_scores_SUD}). 
As for the popular definition of closeness in Equation \ref{eq:popular_closeness}, the medians (the centers retrieved by that closeness) and the centroids coincide on trees \citep{Kang1975a,Slater1975a}. Given the worse performance of Jordan centers (eccentricity) over centroids except for recall in UD (Figures \ref{fig:boxplot_centrality_scores_UD} and \ref{fig:boxplot_centrality_scores_SUD}), we can conclude that eccentricity has been beaten by another soft correlate although we have not investigated that popular version of closeness directly.

\subsection{The puzzle of Japanese}

Japanese exhibits an extreme behavior. Consider mean normalized rank (Figure \ref{fig:boxplot_centrality_scores_ranking}). When using UD annotation style, Japanese is farthest from the random baseline. However, when using SUD style, Japanese is closest to the random baseline becoming an outlier with respect to the other languages but followed closely by Hindi, another SOV language. Perhaps the most shocking result on Japanese, as explained above, is that precision and recall (and hence $F$-measure) are close to the random baseline for non-spatial scores when SUD annotation style is used (Figure \ref{fig:boxplot_centrality_scores_SUD}).

Such a peculiar behavior of Japanese cannot be attributed to the fact that it is an SOV language as we have indeed three other SOV languages in our collection: Korean, Hindi and Turkish (Table \ref{tab:languages}). Within SOV languages, one may expect that the ease in predicting the root for one of them should be the similar to the ease of predicting the root for another. However, these four languages belong to different linguistic families (Japonic, Koreanic, Indo-European and Turkic, respectively) \footnote{According to Glottolog 5.2 \url{https://glottolog.org/glottolog/language}}, and thus may be seen as independent to a large extent in an evolutionary sense \citep{Winter2021a}. Furthermore, Japanese or its syntactic dependency annotation, deviates from the other SOV languages in various domains
\begin{itemize}
\item
{\em Anti dependency distance minimization effects}. Japanese is the only of them that exhibits syntactic dependency distances that are longer than expected by chance in short sequences \citep[Table 3]{Ferrer2019a}. 
\item
{\em Closeness to a minimum linear arrangement}, namely an ordering that minimizes the sum of dependency distances \citep{Ferrer2020b}. When considering the degree of optimality of syntactic dependency distances in a parallel syntactic dependency treebank formed by 20 languages, Turkish, Korean and Hindi exhibit the lowest, the 2nd lowest and the 4th lowest degree of optimization \citep[Figures 6 and 18]{Ferrer2020b}. When using UD annotation style Japanese is optimized to an $85\%$ while Turkish, Korean and Hindi are optimized to a $64\%$, $65\%$ and $74\%$, respectively \citep[Figure 6]{Ferrer2020b}.
When using SUD annotation style, Japanese is optimized to an $83\%$ while Turkish, Korean and Hindi are optimized to a $63\%$, $66\%$ and $73\%$, respectively\citep[Figure 18]{Ferrer2020b}. 
When looking at statistically significant differences, Japanese is always more optimized than Turkish, Hindi and Korean as well as more optimized than German and Chinese \citep[Figure 7]{Ferrer2020b}.
\item
{\em Projectivity}. Japanese diverges from Korean, Hindi and Turkish concerning the percentage of projective sentences \cite[Tables 2 and 3]{Alemany2022c}. Japanese has the largest proportion of projective sentences in PUD when using UD annotation (Japanese $99.7\%$, Korean $93.6\%$, Hindi $74.3\%$ and Turkish $93.5\%$) but becomes the language with the smallest proportion of projective sentences when using SUD annotation (Japanese $35.8\%$, Korean $75.8\%$, Hindi $43.6\%$ and Turkish $87.6\%$).
\end{itemize}

Then, it is not surprising that Japanese is not always aligned with other SOV languages in our analyses. Concerning mean normalized rank, we find that the scores tend to assign lower ranks to the roots in Japanese compared to Korean in UD annotation style but it is the other way around in SUD. However, when we consider traditional evaluation scores (precision, recall and $F$-measure) and local linear distance scores ($D$, $D'$ or $coverage$) we find that SOV languages (Japanese, Korean and Turkish) cluster together as expected for being SOV languages when using UD annotation style (Figure \ref{fig:boxplot_centrality_scores_UD}) but not when using SUD annotation style (Figure \ref{fig:boxplot_centrality_scores_SUD}). Indeed, Japanese is the language where the centrality scores perform the worst when using SUD annotation style. Critically, the performance of many centrality scores is close to the random baseline suggesting some arbitrariness in the identification of roots in Japanese. Such a finding has two implications for future research. First, as SUD annotations are generated automatically from UD annotations \citep{sud}, it follows that both the original UD annotations for Japanese and the automatic rules that are used to transform UD annotations into SUD annotations should be revised. 
If this revision does not reveal any annotation problem, then the peculiar behavior of Japanese may be reflecting some singular but genuine aspect of the concept of root in that language that deserves further theoretical investigation. 

\subsection{Who is the root?}

The question of who is the root of a syntactic dependency structure can be answered in two ways by means of the classification models (Figures \ref{fig:boxplot_centrality_scores_UD} and \ref{fig:boxplot_centrality_scores_SUD}). 
From a precision perspective, the vertex or vertices that maximize $D'$ are likely to be the roots (with an average probability slightly above $40\%$ in UD and slightly below $40\%$ in SUD). From a recall perspective, the root is likely to be a vertex that maximizes coverage (with an average probability slightly above $50\%$ in UD and below $50\%$ in SUD). These findings suggest that long distance dependencies can fool the classification models based on $D$ or $D'$ and that $D'$ does not clear all confusion caused by long distance dependencies. 
Interestingly, the performance of the new centrality scores is $\approx 60\%$ or greater in certain languages that appear as ``outlying'' points in Figures \ref{fig:boxplot_centrality_scores_UD} and \ref{fig:boxplot_centrality_scores_SUD} (for precise values, check Tables \ref{tab:complete_centrality_scores_UD} and \ref{tab:complete_centrality_scores_SUD}\iftoggle{arxiv}{)}{ in Supplementary Material)}. These languages, tend to be Korean and Turkish and Japanese in UD and Korean and Turkish in SUD, which are among the SOV languages in our sample (Table \ref{tab:languages}). 
We believe that their tendency to put the main verb by the end of the sentence increases the chance that the main verb has longer syntactic dependencies and then the chance of confusing it with other heads reduces.
 
If we restrict the answer to the question above to non-spatial scores, the vertex or vertices that maximize the betweenness centrality (or the centroids in case of UD; or the vertices that maximize closeness in SUD) are likely to be the roots (precision). In contrast, the root is likely to be a Jordan center (eccentricity) in UD and simply a hub (the vertex of maximum degree) in SUD (recall). 
For UD annotation style, the best non-spatial model according to the $F$-measure is the centroid (Figure \ref{fig:boxplot_centrality_scores_UD}).
It is surprising that the centroid is able to predict the root of a syntactic dependency structure with an accuracy of $\approx 40\%$ (for UD) just knowing the undirected links and ignoring any other information (the word labels, their part of speech, their position in the sentence,...).
These findings demonstrate the power of the theory of optimal linear arrangements, namely, arrangements that minimize the sum of syntactic dependency distances \iftoggle{anonymous}
{\citep{Shiloach1979,Iordanskii1987a, Hochberg2003a,anonymous}.}
{\citep{Shiloach1979,Iordanskii1987a, Hochberg2003a,Alemany2021a}.}
Our findings suggest that among the distinct kinds of information that the centrality scores exploit (Table \ref{tab:features_of_centrality_scores}), subtree sizes are the most valuable non-spatial information to find the root of a syntactic dependency tree (precision). This is consistent with the importance of subtree sizes in the theory of optimal linear arrangements, whereby subtrees must be laid out around the centroid in a specific way \citep{Shiloach1979,Chung1984,Iordanskii1987a, Hochberg2003a}. 

We have seen that the performance of the centrality scores improves in short sentences (Tables \ref{tab:complete_centrality_scores_UD_small_sequences} and \ref{tab:complete_centrality_scores_SUD_small_sequences}). In this context, we have found that the root is easier to predict in star-like structures and more difficult to predict in path-like structures (Table \ref{tab:performance_versus_tree_size_UD} and \ref{tab:performance_versus_tree_size_SUD}). Indeed, we have found that ease of prediction is positively correlated with the degree of hubiness or star-likeness (Table \ref{tab:performance_versus_hubiness_UD} and \ref{tab:performance_versus_hubiness_SUD}). Interestingly, linear order is practically irrelevant for a successful guess of the root in sufficiently long sentences with a star tree structure. In star trees, the hub is very likely to be the root, no matter where the hub is placed 
For UD annotation style, precision in Table \ref{tab:complete_centrality_scores_UD_small_sequences} indicates that such probability as a function of $n$ is $0.833$ ($n = 3$), $0.881$ ($n=4$) and $0.939$ ($n = 5$) and $1$ ($n = 6$) (see Table \ref{tab:complete_centrality_scores_UD_small_sequences} for SUD).
Therefore, small star trees are likely to be single head structures. The strong association between the hub and the root in star trees can only be partially accounted for by the theoretical consistency between degree centrality and two of the new spatial centrality scores (Property \ref{prop:new_spatial_scores_on_star}).

\section{Future work}

We have used both UD and SUD annotation style mainly to show the robustness of the major conclusions of the article. However, we have also seen that the performance of the scores tends to be higher with UD than with SUD annotation style (Figures \ref{fig:boxplot_centrality_scores_ranking}, \ref{fig:boxplot_centrality_scores_UD} and \ref{fig:boxplot_centrality_scores_SUD}), suggesting that UD is a better format for the discovery or validation of roots. 
Besides, we have not found a clear advantage of scores that satisfy the tree rooting property. The only circumstance where one can see an advantage in root finding of the non-spatial scores that satisfy the tree rooting property \citep{Riveros2023a} is in recall on UD (Figure \ref{fig:boxplot_centrality_scores_UD}). We suspect that the theoretical advantage may be masked by the kind of trees that are found in syntactic dependency structures and their size. This is suggested by the fact that scores that do not satisfy the tree rooting property, find just one vertex or two connected nodes on specific trees (Figure \ref{fig:all_unlabelled_trees_with_centers}). The question of whether UD annotation style is indeed more suitable for the tree rooting property or root prediction in general is subject of future research.

Here we have investigated the problem of finding the root of a vertex with the simplifying assumption that a classification model can only consider a single notion of centrality. Future research should consider models that combine distinct notions of centrality. We have seen that the ratio $N_S/N_M$ \iftoggle{arxiv}{(Tables}{(Supplementary Material, Tables} \ref{tab:summary_centrality_scores_UD} and \ref{tab:summary_centrality_scores_SUD}) yields an upper bound to precision (equation \ref{eq:precision_upper_bound}) and a low value of this ratio is an indication of a high proportion of false positives (Section \ref{subsec:evaluation}). A low value of the ratio $N_S/N_M$ is found in the worst classification models in terms of precision, i.e. the degree centrality model and the eccentricity model independently of the annotation style,
and is due to an excess of guesses per sentence. The problem of reducing the number of guesses of a classification model should be the subject of future research. Such a reduction can be achieved by combining distinct centrality criteria to reduce the number of tied vertices. We have paved the way for unsupervised machine learning methods that find the root vertex given the free tree structure and the positions of vertices.

\backmatter

\bmhead{Supplementary information}


\iftoggle{anonymous}{}
{

\bmhead{Acknowledgements}

We are very grateful to Á. Cancho-Victorio (1934-2023) for his inspiring discussions on first principles. This article is dedicated to his memory. We thank the editor and an anonymous reviewer for helpful comments to improve the article. We thank L. Alemany-Puig and Y. Yao for spotting multiple errors and inconsistencies in previous versions of the manuscript. We also thank L. Alemany-Puig for producing and making available the PUD/PSUD dataset in head vector format for the UD 2.14 release.

}

\section*{Declarations}


\begin{itemize}
\item Availability of data and materials.\\ 
The preprocessed treebanks analyzed during the current study are available in head vector format at 
\url{https://cqllab.upc.edu/lal/universal-dependencies/}. The original treebanks (UD 2.14) are available at \url{http://hdl.handle.net/11234/1-5502}.
\item Competing interests. \\ The authors declare that they have no competing interests.
\item Funding. \\ 
This research is supported by a recognition 2021SGR-Cat (01266 LQMC) from AGAUR (Generalitat de Catalunya) and the grant PID2024-155946NB-I00 funded by Ministerio de Ciencia, Innovación y Universidades (MICIU), Agencia Estatal de Investigación (AEI/10.13039/501100011033) and the European Social Fund Plus (ESF+).
\item Authors' contributions. \\ 
RFC conceived the research project, designed methods, performed the statistical analyses, wrote the original draft and revised the manuscript. MA supervised the research, contributed to the conceptualization and the design of methods and revised the manuscript. All authors read and approved the final manuscript.
\item Acknowledgements. \\ 
We are very grateful to Á. Cancho-Victorio (1934-2023) for his inspiring discussions on first principles. This article is dedicated to his memory. We thank the editor and an anonymous reviewer for helpful comments to improve the article. We thank L. Alemany-Puig and Y. Yao for spotting multiple errors and inconsistencies in previous versions of the manuscript. We also thank L. Alemany-Puig for producing and making available the PUD/PSUD dataset in head vector format for the UD 2.14 release.

\end{itemize}

\iftoggle{arxiv}{

\clearpage 

\begin{appendices}

\section{Ranking}
\label{appendix:ranking_summary}

\subsection{Discounted cumulative gain}

We also consider another ranking approach that we borrow from the field of information retrieval: discounted cumulative gain (DCG) \citep{Croft2010a}. The DCG of a list of $n$ documents retrieved is defined as \citep{Croft2010a}
\begin{equation*}
DCG = \sum_{i=1}^n \frac{\rho_i}{\log_2(i+1)},
\end{equation*}
where $i$ is the position of the document in the list and $\rho_i$ is the relevance of the $i$-th document selected. 
In our application, the documents correspond to the vertices of the tree, there is only one possible relevant vertex that is the root (a tree has only one root) and so the DCG becomes 
\begin{equation*}
DCG = \frac{1}{\log_2(i+1)},
\end{equation*}
where now $i$ is simply the average rank of the root in the sorting \iftoggle{arxiv}{(Equation \ref{eq:average_rank}).}{(main article, Equation \ref{eq:average_rank}).} If there are not tied values among vertices, then $i$ is simply the position of the root in the sorting.
DCG aims to give more importance to finding the root in top positions with respect to the plain definition of rank above. 

As DCGs from sentences of different length are not comparable, we transform them into numbers between 0 and 1 knowing that $1/\log_2 (n+1) \leq DCG \leq 1$. Then the performance of a score on a language is the average value of the normalized DCGs. The following property indicates that the average normalized DCG of the random baseline will never exceed 0.131. 
\begin{property}
\label{prop:ranking_DCG}
Let $\overline{DCG}$ be the normalized $DCG$, namely
\begin{align*}
\overline{DCG} & = \frac{DCG - DCG_{min}}{1 - DCG_{min}},
\end{align*}
where 
\begin{align*}
DCG & = 1/\log_2(r+1) \\
DCG_{min} & = 1/\log_2(n+1).
\end{align*}
Then the expectations according to a random baseline that picks a random vertex as root of a tree of size $n$ are
\begin{align*}
\mathbb{E}[DCG] & \leq \frac{1}{\log_2 \frac{n+3}{2}} \\
\mathbb{E}[\overline{DCG}] & \leq \frac{1}{1 - DCG_{min}}\left(\frac{1}{\log_2 \frac{n+3}{2}} - DCG_{min} \right). \\
                      & \leq 0.131.
\end{align*}
\end{property}
\begin{proof}
As the function $1/log_2(x + 1)$ is convex for $x > 1$ (Figure \ref{fig:expected_DCG_upper_bound}), Jensen's inequality yields 
\begin{equation*}
\mathbb{E}[DCG] \leq \frac{1}{\mathbb{E}[r+1]}.
\end{equation*}
Knowing that (recall Equation \ref{eq:average_rank}) \iftoggle{arxiv}{(recall Equation \ref{eq:average_rank})}{(recall main article, Equation \ref{eq:average_rank})}
$$\mathbb{E}[r+1] = r(2, n+1) = \frac{n+3}{2},$$
we obtain 
\begin{equation*}
\mathbb{E}[DCG] \leq \frac{1}{\log_2 \frac{n+3}{2}}.
\end{equation*}
By the linearity of expectation, 
\begin{equation*}
\label{eq:upper_bound}
\mathbb{E}[\overline{DCG}] \leq f(n) = \frac{1}{1 - DCG_{min}}\left(\frac{1}{\log_2 \frac{n+3}{2}} - \frac{1}{\log_2 (n+1)} \right).
\end{equation*}
When $n \geq 1$, $f(n)$ reaches a maximum at $n = 3$ (Figure \ref{fig:expected_DCG_upper_bound}). Hence 
$$\mathbb{E}[\overline{DCG}] \leq f(3) < 0.131.$$
\end{proof}

\begin{figure}
\centering
\includegraphics[width = 0.9 \textwidth]{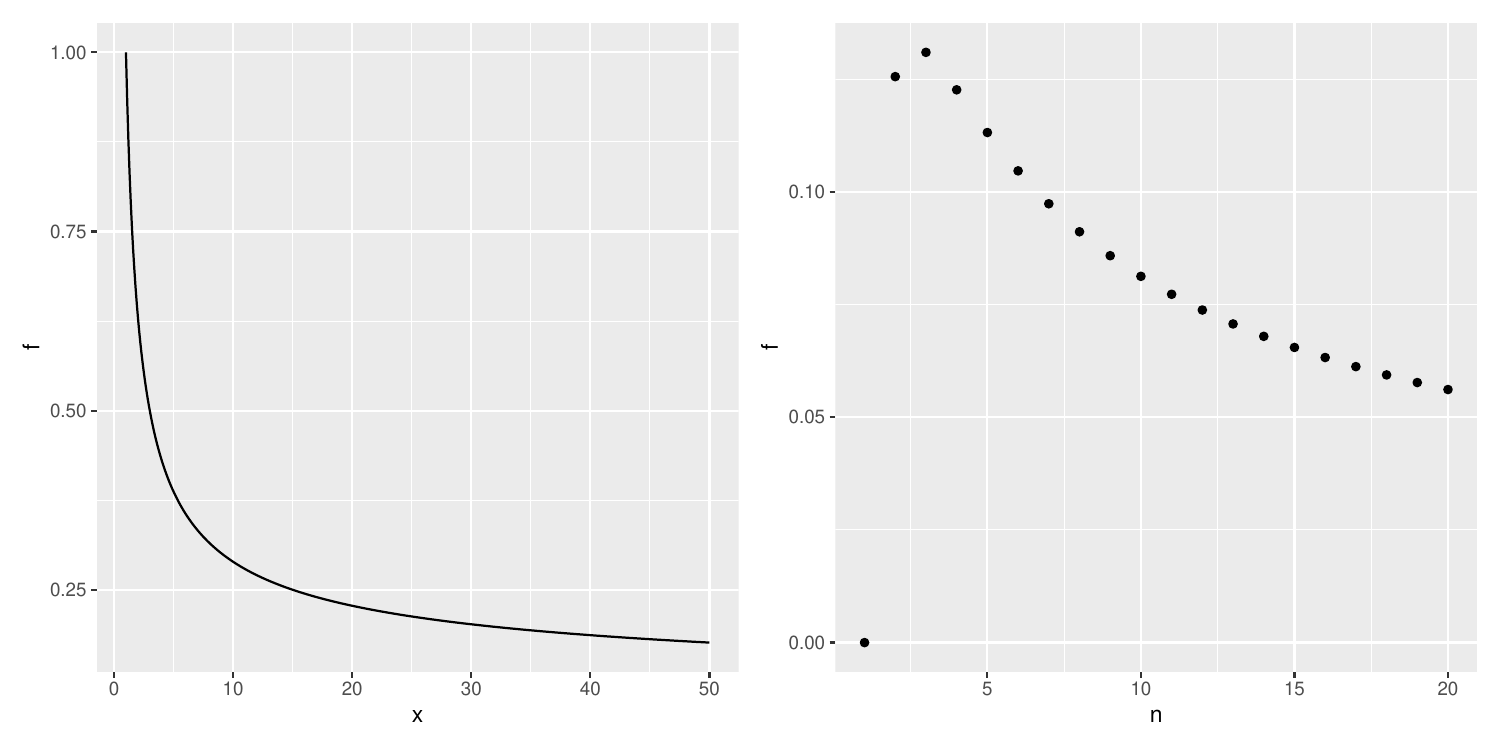}
\caption{\label{fig:expected_DCG_upper_bound} Left. The function $1/log_2(x + 1)$ for $x \geq 1$. Right. The function $f(n)$ (Equation \ref{eq:upper_bound}) for $n \geq 1$ }
\end{figure}

\subsection{Detailed results}

Figure \ref{fig:boxplot_centrality_scores_DCG} shows the  performance of the scores based on DCG for UD and SUD. 
Recall that higher DCG means higher ability to place the root in top positions (the opposite of plain ranks).
The mean normalized DCG is far from the upper bound predicted for the random baseline, that is 0.131 (Property \ref{prop:ranking_DCG}).

DCG supports the overall conclusion that all centrality scores tend to put the root vertex close to top positions and also that the new spatial centrality scores ($D$, $D'$ and coverage) are better suited (Figure \ref{fig:boxplot_centrality_scores_DCG}). With respect to the summary of results in \iftoggle{arxiv}{Section \ref{sec:results},}{Section \ref{sec:results} of the main article,} the only differences are (Figure \ref{fig:boxplot_centrality_scores_DCG})
\begin{enumerate}
\item[2.] The best scores are the new spatial scores both in UD and SUD. 
\item[3.] Among the non-spatial scores, degree centrality and eccentricity are the worst scores, both in UD and SUD. 
\end{enumerate} 
In addition, the distribution over languages is narrower (the violin plots for DCG in Figure \ref{fig:boxplot_centrality_scores_DCG} are wider than those for normalized rank) probably due to the  smoothing effect of the logarithmic correction of ranks performed by DCG. 
Besides, DCG confirms that eccentricity tends to perform worse than degree centrality (Figure \ref{fig:boxplot_centrality_scores_DCG}).
DCG also confirms the slight tendency of $D'$ to perform better than $D$ (Figure \ref{fig:boxplot_centrality_scores_DCG}). It also confirms that $m_2(v)$ is worse than $n_{max}(v)$ in UD and slightly better in SUD (Figure \ref{fig:boxplot_centrality_scores_DCG}).

\begin{figure}
\centering
\includegraphics[height = 0.85 \textheight]{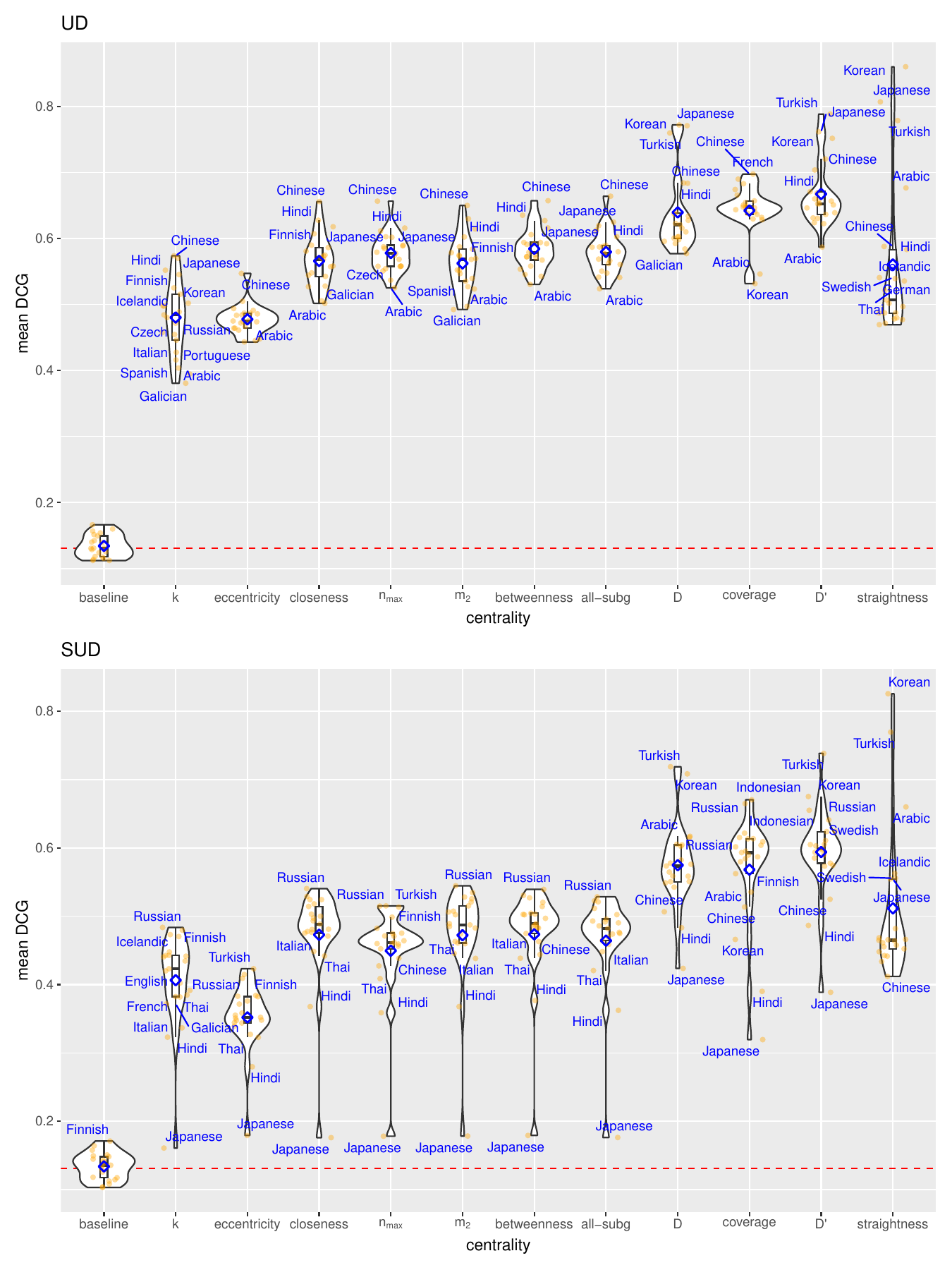}
\caption{\label{fig:boxplot_centrality_scores_DCG} 
The distribution of the DCG
(combined boxplot and violin plot) across languages for each centrality score when using UD (top) and SUD (bottom) annotation style. For each centrality score, black thick lines indicate medians while blue diamonds indicate means. 
The red dashed line indicates the upper bound of the expected value of DCG for the random baseline (Property \ref{prop:ranking_DCG}). 
}
\end{figure}

Tables \ref{tab:summary_centrality_scores_UD_ranking}
and \ref{tab:summary_centrality_scores_SUD_ranking} \iftoggle{arxiv}{}{in this document} summarize 
the distributions shown in  
\iftoggle{arxiv}{
Figures \ref{fig:boxplot_centrality_scores_ranking} and \ref{fig:boxplot_centrality_scores_DCG}, 
respectively. 
}
{Figure \ref{fig:boxplot_centrality_scores_ranking},
in the main text as well as Figure
\ref{fig:boxplot_centrality_scores_DCG} here, 
respectively. 
}
For the sake of completeness, Tables \ref{tab:complete_centrality_scores_UD_ranking} and \ref{tab:complete_centrality_scores_SUD_ranking} detail the performance of the centrality score on each language. 

\begin{table}
\caption{\label{tab:summary_centrality_scores_UD_ranking} The distribution of the performance of each centrality score across languages depending on the evaluation metrics (normalized rank or DCG) when UD annotation style is used. ``Aggregation'' indicates how, for each language, normalized rank or DCG are aggregated: by applying the mean or the median over all sentences. The distribution is described by the minimum value (min), the mean, the median, the maximum value (max) and the standard deviation (sd).}
\begin{tabular}{ccclllll}
evaluation & aggregation & centrality & min & mean & median & max & sd \\
\hline
rank & mean & $k$ & 0.092 & 0.138 & 0.146 & 0.19 & 0.028 \\
rank & mean & eccentricity & 0.127 & 0.167 & 0.163 & 0.209 & 0.021 \\
rank & mean & closeness & 0.07 & 0.113 & 0.11 & 0.15 & 0.022 \\
rank & mean & $n_{max}$ & 0.072 & 0.106 & 0.104 & 0.152 & 0.022 \\
rank & mean & $m_2$ & 0.074 & 0.115 & 0.116 & 0.156 & 0.023 \\
rank & mean & betweenness & 0.07 & 0.107 & 0.103 & 0.147 & 0.022 \\
rank & mean & all-subg & 0.068 & 0.107 & 0.104 & 0.151 & 0.022 \\
rank & mean & $D$ & 0.039 & 0.087 & 0.089 & 0.115 & 0.018 \\
rank & mean & coverage & 0.048 & 0.082 & 0.077 & 0.117 & 0.018 \\
rank & mean & $D'$ & 0.04 & 0.08 & 0.08 & 0.11 & 0.017 \\
rank & mean & straightness & 0.029 & 0.108 & 0.117 & 0.151 & 0.033 \\
rank & median & $k$ & 0.053 & 0.106 & 0.1 & 0.167 & 0.031 \\
rank & median & eccentricity & 0.05 & 0.104 & 0.1 & 0.133 & 0.02 \\
rank & median & closeness & 0.037 & 0.071 & 0.071 & 0.1 & 0.017 \\
rank & median & $n_{max}$ & 0.038 & 0.064 & 0.065 & 0.083 & 0.013 \\
rank & median & $m_2$ & 0.038 & 0.074 & 0.077 & 0.1 & 0.018 \\
rank & median & betweenness & 0.036 & 0.064 & 0.067 & 0.091 & 0.014 \\
rank & median & all-subg & 0.038 & 0.067 & 0.069 & 0.095 & 0.015 \\
rank & median & $D$ & 0 & 0.044 & 0.05 & 0.067 & 0.021 \\
rank & median & coverage & 0.025 & 0.042 & 0.04 & 0.062 & 0.011 \\
rank & median & $D'$ & 0 & 0.035 & 0.043 & 0.059 & 0.021 \\
rank & median & straightness & 0 & 0.08 & 0.091 & 0.125 & 0.04 \\
DCG & mean & $k$ & 0.381 & 0.48 & 0.484 & 0.574 & 0.056 \\
DCG & mean & eccentricity & 0.443 & 0.477 & 0.475 & 0.547 & 0.023 \\
DCG & mean & closeness & 0.502 & 0.566 & 0.569 & 0.656 & 0.04 \\
DCG & mean & $n_{max}$ & 0.525 & 0.578 & 0.577 & 0.656 & 0.03 \\
DCG & mean & $m_2$ & 0.492 & 0.562 & 0.568 & 0.65 & 0.042 \\
DCG & mean & betweenness & 0.53 & 0.584 & 0.579 & 0.657 & 0.032 \\
DCG & mean & all-subg & 0.524 & 0.579 & 0.577 & 0.664 & 0.033 \\
DCG & mean & $D$ & 0.577 & 0.639 & 0.621 & 0.772 & 0.06 \\
DCG & mean & coverage & 0.531 & 0.642 & 0.647 & 0.697 & 0.039 \\
DCG & mean & $D'$ & 0.587 & 0.667 & 0.652 & 0.788 & 0.05 \\
DCG & mean & straightness & 0.469 & 0.561 & 0.507 & 0.86 & 0.118 \\
DCG & median & $k$ & 0.275 & 0.397 & 0.4 & 0.515 & 0.068 \\
DCG & median & eccentricity & 0.308 & 0.381 & 0.353 & 0.662 & 0.088 \\
DCG & median & closeness & 0.38 & 0.494 & 0.508 & 0.538 & 0.04 \\
DCG & median & $n_{max}$ & 0.472 & 0.517 & 0.511 & 0.644 & 0.032 \\
DCG & median & $m_2$ & 0.37 & 0.488 & 0.504 & 0.535 & 0.048 \\
DCG & median & betweenness & 0.481 & 0.512 & 0.511 & 0.536 & 0.014 \\
DCG & median & all-subg & 0.481 & 0.509 & 0.508 & 0.539 & 0.014 \\
DCG & median & $D$ & 0.511 & 0.599 & 0.526 & 1 & 0.17 \\
DCG & median & coverage & 0.515 & 0.605 & 0.622 & 0.686 & 0.058 \\
DCG & median & $D'$ & 0.517 & 0.641 & 0.531 & 1 & 0.206 \\
DCG & median & straightness & 0.355 & 0.507 & 0.446 & 1 & 0.218 \\

\end{tabular}
\end{table}

\begin{table}
\caption{\label{tab:summary_centrality_scores_SUD_ranking} The distribution of the performance of each centrality score across languages when SUD annotation style is used. The format is the same as in Table \ref{tab:summary_centrality_scores_UD_ranking}. }
\begin{tabular}{ccclllll}
evaluation & aggregation & centrality & min & mean & median & max & sd \\
\hline
rank & mean & $k$ & 0.153 & 0.207 & 0.194 & 0.419 & 0.059 \\
rank & mean & eccentricity & 0.218 & 0.263 & 0.258 & 0.395 & 0.037 \\
rank & mean & closeness & 0.146 & 0.193 & 0.177 & 0.407 & 0.057 \\
rank & mean & $n_{max}$ & 0.151 & 0.199 & 0.18 & 0.42 & 0.059 \\
rank & mean & $m_2$ & 0.142 & 0.189 & 0.166 & 0.416 & 0.062 \\
rank & mean & betweenness & 0.146 & 0.193 & 0.174 & 0.419 & 0.06 \\
rank & mean & all-subg & 0.149 & 0.197 & 0.179 & 0.407 & 0.056 \\
rank & mean & $D$ & 0.084 & 0.143 & 0.129 & 0.257 & 0.037 \\
rank & mean & coverage & 0.103 & 0.143 & 0.126 & 0.283 & 0.044 \\
rank & mean & $D'$ & 0.085 & 0.139 & 0.126 & 0.268 & 0.039 \\
rank & mean & straightness & 0.055 & 0.142 & 0.144 & 0.197 & 0.039 \\
rank & median & $k$ & 0.1 & 0.153 & 0.132 & 0.423 & 0.081 \\
rank & median & eccentricity & 0.167 & 0.202 & 0.19 & 0.382 & 0.047 \\
rank & median & closeness & 0.077 & 0.118 & 0.1 & 0.385 & 0.066 \\
rank & median & $n_{max}$ & 0.1 & 0.138 & 0.125 & 0.397 & 0.063 \\
rank & median & $m_2$ & 0.079 & 0.121 & 0.1 & 0.389 & 0.068 \\
rank & median & betweenness & 0.083 & 0.124 & 0.1 & 0.397 & 0.067 \\
rank & median & all-subg & 0.1 & 0.133 & 0.115 & 0.388 & 0.064 \\
rank & median & $D$ & 0 & 0.058 & 0.056 & 0.103 & 0.025 \\
rank & median & coverage & 0.026 & 0.056 & 0.05 & 0.13 & 0.027 \\
rank & median & $D'$ & 0 & 0.052 & 0.048 & 0.12 & 0.027 \\
rank & median & straightness & 0 & 0.091 & 0.1 & 0.15 & 0.04 \\
DCG & mean & $k$ & 0.161 & 0.406 & 0.423 & 0.484 & 0.072 \\
DCG & mean & eccentricity & 0.179 & 0.352 & 0.352 & 0.423 & 0.052 \\
DCG & mean & closeness & 0.176 & 0.473 & 0.489 & 0.541 & 0.078 \\
DCG & mean & $n_{max}$ & 0.178 & 0.45 & 0.462 & 0.515 & 0.072 \\
DCG & mean & $m_2$ & 0.178 & 0.472 & 0.487 & 0.545 & 0.078 \\
DCG & mean & betweenness & 0.179 & 0.474 & 0.489 & 0.539 & 0.077 \\
DCG & mean & all-subg & 0.176 & 0.464 & 0.482 & 0.528 & 0.076 \\
DCG & mean & $D$ & 0.424 & 0.575 & 0.574 & 0.719 & 0.065 \\
DCG & mean & coverage & 0.319 & 0.568 & 0.593 & 0.67 & 0.085 \\
DCG & mean & $D'$ & 0.389 & 0.594 & 0.599 & 0.738 & 0.069 \\
DCG & mean & straightness & 0.412 & 0.512 & 0.465 & 0.826 & 0.111 \\
DCG & median & $k$ & 0.084 & 0.313 & 0.333 & 0.412 & 0.079 \\
DCG & median & eccentricity & 0.101 & 0.241 & 0.242 & 0.306 & 0.044 \\
DCG & median & closeness & 0.096 & 0.381 & 0.372 & 0.499 & 0.096 \\
DCG & median & $n_{max}$ & 0.093 & 0.333 & 0.342 & 0.424 & 0.07 \\
DCG & median & $m_2$ & 0.099 & 0.383 & 0.369 & 0.494 & 0.096 \\
DCG & median & betweenness & 0.094 & 0.372 & 0.372 & 0.488 & 0.09 \\
DCG & median & all-subg & 0.094 & 0.346 & 0.355 & 0.481 & 0.08 \\
DCG & median & $D$ & 0.321 & 0.547 & 0.524 & 1 & 0.16 \\
DCG & median & coverage & 0.274 & 0.553 & 0.536 & 0.683 & 0.107 \\
DCG & median & $D'$ & 0.283 & 0.555 & 0.526 & 1 & 0.163 \\
DCG & median & straightness & 0.315 & 0.448 & 0.365 & 1 & 0.194 \\

\end{tabular}
\end{table}

\begin{longtable}[c]{ccllll}
 \caption{\label{tab:complete_centrality_scores_UD_ranking} The performance of each centrality score for each language in the PUD treebank using UD annotation style. For the rank and the DCG of the root, we show the mean and the median over all sentences of the language. Rank is the normalized rank. For each language, evaluation metric (rank or DCG) and aggregation (mean or media), the best score is marked with boldface and underline whereas the 2nd best score is marked just with boldface. }\\ 

 \hline
          &            & \multicolumn{2}{c}{rank} & \multicolumn{2}{c}{DCG} \\
 \cmidrule{3-4} \cmidrule{5-6} 
 language & centrality & mean & median & mean & median \\ 
 \hline
 \endfirsthead

 \hline
          &            & \multicolumn{2}{c}{rank} & \multicolumn{2}{c}{DCG} \\
 \cmidrule{3-4} \cmidrule{5-6} 
 language & centrality & rank & median & mean & median \\  
 \hline
 \endhead

 \hline
 \endfoot


 Arabic & $k$ & 0.19 & 0.156 & 0.395 & 0.297\\
Arabic & eccentricity & 0.198 & 0.132 & 0.443 & 0.308\\
Arabic & closeness & 0.15 & 0.091 & 0.502 & 0.427\\
Arabic & $n_{max}$ & 0.141 & 0.079 & 0.525 & 0.488\\
Arabic & $m_2$ & 0.156 & 0.1 & 0.497 & 0.382\\
Arabic & betweenness & 0.143 & 0.083 & 0.53 & 0.488\\
Arabic & all-subg & 0.142 & 0.087 & 0.524 & 0.488\\
Arabic & $D$ & \textbf{0.109} & \textbf{0.059} & \textbf{0.592} & \textbf{0.52}\\
Arabic & coverage & 0.117 & 0.062 & 0.546 & 0.515\\
Arabic & $D'$ & 0.11 & \textbf{0.059} & 0.587 & 0.517\\
Arabic & straightness & \underline{\textbf{0.081}} & \underline{\textbf{0.026}} & \underline{\textbf{0.676}} & \underline{\textbf{0.635}}\\
Chinese & $k$ & 0.109 & 0.062 & 0.574 & 0.515\\
Chinese & eccentricity & 0.127 & 0.05 & 0.547 & 0.662\\
Chinese & closeness & 0.085 & 0.042 & 0.656 & 0.538\\
Chinese & $n_{max}$ & 0.083 & 0.038 & 0.656 & 0.644\\
Chinese & $m_2$ & 0.089 & 0.042 & 0.65 & 0.534\\
Chinese & betweenness & 0.083 & 0.04 & 0.657 & 0.535\\
Chinese & all-subg & 0.081 & 0.038 & 0.664 & 0.539\\
Chinese & $D$ & 0.071 & 0.033 & 0.684 & 0.657\\
Chinese & coverage & \textbf{0.064} & \textbf{0.031} & \textbf{0.697} & \textbf{0.673}\\
Chinese & $D'$ & \underline{\textbf{0.063}} & \underline{\textbf{0}} & \underline{\textbf{0.72}} & \underline{\textbf{1}}\\
Chinese & straightness & 0.094 & 0.071 & 0.588 & 0.499\\
Czech & $k$ & 0.168 & 0.136 & 0.446 & 0.361\\
Czech & eccentricity & 0.196 & 0.125 & 0.461 & 0.338\\
Czech & closeness & 0.146 & 0.1 & 0.528 & 0.461\\
Czech & $n_{max}$ & 0.136 & 0.083 & 0.537 & 0.472\\
Czech & $m_2$ & 0.144 & 0.1 & 0.527 & 0.472\\
Czech & betweenness & 0.136 & 0.091 & 0.545 & 0.481\\
Czech & all-subg & 0.139 & 0.095 & 0.541 & 0.481\\
Czech & $D$ & 0.115 & 0.067 & 0.584 & 0.511\\
Czech & coverage & \underline{\textbf{0.1}} & \underline{\textbf{0.045}} & \underline{\textbf{0.64}} & \underline{\textbf{0.622}}\\
Czech & $D'$ & \textbf{0.105} & \textbf{0.059} & \textbf{0.624} & \textbf{0.517}\\
Czech & straightness & 0.146 & 0.125 & 0.469 & 0.363\\
English & $k$ & 0.123 & 0.1 & 0.484 & 0.4\\
English & eccentricity & 0.157 & 0.1 & 0.487 & 0.367\\
English & closeness & 0.099 & 0.071 & 0.569 & 0.508\\
English & $n_{max}$ & 0.091 & 0.06 & 0.587 & 0.515\\
English & $m_2$ & 0.099 & 0.071 & 0.568 & 0.504\\
English & betweenness & 0.092 & 0.059 & 0.592 & 0.52\\
English & all-subg & 0.092 & 0.062 & 0.587 & 0.515\\
English & $D$ & 0.089 & 0.053 & 0.602 & 0.522\\
English & coverage & \underline{\textbf{0.071}} & \underline{\textbf{0.04}} & \underline{\textbf{0.657}} & \underline{\textbf{0.622}}\\
English & $D'$ & \textbf{0.075} & \textbf{0.045} & \textbf{0.651} & \textbf{0.53}\\
English & straightness & 0.125 & 0.111 & 0.478 & 0.355\\
Finnish & $k$ & 0.122 & 0.091 & 0.552 & 0.488\\
Finnish & eccentricity & 0.176 & 0.115 & 0.494 & 0.404\\
Finnish & closeness & 0.11 & 0.077 & 0.607 & 0.508\\
Finnish & $n_{max}$ & 0.111 & 0.071 & 0.602 & 0.511\\
Finnish & $m_2$ & 0.11 & 0.083 & 0.602 & 0.504\\
Finnish & betweenness & 0.108 & 0.071 & 0.616 & 0.511\\
Finnish & all-subg & 0.11 & 0.071 & 0.612 & 0.511\\
Finnish & $D$ & 0.103 & 0.062 & 0.627 & 0.517\\
Finnish & coverage & \textbf{0.097} & \textbf{0.056} & \textbf{0.647} & \underline{\textbf{0.622}}\\
Finnish & $D'$ & \underline{\textbf{0.096}} & \underline{\textbf{0.053}} & \underline{\textbf{0.659}} & \textbf{0.524}\\
Finnish & straightness & 0.145 & 0.125 & 0.502 & 0.446\\
French & $k$ & 0.115 & 0.087 & 0.48 & 0.4\\
French & eccentricity & 0.147 & 0.086 & 0.484 & 0.365\\
French & closeness & 0.09 & 0.056 & 0.581 & 0.517\\
French & $n_{max}$ & 0.083 & 0.054 & 0.59 & 0.52\\
French & $m_2$ & 0.092 & 0.059 & 0.573 & 0.515\\
French & betweenness & 0.083 & 0.051 & 0.594 & 0.522\\
French & all-subg & 0.085 & 0.056 & 0.59 & 0.517\\
French & $D$ & 0.069 & 0.04 & 0.639 & 0.534\\
French & coverage & \underline{\textbf{0.056}} & \underline{\textbf{0.025}} & \underline{\textbf{0.69}} & \underline{\textbf{0.686}}\\
French & $D'$ & \textbf{0.062} & \textbf{0.032} & \textbf{0.671} & \textbf{0.54}\\
French & straightness & 0.111 & 0.094 & 0.48 & 0.363\\
Galician & $k$ & 0.173 & 0.167 & 0.381 & 0.275\\
Galician & eccentricity & 0.163 & 0.091 & 0.476 & 0.353\\
Galician & closeness & 0.126 & 0.083 & 0.505 & 0.38\\
Galician & $n_{max}$ & 0.111 & 0.065 & 0.551 & 0.508\\
Galician & $m_2$ & 0.135 & 0.097 & 0.492 & 0.37\\
Galician & betweenness & 0.116 & 0.069 & 0.543 & 0.494\\
Galician & all-subg & 0.113 & 0.071 & 0.541 & 0.494\\
Galician & $D$ & 0.1 & 0.053 & 0.577 & 0.522\\
Galician & coverage & \underline{\textbf{0.081}} & \underline{\textbf{0.038}} & \underline{\textbf{0.628}} & \underline{\textbf{0.54}}\\
Galician & $D'$ & \textbf{0.088} & \textbf{0.042} & \textbf{0.622} & \textbf{0.531}\\
Galician & straightness & 0.12 & 0.091 & 0.497 & 0.374\\
German & $k$ & 0.116 & 0.089 & 0.498 & 0.412\\
German & eccentricity & 0.145 & 0.083 & 0.505 & \underline{\textbf{0.603}}\\
German & closeness & 0.092 & 0.059 & 0.586 & 0.517\\
German & $n_{max}$ & 0.083 & 0.056 & 0.602 & 0.52\\
German & $m_2$ & 0.094 & 0.062 & 0.584 & 0.515\\
German & betweenness & 0.084 & 0.059 & 0.606 & 0.517\\
German & all-subg & 0.084 & 0.059 & 0.602 & 0.517\\
German & $D$ & 0.08 & 0.047 & 0.634 & 0.528\\
German & coverage & \textbf{0.075} & \textbf{0.042} & \textbf{0.65} & \underline{\textbf{0.603}}\\
German & $D'$ & \underline{\textbf{0.072}} & \underline{\textbf{0.04}} & \underline{\textbf{0.668}} & \textbf{0.533}\\
German & straightness & 0.101 & 0.08 & 0.536 & 0.494\\
Hindi & $k$ & 0.092 & 0.053 & 0.57 & 0.515\\
Hindi & eccentricity & 0.16 & 0.094 & 0.475 & 0.349\\
Hindi & closeness & 0.087 & 0.04 & 0.627 & 0.534\\
Hindi & $n_{max}$ & 0.08 & 0.043 & 0.616 & 0.533\\
Hindi & $m_2$ & 0.078 & 0.043 & 0.63 & 0.531\\
Hindi & betweenness & 0.078 & 0.04 & 0.635 & 0.534\\
Hindi & all-subg & 0.082 & 0.043 & 0.619 & 0.53\\
Hindi & $D$ & \underline{\textbf{0.065}} & \textbf{0.029} & \textbf{0.684} & 0.548\\
Hindi & coverage & 0.079 & 0.038 & 0.632 & \textbf{0.652}\\
Hindi & $D'$ & \textbf{0.065} & \underline{\textbf{0}} & \underline{\textbf{0.704}} & \underline{\textbf{1}}\\
Hindi & straightness & 0.109 & 0.062 & 0.583 & 0.508\\
Icelandic & $k$ & 0.113 & 0.083 & 0.525 & 0.461\\
Icelandic & eccentricity & 0.176 & 0.121 & 0.472 & 0.347\\
Icelandic & closeness & 0.102 & 0.067 & 0.59 & 0.515\\
Icelandic & $n_{max}$ & 0.1 & 0.071 & 0.585 & 0.515\\
Icelandic & $m_2$ & 0.098 & 0.071 & 0.587 & 0.511\\
Icelandic & betweenness & 0.099 & 0.067 & 0.591 & 0.515\\
Icelandic & all-subg & 0.1 & 0.067 & 0.588 & 0.511\\
Icelandic & $D$ & 0.084 & \textbf{0.05} & 0.637 & \textbf{0.53}\\
Icelandic & coverage & \underline{\textbf{0.076}} & \underline{\textbf{0.043}} & \textbf{0.648} & \underline{\textbf{0.603}}\\
Icelandic & $D'$ & \textbf{0.078} & \underline{\textbf{0.043}} & \underline{\textbf{0.656}} & \textbf{0.53}\\
Icelandic & straightness & 0.121 & 0.095 & 0.539 & 0.472\\
Indonesian & $k$ & 0.153 & 0.133 & 0.454 & 0.361\\
Indonesian & eccentricity & 0.165 & 0.1 & 0.484 & 0.378\\
Indonesian & closeness & 0.124 & 0.083 & 0.542 & 0.494\\
Indonesian & $n_{max}$ & 0.109 & 0.069 & 0.577 & 0.515\\
Indonesian & $m_2$ & 0.124 & 0.083 & 0.539 & 0.494\\
Indonesian & betweenness & 0.111 & 0.071 & 0.576 & 0.508\\
Indonesian & all-subg & 0.111 & 0.071 & 0.577 & 0.504\\
Indonesian & $D$ & 0.098 & 0.059 & 0.61 & 0.522\\
Indonesian & coverage & \underline{\textbf{0.081}} & \underline{\textbf{0.031}} & \underline{\textbf{0.674}} & \underline{\textbf{0.678}}\\
Indonesian & $D'$ & \textbf{0.088} & \textbf{0.045} & \textbf{0.652} & \textbf{0.528}\\
Indonesian & straightness & 0.135 & 0.111 & 0.487 & 0.361\\
Italian & $k$ & 0.146 & 0.115 & 0.416 & 0.322\\
Italian & eccentricity & 0.159 & 0.1 & 0.467 & 0.328\\
Italian & closeness & 0.113 & 0.077 & 0.527 & 0.481\\
Italian & $n_{max}$ & 0.101 & 0.062 & 0.554 & 0.508\\
Italian & $m_2$ & 0.116 & 0.077 & 0.523 & 0.481\\
Italian & betweenness & 0.102 & 0.067 & 0.558 & 0.504\\
Italian & all-subg & 0.103 & 0.069 & 0.551 & 0.504\\
Italian & $D$ & 0.091 & 0.05 & 0.595 & 0.526\\
Italian & coverage & \underline{\textbf{0.075}} & \underline{\textbf{0.038}} & \underline{\textbf{0.643}} & \underline{\textbf{0.546}}\\
Italian & $D'$ & \textbf{0.081} & \textbf{0.043} & \textbf{0.63} & \textbf{0.531}\\
Italian & straightness & 0.117 & 0.1 & 0.487 & 0.361\\
Japanese & $k$ & 0.092 & 0.053 & 0.545 & 0.462\\
Japanese & eccentricity & 0.138 & 0.095 & 0.464 & 0.329\\
Japanese & closeness & 0.07 & 0.037 & 0.618 & 0.535\\
Japanese & $n_{max}$ & 0.072 & 0.038 & 0.611 & 0.535\\
Japanese & $m_2$ & 0.074 & 0.038 & 0.614 & 0.535\\
Japanese & betweenness & 0.07 & 0.036 & 0.627 & 0.536\\
Japanese & all-subg & 0.068 & 0.038 & 0.625 & 0.535\\
Japanese & $D$ & \textbf{0.039} & \underline{\textbf{0}} & \textbf{0.771} & \underline{\textbf{1}}\\
Japanese & coverage & 0.048 & \textbf{0.029} & 0.666 & \textbf{0.547}\\
Japanese & $D'$ & 0.04 & \underline{\textbf{0}} & 0.761 & \underline{\textbf{1}}\\
Japanese & straightness & \underline{\textbf{0.029}} & \underline{\textbf{0}} & \underline{\textbf{0.807}} & \underline{\textbf{1}}\\
Korean & $k$ & 0.163 & 0.094 & 0.515 & 0.433\\
Korean & eccentricity & 0.209 & 0.133 & 0.448 & 0.347\\
Korean & closeness & 0.15 & 0.071 & 0.573 & 0.508\\
Korean & $n_{max}$ & 0.152 & 0.077 & 0.558 & 0.508\\
Korean & $m_2$ & 0.147 & 0.071 & 0.575 & 0.511\\
Korean & betweenness & 0.147 & 0.071 & 0.579 & 0.511\\
Korean & all-subg & 0.151 & 0.077 & 0.577 & 0.508\\
Korean & $D$ & \textbf{0.076} & \underline{\textbf{0}} & \textbf{0.772} & \underline{\textbf{1}}\\
Korean & coverage & 0.117 & \textbf{0.062} & 0.531 & \textbf{0.52}\\
Korean & $D'$ & 0.081 & \underline{\textbf{0}} & 0.751 & \underline{\textbf{1}}\\
Korean & straightness & \underline{\textbf{0.044}} & \underline{\textbf{0}} & \underline{\textbf{0.86}} & \underline{\textbf{1}}\\
Polish & $k$ & 0.165 & 0.125 & 0.472 & 0.394\\
Polish & eccentricity & 0.192 & 0.125 & 0.464 & 0.342\\
Polish & closeness & 0.139 & 0.083 & 0.557 & 0.499\\
Polish & $n_{max}$ & 0.136 & 0.077 & 0.558 & 0.499\\
Polish & $m_2$ & 0.142 & 0.083 & 0.552 & 0.499\\
Polish & betweenness & 0.135 & 0.077 & 0.572 & 0.504\\
Polish & all-subg & 0.134 & 0.083 & 0.566 & 0.499\\
Polish & $D$ & 0.109 & 0.062 & 0.616 & 0.52\\
Polish & coverage & \underline{\textbf{0.101}} & \underline{\textbf{0.05}} & \textbf{0.63} & \underline{\textbf{0.572}}\\
Polish & $D'$ & \textbf{0.103} & \textbf{0.056} & \underline{\textbf{0.636}} & \textbf{0.522}\\
Polish & straightness & 0.137 & 0.107 & 0.507 & 0.398\\
Portuguese & $k$ & 0.147 & 0.121 & 0.428 & 0.322\\
Portuguese & eccentricity & 0.148 & 0.086 & 0.491 & 0.4\\
Portuguese & closeness & 0.109 & 0.071 & 0.544 & 0.504\\
Portuguese & $n_{max}$ & 0.097 & 0.059 & 0.574 & 0.52\\
Portuguese & $m_2$ & 0.116 & 0.077 & 0.535 & 0.494\\
Portuguese & betweenness & 0.1 & 0.062 & 0.572 & 0.515\\
Portuguese & all-subg & 0.099 & 0.062 & 0.569 & 0.515\\
Portuguese & $D$ & 0.091 & 0.048 & 0.603 & 0.526\\
Portuguese & coverage & \underline{\textbf{0.075}} & \underline{\textbf{0.034}} & \underline{\textbf{0.654}} & \underline{\textbf{0.635}}\\
Portuguese & $D'$ & \textbf{0.08} & \textbf{0.04} & \textbf{0.646} & \textbf{0.534}\\
Portuguese & straightness & 0.118 & 0.097 & 0.477 & 0.363\\
Russian & $k$ & 0.152 & 0.125 & 0.458 & 0.388\\
Russian & eccentricity & 0.178 & 0.125 & 0.474 & 0.361\\
Russian & closeness & 0.126 & 0.083 & 0.548 & 0.494\\
Russian & $n_{max}$ & 0.117 & 0.071 & 0.568 & 0.508\\
Russian & $m_2$ & 0.125 & 0.083 & 0.547 & 0.494\\
Russian & betweenness & 0.117 & 0.075 & 0.571 & 0.504\\
Russian & all-subg & 0.12 & 0.077 & 0.562 & 0.499\\
Russian & $D$ & 0.107 & 0.06 & 0.6 & 0.517\\
Russian & coverage & \underline{\textbf{0.092}} & \underline{\textbf{0.043}} & \underline{\textbf{0.646}} & \underline{\textbf{0.657}}\\
Russian & $D'$ & \textbf{0.097} & \textbf{0.053} & \textbf{0.639} & \textbf{0.524}\\
Russian & straightness & 0.151 & 0.125 & 0.471 & 0.355\\
Spanish & $k$ & 0.16 & 0.138 & 0.404 & 0.307\\
Spanish & eccentricity & 0.159 & 0.095 & 0.474 & 0.349\\
Spanish & closeness & 0.119 & 0.083 & 0.519 & 0.422\\
Spanish & $n_{max}$ & 0.106 & 0.062 & 0.558 & 0.508\\
Spanish & $m_2$ & 0.127 & 0.091 & 0.51 & 0.388\\
Spanish & betweenness & 0.11 & 0.067 & 0.553 & 0.508\\
Spanish & all-subg & 0.108 & 0.067 & 0.548 & 0.499\\
Spanish & $D$ & 0.094 & 0.05 & 0.598 & 0.524\\
Spanish & coverage & \underline{\textbf{0.077}} & \underline{\textbf{0.038}} & \underline{\textbf{0.638}} & \underline{\textbf{0.548}}\\
Spanish & $D'$ & \textbf{0.084} & \textbf{0.043} & \textbf{0.631} & \textbf{0.531}\\
Spanish & straightness & 0.119 & 0.091 & 0.487 & 0.365\\
Swedish & $k$ & 0.122 & 0.096 & 0.502 & 0.416\\
Swedish & eccentricity & 0.163 & 0.1 & 0.486 & 0.363\\
Swedish & closeness & 0.101 & 0.071 & 0.571 & 0.504\\
Swedish & $n_{max}$ & 0.094 & 0.071 & 0.589 & 0.508\\
Swedish & $m_2$ & 0.103 & 0.077 & 0.57 & 0.499\\
Swedish & betweenness & 0.094 & 0.067 & 0.594 & 0.511\\
Swedish & all-subg & 0.096 & 0.071 & 0.587 & 0.504\\
Swedish & $D$ & 0.088 & 0.056 & 0.621 & 0.522\\
Swedish & coverage & \underline{\textbf{0.077}} & \textbf{0.048} & \textbf{0.636} & \underline{\textbf{0.534}}\\
Swedish & $D'$ & \textbf{0.079} & \underline{\textbf{0.045}} & \underline{\textbf{0.652}} & \textbf{0.528}\\
Swedish & straightness & 0.115 & 0.091 & 0.541 & 0.481\\
Thai & $k$ & 0.128 & 0.083 & 0.489 & 0.412\\
Thai & eccentricity & 0.17 & 0.114 & 0.446 & 0.321\\
Thai & closeness & 0.107 & 0.059 & 0.558 & 0.515\\
Thai & $n_{max}$ & 0.104 & 0.057 & 0.557 & 0.515\\
Thai & $m_2$ & 0.108 & 0.062 & 0.552 & 0.511\\
Thai & betweenness & 0.103 & 0.059 & 0.567 & 0.515\\
Thai & all-subg & 0.104 & 0.059 & 0.561 & 0.515\\
Thai & $D$ & 0.078 & 0.042 & 0.623 & 0.533\\
Thai & coverage & \underline{\textbf{0.066}} & \underline{\textbf{0.026}} & \underline{\textbf{0.683}} & \underline{\textbf{0.683}}\\
Thai & $D'$ & \textbf{0.071} & \textbf{0.034} & \textbf{0.66} & \textbf{0.539}\\
Thai & straightness & 0.103 & 0.074 & 0.521 & 0.461\\
Turkish & $k$ & 0.155 & 0.111 & 0.498 & 0.404\\
Turkish & eccentricity & 0.176 & 0.115 & 0.487 & 0.394\\
Turkish & closeness & 0.129 & 0.077 & 0.575 & 0.508\\
Turkish & $n_{max}$ & 0.129 & 0.071 & 0.579 & 0.511\\
Turkish & $m_2$ & 0.131 & 0.077 & 0.578 & 0.508\\
Turkish & betweenness & 0.127 & 0.071 & 0.592 & 0.511\\
Turkish & all-subg & 0.128 & 0.077 & 0.58 & 0.504\\
Turkish & $D$ & 0.066 & \underline{\textbf{0}} & 0.76 & \underline{\textbf{1}}\\
Turkish & coverage & 0.087 & \textbf{0.05} & 0.648 & \textbf{0.657}\\
Turkish & $D'$ & \textbf{0.063} & \underline{\textbf{0}} & \underline{\textbf{0.788}} & \underline{\textbf{1}}\\
Turkish & straightness & \underline{\textbf{0.055}} & \underline{\textbf{0}} & \textbf{0.778} & \underline{\textbf{1}}\\

\end{longtable}

\clearpage

\begin{longtable}[c]{ccllllll} \caption{\label{tab:complete_centrality_scores_SUD_ranking} The performance of each centrality score for each language in the PUD treebank using SUD annotation style. The format is the same as in Table \ref{tab:complete_centrality_scores_UD_ranking}.}\\ 

 \toprule
           &            & \multicolumn{2}{c}{rank} & \multicolumn{2}{c}{DCG} \\
 \cmidrule{3-4} \cmidrule{5-6} 
 language & centrality & mean & median & mean & median \\ 
 \midrule
 \endfirsthead

 \toprule
           &            & \multicolumn{2}{c}{rank} & \multicolumn{2}{c}{DCG} \\
 \cmidrule{3-4} \cmidrule{5-6} 
 language & centrality & mean & median & mean & median \\  
 \midrule
 \endhead

 \botrule
 \endfoot


 Arabic & $k$ & 0.196 & 0.111 & 0.441 & 0.355\\
Arabic & eccentricity & 0.267 & 0.2 & 0.348 & 0.244\\
Arabic & closeness & 0.189 & 0.1 & 0.497 & 0.38\\
Arabic & $n_{max}$ & 0.198 & 0.125 & 0.46 & 0.338\\
Arabic & $m_2$ & 0.179 & 0.091 & 0.507 & 0.431\\
Arabic & betweenness & 0.189 & 0.1 & 0.502 & 0.374\\
Arabic & all-subg & 0.199 & 0.118 & 0.475 & 0.347\\
Arabic & $D$ & \textbf{0.123} & \textbf{0.047} & \textbf{0.617} & \underline{\textbf{0.536}}\\
Arabic & coverage & 0.15 & 0.071 & 0.535 & 0.511\\
Arabic & $D'$ & 0.131 & 0.053 & 0.603 & 0.524\\
Arabic & straightness & \underline{\textbf{0.089}} & \underline{\textbf{0.038}} & \underline{\textbf{0.66}} & \textbf{0.535}\\
Chinese & $k$ & 0.232 & 0.136 & 0.42 & 0.302\\
Chinese & eccentricity & 0.25 & 0.196 & 0.339 & 0.235\\
Chinese & closeness & \textbf{0.189} & 0.125 & 0.475 & 0.346\\
Chinese & $n_{max}$ & 0.235 & 0.147 & 0.427 & 0.315\\
Chinese & $m_2$ & 0.222 & 0.129 & 0.46 & 0.342\\
Chinese & betweenness & 0.229 & 0.132 & 0.444 & 0.338\\
Chinese & all-subg & 0.205 & 0.133 & 0.459 & 0.342\\
Chinese & $D$ & 0.191 & 0.091 & 0.507 & 0.461\\
Chinese & coverage & 0.192 & \underline{\textbf{0.071}} & \textbf{0.514} & \underline{\textbf{0.508}}\\
Chinese & $D'$ & \underline{\textbf{0.189}} & \textbf{0.077} & \underline{\textbf{0.525}} & \textbf{0.494}\\
Chinese & straightness & 0.197 & 0.143 & 0.412 & 0.315\\
Czech & $k$ & 0.205 & 0.132 & 0.423 & 0.333\\
Czech & eccentricity & 0.268 & 0.208 & 0.371 & 0.255\\
Czech & closeness & 0.202 & 0.118 & 0.478 & 0.358\\
Czech & $n_{max}$ & 0.201 & 0.147 & 0.456 & 0.33\\
Czech & $m_2$ & 0.188 & 0.118 & 0.487 & 0.361\\
Czech & betweenness & 0.195 & 0.133 & 0.483 & 0.349\\
Czech & all-subg & 0.208 & 0.143 & 0.464 & 0.333\\
Czech & $D$ & 0.153 & 0.071 & 0.559 & 0.511\\
Czech & coverage & \underline{\textbf{0.143}} & \underline{\textbf{0.053}} & \underline{\textbf{0.604}} & \underline{\textbf{0.622}}\\
Czech & $D'$ & \textbf{0.145} & \textbf{0.059} & \textbf{0.599} & \textbf{0.517}\\
Czech & straightness & 0.176 & 0.125 & 0.442 & 0.349\\
English & $k$ & 0.174 & 0.132 & 0.402 & 0.322\\
English & eccentricity & 0.261 & 0.2 & 0.344 & 0.234\\
English & closeness & 0.168 & 0.095 & 0.481 & 0.374\\
English & $n_{max}$ & 0.174 & 0.125 & 0.46 & 0.328\\
English & $m_2$ & 0.159 & 0.1 & 0.483 & 0.369\\
English & betweenness & 0.167 & 0.109 & 0.483 & 0.361\\
English & all-subg & 0.174 & 0.12 & 0.476 & 0.346\\
English & $D$ & 0.142 & 0.062 & 0.559 & 0.517\\
English & coverage & \underline{\textbf{0.131}} & \underline{\textbf{0.045}} & \underline{\textbf{0.593}} & \underline{\textbf{0.538}}\\
English & $D'$ & \textbf{0.136} & \textbf{0.053} & \textbf{0.59} & \textbf{0.524}\\
English & straightness & 0.157 & 0.111 & 0.464 & 0.365\\
Finnish & $k$ & 0.175 & 0.125 & 0.48 & 0.398\\
Finnish & eccentricity & 0.241 & 0.182 & 0.415 & 0.306\\
Finnish & closeness & 0.171 & 0.1 & 0.529 & 0.481\\
Finnish & $n_{max}$ & 0.172 & 0.115 & 0.513 & 0.4\\
Finnish & $m_2$ & 0.163 & 0.1 & 0.535 & 0.481\\
Finnish & betweenness & 0.168 & 0.105 & 0.531 & 0.472\\
Finnish & all-subg & 0.176 & 0.111 & 0.521 & 0.446\\
Finnish & $D$ & 0.162 & \textbf{0.083} & 0.551 & 0.499\\
Finnish & coverage & \underline{\textbf{0.152}} & \underline{\textbf{0.077}} & \textbf{0.562} & \underline{\textbf{0.515}}\\
Finnish & $D'$ & \textbf{0.154} & \underline{\textbf{0.077}} & \underline{\textbf{0.578}} & \textbf{0.508}\\
Finnish & straightness & 0.19 & 0.15 & 0.466 & 0.358\\
French & $k$ & 0.182 & 0.14 & 0.382 & 0.274\\
French & eccentricity & 0.246 & 0.184 & 0.348 & 0.229\\
French & closeness & 0.161 & 0.091 & 0.471 & 0.363\\
French & $n_{max}$ & 0.162 & 0.111 & 0.452 & 0.338\\
French & $m_2$ & 0.158 & 0.103 & 0.461 & 0.353\\
French & betweenness & 0.157 & 0.1 & 0.475 & 0.355\\
French & all-subg & 0.16 & 0.1 & 0.474 & 0.353\\
French & $D$ & 0.121 & 0.05 & 0.565 & 0.524\\
French & coverage & \underline{\textbf{0.105}} & \underline{\textbf{0.031}} & \underline{\textbf{0.622}} & \underline{\textbf{0.675}}\\
French & $D'$ & \textbf{0.113} & \textbf{0.043} & \textbf{0.594} & \textbf{0.531}\\
French & straightness & 0.142 & 0.091 & 0.455 & 0.365\\
Galician & $k$ & 0.204 & 0.143 & 0.372 & 0.272\\
Galician & eccentricity & 0.256 & 0.184 & 0.351 & 0.232\\
Galician & closeness & 0.177 & 0.097 & 0.477 & 0.367\\
Galician & $n_{max}$ & 0.18 & 0.107 & 0.459 & 0.346\\
Galician & $m_2$ & 0.176 & 0.1 & 0.463 & 0.361\\
Galician & betweenness & 0.174 & 0.1 & 0.485 & 0.369\\
Galician & all-subg & 0.179 & 0.105 & 0.47 & 0.353\\
Galician & $D$ & 0.131 & 0.048 & 0.576 & 0.53\\
Galician & coverage & \underline{\textbf{0.123}} & \underline{\textbf{0.037}} & \underline{\textbf{0.609}} & \underline{\textbf{0.622}}\\
Galician & $D'$ & \textbf{0.126} & \textbf{0.042} & \textbf{0.605} & \textbf{0.533}\\
Galician & straightness & 0.138 & 0.088 & 0.482 & 0.369\\
German & $k$ & 0.162 & 0.125 & 0.443 & 0.338\\
German & eccentricity & 0.229 & 0.175 & 0.382 & 0.259\\
German & closeness & 0.146 & 0.091 & 0.503 & 0.39\\
German & $n_{max}$ & 0.151 & 0.114 & 0.482 & 0.361\\
German & $m_2$ & 0.142 & 0.091 & 0.506 & 0.394\\
German & betweenness & 0.146 & 0.1 & 0.503 & 0.372\\
German & all-subg & 0.149 & 0.107 & 0.493 & 0.358\\
German & $D$ & 0.127 & 0.077 & 0.543 & 0.494\\
German & coverage & \underline{\textbf{0.113}} & \underline{\textbf{0.05}} & \underline{\textbf{0.595}} & \underline{\textbf{0.53}}\\
German & $D'$ & \textbf{0.119} & \textbf{0.062} & \textbf{0.575} & \textbf{0.515}\\
German & straightness & 0.134 & 0.107 & 0.465 & 0.365\\
Hindi & $k$ & 0.308 & 0.357 & 0.323 & 0.16\\
Hindi & eccentricity & 0.32 & 0.265 & 0.28 & 0.18\\
Hindi & closeness & 0.279 & 0.2 & 0.368 & 0.219\\
Hindi & $n_{max}$ & 0.285 & 0.2 & 0.359 & 0.219\\
Hindi & $m_2$ & 0.289 & 0.217 & 0.368 & 0.202\\
Hindi & betweenness & 0.283 & 0.2 & 0.377 & 0.224\\
Hindi & all-subg & 0.278 & 0.214 & 0.362 & 0.212\\
Hindi & $D$ & \textbf{0.201} & \underline{\textbf{0.091}} & \textbf{0.483} & \underline{\textbf{0.367}}\\
Hindi & coverage & 0.227 & 0.111 & 0.39 & 0.313\\
Hindi & $D'$ & 0.205 & \textbf{0.097} & \underline{\textbf{0.487}} & \underline{\textbf{0.367}}\\
Hindi & straightness & \underline{\textbf{0.194}} & 0.12 & 0.461 & \textbf{0.328}\\
Icelandic & $k$ & 0.162 & 0.107 & 0.473 & 0.4\\
Icelandic & eccentricity & 0.278 & 0.225 & 0.344 & 0.228\\
Icelandic & closeness & 0.178 & 0.087 & 0.518 & 0.494\\
Icelandic & $n_{max}$ & 0.191 & 0.132 & 0.462 & 0.342\\
Icelandic & $m_2$ & 0.16 & 0.091 & 0.526 & 0.488\\
Icelandic & betweenness & 0.179 & 0.1 & 0.503 & 0.404\\
Icelandic & all-subg & 0.191 & 0.115 & 0.49 & 0.355\\
Icelandic & $D$ & \textbf{0.122} & \textbf{0.05} & \textbf{0.605} & \underline{\textbf{0.53}}\\
Icelandic & coverage & 0.124 & 0.056 & 0.59 & \underline{\textbf{0.53}}\\
Icelandic & $D'$ & \underline{\textbf{0.119}} & \underline{\textbf{0.048}} & \underline{\textbf{0.624}} & \underline{\textbf{0.53}}\\
Icelandic & straightness & 0.14 & 0.077 & 0.563 & \textbf{0.504}\\
Indonesian & $k$ & 0.177 & 0.119 & 0.442 & 0.349\\
Indonesian & eccentricity & 0.242 & 0.179 & 0.383 & 0.272\\
Indonesian & closeness & 0.161 & 0.091 & 0.514 & 0.488\\
Indonesian & $n_{max}$ & 0.166 & 0.105 & 0.499 & 0.388\\
Indonesian & $m_2$ & 0.157 & 0.091 & 0.515 & 0.472\\
Indonesian & betweenness & 0.159 & 0.091 & 0.52 & 0.481\\
Indonesian & all-subg & 0.165 & 0.111 & 0.512 & 0.372\\
Indonesian & $D$ & 0.12 & 0.056 & 0.602 & 0.526\\
Indonesian & coverage & \underline{\textbf{0.103}} & \underline{\textbf{0.026}} & \underline{\textbf{0.67}} & \underline{\textbf{0.683}}\\
Indonesian & $D'$ & \textbf{0.111} & \textbf{0.045} & \textbf{0.641} & \textbf{0.531}\\
Indonesian & straightness & 0.155 & 0.111 & 0.448 & 0.367\\
Italian & $k$ & 0.216 & 0.158 & 0.337 & 0.244\\
Italian & eccentricity & 0.264 & 0.198 & 0.327 & 0.228\\
Italian & closeness & 0.192 & 0.111 & 0.449 & 0.355\\
Italian & $n_{max}$ & 0.185 & 0.125 & 0.439 & 0.307\\
Italian & $m_2$ & 0.187 & 0.115 & 0.438 & 0.349\\
Italian & betweenness & 0.18 & 0.105 & 0.463 & 0.358\\
Italian & all-subg & 0.189 & 0.118 & 0.448 & 0.338\\
Italian & $D$ & 0.15 & 0.056 & 0.551 & 0.522\\
Italian & coverage & \underline{\textbf{0.138}} & \underline{\textbf{0.042}} & \underline{\textbf{0.584}} & \underline{\textbf{0.536}}\\
Italian & $D'$ & \textbf{0.144} & \textbf{0.05} & \textbf{0.573} & \textbf{0.524}\\
Italian & straightness & 0.153 & 0.1 & 0.443 & 0.355\\
Japanese & $k$ & 0.419 & 0.423 & 0.161 & 0.084\\
Japanese & eccentricity & 0.395 & 0.382 & 0.179 & 0.101\\
Japanese & closeness & 0.407 & 0.385 & 0.176 & 0.096\\
Japanese & $n_{max}$ & 0.42 & 0.397 & 0.178 & 0.093\\
Japanese & $m_2$ & 0.416 & 0.389 & 0.178 & 0.099\\
Japanese & betweenness & 0.419 & 0.397 & 0.179 & 0.094\\
Japanese & all-subg & 0.407 & 0.388 & 0.176 & 0.094\\
Japanese & $D$ & \textbf{0.257} & \textbf{0.103} & \textbf{0.424} & \textbf{0.321}\\
Japanese & coverage & 0.283 & 0.13 & 0.319 & 0.274\\
Japanese & $D'$ & 0.268 & 0.12 & 0.389 & 0.283\\
Japanese & straightness & \underline{\textbf{0.1}} & \underline{\textbf{0.048}} & \underline{\textbf{0.555}} & \underline{\textbf{0.524}}\\
Korean & $k$ & 0.238 & 0.136 & 0.435 & 0.338\\
Korean & eccentricity & 0.26 & 0.188 & 0.384 & 0.272\\
Korean & closeness & 0.218 & 0.125 & 0.493 & 0.361\\
Korean & $n_{max}$ & 0.22 & 0.133 & 0.471 & 0.358\\
Korean & $m_2$ & 0.218 & 0.136 & 0.489 & 0.361\\
Korean & betweenness & 0.216 & 0.125 & 0.489 & 0.372\\
Korean & all-subg & 0.218 & 0.125 & 0.492 & 0.358\\
Korean & $D$ & \textbf{0.116} & \underline{\textbf{0}} & \textbf{0.708} & \underline{\textbf{1}}\\
Korean & coverage & 0.163 & \textbf{0.077} & 0.466 & \textbf{0.511}\\
Korean & $D'$ & 0.126 & \underline{\textbf{0}} & 0.675 & \underline{\textbf{1}}\\
Korean & straightness & \underline{\textbf{0.055}} & \underline{\textbf{0}} & \underline{\textbf{0.826}} & \underline{\textbf{1}}\\
Polish & $k$ & 0.209 & 0.125 & 0.443 & 0.353\\
Polish & eccentricity & 0.258 & 0.19 & 0.376 & 0.259\\
Polish & closeness & 0.196 & 0.105 & 0.5 & 0.376\\
Polish & $n_{max}$ & 0.2 & 0.125 & 0.475 & 0.37\\
Polish & $m_2$ & 0.191 & 0.1 & 0.502 & 0.446\\
Polish & betweenness & 0.193 & 0.111 & 0.504 & 0.412\\
Polish & all-subg & 0.199 & 0.125 & 0.496 & 0.358\\
Polish & $D$ & 0.148 & 0.062 & 0.578 & 0.517\\
Polish & coverage & \underline{\textbf{0.138}} & \underline{\textbf{0.05}} & \underline{\textbf{0.615}} & \underline{\textbf{0.635}}\\
Polish & $D'$ & \textbf{0.141} & \textbf{0.059} & \textbf{0.605} & \textbf{0.52}\\
Polish & straightness & 0.167 & 0.118 & 0.458 & 0.363\\
Portuguese & $k$ & 0.191 & 0.139 & 0.385 & 0.291\\
Portuguese & eccentricity & 0.246 & 0.175 & 0.358 & 0.242\\
Portuguese & closeness & 0.167 & 0.095 & 0.481 & 0.369\\
Portuguese & $n_{max}$ & 0.172 & 0.111 & 0.464 & 0.349\\
Portuguese & $m_2$ & 0.166 & 0.103 & 0.474 & 0.358\\
Portuguese & betweenness & 0.167 & 0.1 & 0.487 & 0.365\\
Portuguese & all-subg & 0.169 & 0.105 & 0.482 & 0.355\\
Portuguese & $D$ & 0.128 & 0.056 & 0.574 & 0.524\\
Portuguese & coverage & \underline{\textbf{0.119}} & \underline{\textbf{0.042}} & \underline{\textbf{0.602}} & \underline{\textbf{0.572}}\\
Portuguese & $D'$ & \textbf{0.123} & \textbf{0.048} & \textbf{0.597} & \textbf{0.528}\\
Portuguese & straightness & 0.139 & 0.094 & 0.466 & 0.369\\
Russian & $k$ & 0.159 & 0.1 & 0.484 & 0.412\\
Russian & eccentricity & 0.231 & 0.167 & 0.409 & 0.286\\
Russian & closeness & 0.153 & 0.077 & 0.541 & 0.499\\
Russian & $n_{max}$ & 0.161 & 0.1 & 0.515 & 0.404\\
Russian & $m_2$ & 0.144 & 0.079 & 0.545 & 0.494\\
Russian & betweenness & 0.153 & 0.083 & 0.539 & 0.488\\
Russian & all-subg & 0.16 & 0.1 & 0.528 & 0.461\\
Russian & $D$ & 0.123 & 0.05 & 0.615 & 0.526\\
Russian & coverage & \underline{\textbf{0.109}} & \underline{\textbf{0.029}} & \underline{\textbf{0.665}} & \underline{\textbf{0.681}}\\
Russian & $D'$ & \textbf{0.115} & \textbf{0.036} & \textbf{0.655} & \textbf{0.644}\\
Russian & straightness & 0.166 & 0.118 & 0.452 & 0.361\\
Spanish & $k$ & 0.191 & 0.133 & 0.392 & 0.299\\
Spanish & eccentricity & 0.249 & 0.167 & 0.352 & 0.246\\
Spanish & closeness & 0.17 & 0.091 & 0.489 & 0.372\\
Spanish & $n_{max}$ & 0.171 & 0.107 & 0.471 & 0.361\\
Spanish & $m_2$ & 0.165 & 0.1 & 0.485 & 0.37\\
Spanish & betweenness & 0.166 & 0.1 & 0.497 & 0.375\\
Spanish & all-subg & 0.17 & 0.1 & 0.486 & 0.361\\
Spanish & $D$ & 0.129 & 0.045 & 0.582 & 0.533\\
Spanish & coverage & \underline{\textbf{0.117}} & \underline{\textbf{0.034}} & \underline{\textbf{0.614}} & \underline{\textbf{0.662}}\\
Spanish & $D'$ & \textbf{0.123} & \textbf{0.04} & \textbf{0.611} & \textbf{0.534}\\
Spanish & straightness & 0.144 & 0.094 & 0.469 & 0.365\\
Swedish & $k$ & 0.153 & 0.111 & 0.47 & 0.38\\
Swedish & eccentricity & 0.264 & 0.2 & 0.355 & 0.241\\
Swedish & closeness & 0.157 & 0.091 & 0.524 & 0.488\\
Swedish & $n_{max}$ & 0.174 & 0.13 & 0.475 & 0.342\\
Swedish & $m_2$ & 0.146 & 0.091 & 0.528 & 0.472\\
Swedish & betweenness & 0.161 & 0.091 & 0.514 & 0.461\\
Swedish & all-subg & 0.168 & 0.111 & 0.502 & 0.361\\
Swedish & $D$ & \textbf{0.125} & \textbf{0.056} & \textbf{0.604} & \textbf{0.526}\\
Swedish & coverage & 0.126 & \textbf{0.056} & 0.58 & \underline{\textbf{0.528}}\\
Swedish & $D'$ & \underline{\textbf{0.122}} & \underline{\textbf{0.048}} & \underline{\textbf{0.624}} & \textbf{0.526}\\
Swedish & straightness & 0.133 & 0.083 & 0.556 & 0.494\\
Thai & $k$ & 0.205 & 0.129 & 0.382 & 0.291\\
Thai & eccentricity & 0.276 & 0.207 & 0.323 & 0.207\\
Thai & closeness & 0.202 & 0.111 & 0.442 & 0.342\\
Thai & $n_{max}$ & 0.202 & 0.135 & 0.409 & 0.286\\
Thai & $m_2$ & 0.187 & 0.107 & 0.446 & 0.349\\
Thai & betweenness & 0.195 & 0.125 & 0.439 & 0.315\\
Thai & all-subg & 0.208 & 0.136 & 0.421 & 0.281\\
Thai & $D$ & 0.138 & 0.058 & 0.549 & 0.517\\
Thai & coverage & \underline{\textbf{0.126}} & \underline{\textbf{0.033}} & \underline{\textbf{0.613}} & \underline{\textbf{0.635}}\\
Thai & $D'$ & \textbf{0.131} & \textbf{0.045} & \textbf{0.585} & \textbf{0.528}\\
Thai & straightness & 0.156 & 0.1 & 0.434 & 0.358\\
Turkish & $k$ & 0.194 & 0.125 & 0.448 & 0.372\\
Turkish & eccentricity & 0.218 & 0.167 & 0.423 & 0.305\\
Turkish & closeness & 0.16 & 0.1 & 0.524 & 0.488\\
Turkish & $n_{max}$ & 0.165 & 0.1 & 0.515 & 0.424\\
Turkish & $m_2$ & 0.166 & 0.1 & 0.521 & 0.481\\
Turkish & betweenness & 0.162 & 0.1 & 0.527 & 0.481\\
Turkish & all-subg & 0.16 & 0.1 & 0.523 & 0.481\\
Turkish & $D$ & \textbf{0.084} & \underline{\textbf{0}} & 0.719 & \underline{\textbf{1}}\\
Turkish & coverage & 0.113 & \textbf{0.062} & 0.587 & \textbf{0.53}\\
Turkish & $D'$ & 0.085 & \underline{\textbf{0}} & \textbf{0.738} & \underline{\textbf{1}}\\
Turkish & straightness & \underline{\textbf{0.058}} & \underline{\textbf{0}} & \underline{\textbf{0.77}} & \underline{\textbf{1}}\\

\end{longtable}

\clearpage

\section{Evaluation}
\label{appendix:evaluation_summary}

Tables \ref{tab:summary_centrality_scores_UD}
and \ref{tab:summary_centrality_scores_SUD}
\iftoggle{arxiv}{}{in this document}
summarize the distributions shown in  
\iftoggle{arxiv}{Figures  \ref{fig:boxplot_centrality_scores_UD}
and \ref{fig:boxplot_centrality_scores_SUD}, 
respectively.
}
{Figures  \ref{fig:boxplot_centrality_scores_UD}
and \ref{fig:boxplot_centrality_scores_SUD} in the main text, 
respectively. 
}

For the sake of completeness, Tables \ref{tab:complete_centrality_scores_UD} and \ref{tab:complete_centrality_scores_SUD} detail the performance of the model on each language. 
Tables \ref{tab:complete_centrality_scores_UD_small_sequences} and \ref{tab:complete_centrality_scores_SUD_small_sequences} detail the performance of the scores on small sentences with $3 \leq n \leq 6$.

\begin{table}
\caption{\label{tab:summary_centrality_scores_UD} The distribution of the performance of each model across languages depending on the evaluation metrics (the ratio $N_S/N_M$, precision, recall and $F$-measure) when UD annotation style is used. The distribution is described by the minimum value (min), the mean, the median, the maximum value (max) and the standard deviation (sd).}
\begin{tabular}{cclllll}
evaluation & centrality & min & mean & median & max & sd \\
\hline
ratio & $k$ & 0.613 & 0.656 & 0.643 & 0.733 & 0.033 \\
ratio & eccentricity & 0.655 & 0.67 & 0.67 & 0.681 & 0.007 \\
ratio & closeness & 0.95 & 0.969 & 0.968 & 0.988 & 0.009 \\
ratio & $n_{max}$ & 0.853 & 0.888 & 0.89 & 0.915 & 0.018 \\
ratio & $m_2$ & 0.948 & 0.973 & 0.975 & 0.989 & 0.01 \\
ratio & betweenness & 0.924 & 0.96 & 0.961 & 0.98 & 0.015 \\
ratio & all-subg & 0.946 & 0.976 & 0.976 & 0.992 & 0.012 \\
\end{tabular}
\end{table}

\clearpage

\begin{table}
\caption{\label{tab:summary_centrality_scores_SUD} The distribution of the performance of each model across languages when SUD annotation style is used. The format is the same as in Table \ref{tab:summary_centrality_scores_UD}. }
\begin{tabular}{cclllll}
evaluation & centrality & min & mean & median & max & sd \\
\hline
ratio & $k$ & 0.508 & 0.569 & 0.569 & 0.623 & 0.026 \\
ratio & eccentricity & 0.655 & 0.668 & 0.669 & 0.692 & 0.009 \\
ratio & closeness & 0.953 & 0.971 & 0.968 & 0.99 & 0.011 \\
ratio & $n_{max}$ & 0.828 & 0.846 & 0.848 & 0.87 & 0.011 \\
ratio & $m_2$ & 0.948 & 0.965 & 0.962 & 0.981 & 0.011 \\
ratio & betweenness & 0.919 & 0.944 & 0.941 & 0.968 & 0.016 \\
ratio & all-subg & 0.934 & 0.962 & 0.964 & 0.983 & 0.013 \\
\end{tabular}
\end{table}

\clearpage

\begin{longtable}[c]{ccllllll}
 \caption{\label{tab:complete_centrality_scores_UD} The performance of each centrality score for each language in the PUD treebank using UD annotation style. $N_M$ is the number of guesses of the model, $h$ is the number of hits of the model (the intersection between the guesses produced by the model and the actual roots of each sentence). The evaluation metrics are precision, recall and $F$-measure. For each kind of tree and evaluation metric, the best score is marked with boldface and underline whereas the 2nd best score is marked just with boldface. ``baseline'' indicates the expected precision, recall and $F$-measure of the random baseline model (Property \ref{prop:baseline}).}\\ 

 \toprule
 language & centrality & $N_M$ & $h$ & baseline & precision & recall & $F$-measure \\ 
 \midrule
 \endfirsthead

 \toprule
 language & centrality & $N_M$ & $h$ & baseline & precision & recall & $F$-measure \\
 \midrule
 \endhead

 \botrule
 \endfoot


 Arabic & $k$ & 1584 & 313 & 0.066 & 0.198 & 0.315 & 0.243\\
Arabic & eccentricity & 1461 & 407 & 0.066 & 0.279 & \textbf{0.409} & 0.331\\
Arabic & closeness & 1024 & 287 & 0.066 & 0.28 & 0.288 & 0.284\\
Arabic & $n_{max}$ & 1152 & 359 & 0.066 & 0.312 & 0.361 & 0.334\\
Arabic & $m_2$ & 1026 & 293 & 0.066 & 0.286 & 0.294 & 0.29\\
Arabic & betweenness & 1055 & 340 & 0.066 & 0.322 & 0.342 & 0.332\\
Arabic & all-subg & 1025 & 313 & 0.066 & 0.305 & 0.315 & 0.31\\
Arabic & $D$ & 1100 & 405 & 0.066 & \textbf{0.368} & 0.407 & \textbf{0.387}\\
\end{longtable}

\clearpage

\begin{longtable}[c]{ccllllll}
 \caption{\label{tab:complete_centrality_scores_SUD} The performance of each centrality score for each language in the PUD treebank using SUD annotation style. The format is the same as in Table \ref{tab:complete_centrality_scores_UD}.}\\ 
 \toprule
 language & centrality & $N_M$ & $h$ & baseline & precision & recall & $F$-measure \\ 
 \midrule
 \endfirsthead

 \toprule
 language & centrality & $N_M$ & $h$ & baseline & precision & recall & $F$-measure \\
 \midrule
 \endhead

 \botrule
 \endfoot


 Arabic & $k$ & 1726 & 411 & 0.066 & 0.238 & 0.413 & 0.302\\
Arabic & eccentricity & 1478 & 274 & 0.066 & 0.185 & 0.275 & 0.222\\
Arabic & closeness & 1028 & 321 & 0.066 & 0.312 & 0.323 & 0.317\\
Arabic & $n_{max}$ & 1202 & 328 & 0.066 & 0.273 & 0.33 & 0.299\\
Arabic & $m_2$ & 1034 & 331 & 0.066 & 0.32 & 0.333 & 0.326\\
Arabic & betweenness & 1067 & 354 & 0.066 & 0.332 & 0.356 & 0.343\\
Arabic & all-subg & 1045 & 311 & 0.066 & 0.298 & 0.313 & 0.305\\
Arabic & $D$ & 1168 & 504 & 0.066 & \textbf{0.432} & \underline{\textbf{0.507}} & \textbf{0.466}\\
\end{longtable}

\clearpage

\begin{longtable}[c]{lclllllll}
 \caption{\label{tab:complete_centrality_scores_UD_small_sequences} The performance of the centrality scores on small trees mixing languages in PUD and using UD annotation style. Structure indicates the size of the tree, followed optionally by the kind of tree (b-bistar stands for balanced bistar). Centrality indicates the centrality score chosen as representative. When only tree size is given, the representative for non-spatial centrality scores are chosen according to Property \ref{prop:classes_given_tree_size}. When tree kind is also given, the representative is for non-spatial centrality scores are choosing according to Property \ref{prop:classes_given_tree_kind_and_tree_size}. In addition, 
For star trees, we omit the spatial centrality scores that are equivalent to degree centrality, i.e. $D(v)$ and $D'(v)$ (Property \ref{prop:new_spatial_scores_on_star}).
 $N_S$ is the number of sentences, $N_M$ is the number of guesses of the classification model, $h$ is the number of hits of the model (the intersection between the guesses produced by the model and the actual roots of each sentence). The evaluation metrics are precision, recall and $F$-measure. For each kind of tree and evaluation metric, the best score is marked with boldface and underline whereas the 2nd best score is marked just with boldface.  ``baseline'' indicates the expected precision, recall and $F$-measure of the random baseline model (Property \ref{prop:baseline}).}\\ 

 \toprule
 structure & centrality & $N_S$ & $N_M$ & $h$ & baseline & precision & recall & $F$-measure \\ 
 \midrule
 \endfirsthead

 \toprule
 structure & centrality & $N_S$ & $N_M$ & $h$ & baseline & precision & recall & $F$-measure \\
 \midrule
 \endhead

 \botrule
 \endfoot


 3 & $k$ & 36 & 36 & 30 & 0.333 & \underline{\textbf{0.833}} & 0.833 & \underline{\textbf{0.833}}\\
3 & coverage & 36 & 63 & 34 & 0.333 & 0.54 & \underline{\textbf{0.944}} & 0.687\\
3 & straightness & 36 & 54 & 32 & 0.333 & \textbf{0.593} & \textbf{0.889} & \textbf{0.711}\\
4 & $k$ & 112 & 182 & 105 & 0.25 & 0.577 & \underline{\textbf{0.938}} & 0.714\\
4 & $D$ & 112 & 167 & 105 & 0.25 & \textbf{0.629} & \underline{\textbf{0.938}} & \textbf{0.753}\\
4 & coverage & 112 & 167 & 96 & 0.25 & 0.575 & \textbf{0.857} & 0.688\\
4 & $D'$ & 112 & 134 & 94 & 0.25 & \underline{\textbf{0.701}} & 0.839 & \underline{\textbf{0.764}}\\
4 & straightness & 112 & 184 & 85 & 0.25 & 0.462 & 0.759 & 0.574\\
\end{longtable}

\clearpage

\begin{longtable}[c]{lclllllll}
 \caption{\label{tab:complete_centrality_scores_SUD_small_sequences} 
 The performance of the centrality score on small trees mixing languages in PUD and using SUD annotation style. The format is the same as in Table \ref{tab:complete_centrality_scores_UD_small_sequences}.}\\ 
 \toprule
 structure & centrality & $N_S$ & $N_M$ & $h$ & baseline & precision & recall & $F$-measure \\ 
 \midrule
 \endfirsthead

 \toprule
 structure & centrality & $N_S$ & $N_M$ & $h$ & baseline & precision & recall & $F$-measure \\ 
 \midrule
 \endhead

 \botrule
 \endfoot


 3 & $k$ & 36 & 36 & 28 & 0.333 & \underline{\textbf{0.778}} & 0.778 & \underline{\textbf{0.778}}\\
3 & coverage & 36 & 62 & 34 & 0.333 & \textbf{0.548} & \underline{\textbf{0.944}} & \textbf{0.694}\\
3 & straightness & 36 & 56 & 30 & 0.333 & 0.536 & \textbf{0.833} & 0.652\\
4 & $k$ & 112 & 189 & 101 & 0.25 & 0.534 & \underline{\textbf{0.902}} & 0.671\\
4 & $D$ & 112 & 172 & 101 & 0.25 & \textbf{0.587} & \underline{\textbf{0.902}} & \textbf{0.711}\\
4 & coverage & 112 & 173 & 95 & 0.25 & 0.549 & \textbf{0.848} & 0.667\\
4 & $D'$ & 112 & 151 & 95 & 0.25 & \underline{\textbf{0.629}} & \textbf{0.848} & \underline{\textbf{0.722}}\\
4 & straightness & 112 & 247 & 94 & 0.25 & 0.381 & 0.839 & 0.524\\
\end{longtable}




\end{appendices}

}



\bibliography{anonymous_references,../../../Dropbox/biblio/rferrericancho,../../../Dropbox/biblio/complex,../../../Dropbox/biblio/ling,../../../Dropbox/biblio/cl,../../../Dropbox/biblio/cs,../../../Dropbox/biblio/maths}

\end{document}